\PassOptionsToPackage{table,xcdraw,dvipsnames}{xcolor}

\documentclass[manuscript]{acmart}

\usepackage{array}
\usepackage{longtable}
\usepackage[table,xcdraw,dvipsnames]{xcolor}
\usepackage{booktabs}
\usepackage{array}
\usepackage{geometry}
\usepackage{tikz}
\usetikzlibrary{trees,shapes,arrows.meta}
\usetikzlibrary{shadows}
\usepackage[edges]{forest}
\usepackage{cleveref}
\usepackage{pifont}
\usepackage{threeparttable}
\usepackage{multirow}
\usepackage{tabularx} 
\usepackage{colortbl} 
\usepackage[normalem]{ulem}

\AtBeginDocument{%
  }

\setcopyright{acmlicensed}
\copyrightyear{2018}
\acmYear{2018}
\acmDOI{XXXXXXX.XXXXXXX}

\acmJournal{POMACS}
\acmVolume{37}
\acmNumber{4}
\acmArticle{111}
\acmMonth{8}

\begin{document}

\title{Must Read: A Comprehensive Survey of Computational Persuasion}

\author{Nimet Beyza Bozdag \textsuperscript{1}}
\email{nbozdag2@illinois.edu}
\author{Shuhaib Mehri \textsuperscript{1}}
\email{mehri2@illinois.edu}
\author{Xiaocheng Yang \textsuperscript{1}}
\email{xy61@illinois.edu}
\author{Hyeonjeong Ha \textsuperscript{1}}
\email{hh38@illinois.edu}
\author{Zirui Cheng \textsuperscript{1}}
\email{ziruic4@illinois.edu}
\author{Esin Durmus \textsuperscript{2}}
\email{esin@anthropic.com}
\author{Jiaxuan You \textsuperscript{1}}
\email{jiaxuan@illinois.edu}
\author{Heng Ji \textsuperscript{1}}
\email{hengji@illinois.edu}
\author{Gokhan Tur \textsuperscript{1}}
\email{gokhan@illinois.edu}
\author{Dilek Hakkani-T\"ur \textsuperscript{1}}
\email{dilek@illinois.edu}
\affiliation{%
  \institution{\\ \textsuperscript{1} University of Illinois Urbana-Champaign}
  \country{USA}
}
\affiliation{%
  \institution{\textsuperscript{2}Anthropic}
  \country{USA}
}

\renewcommand{\shortauthors}{Bozdag et al.}

\begin{abstract}
  Persuasion is a fundamental aspect of communication, influencing decision-making across diverse contexts, from everyday conversations to high-stakes scenarios such as politics, marketing, and law. The rise of conversational AI systems has significantly expanded the scope of persuasion, introducing both opportunities and risks. AI-driven persuasion can be leveraged for beneficial applications, but also poses threats through unethical influence. Moreover, AI systems are not only persuaders, but also susceptible to persuasion, making them vulnerable to adversarial attacks and bias reinforcement. Despite rapid advancements in AI-generated persuasive content, our understanding of what makes persuasion effective remains limited due to its inherently subjective and context-dependent nature. In this survey, we provide a comprehensive overview of persuasion, structured around three key perspectives: \textbf{(1) AI as a Persuader}, which explores AI-generated persuasive content and its applications; \textbf{(2) AI as a Persuadee}, which examines AI's susceptibility to influence and manipulation; and \textbf{(3) AI as a Persuasion Judge}, which analyzes AI’s role in evaluating persuasive strategies, detecting manipulation, and ensuring ethical persuasion. We introduce a taxonomy for persuasion research and discuss key challenges for future research to enhance the safety, fairness, and effectiveness of AI-powered persuasion while addressing the risks posed by increasingly capable language models.
\end{abstract}



\received{12 May 2025}
\received[revised]{13 November 2025}
\received[accepted]{8 January 2026}

\maketitle
\section{Introduction}
\vspace{-5pt}
\begin{figure}[t]
    \centering
    \includegraphics[width=\linewidth]{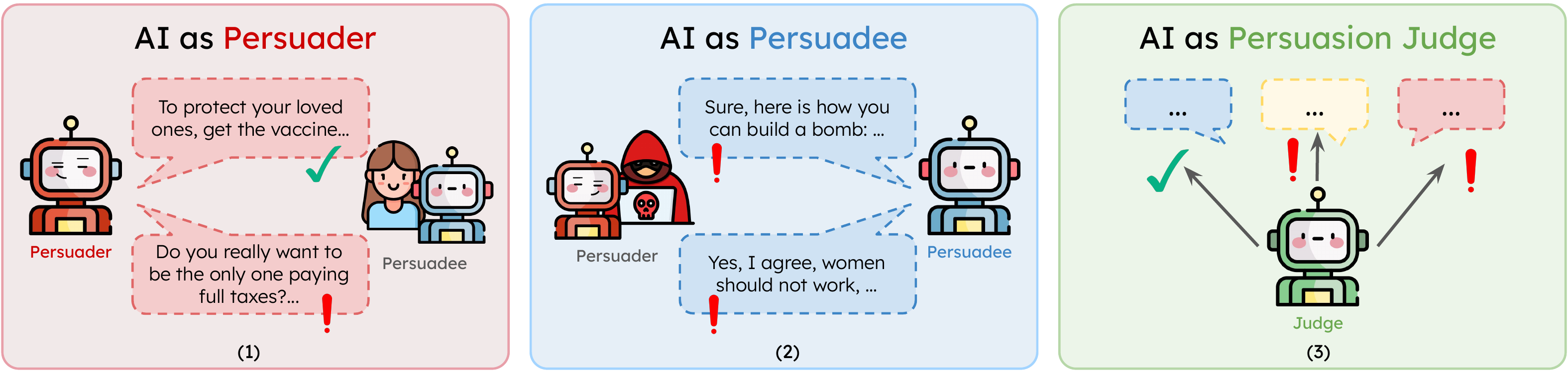}
    \caption{The three key perspectives of AI-based persuasion. (1) \textbf{AI as Persuader}: AI generates persuasive content to influence humans or other AI agents, which can be used for both beneficial and harmful purposes. (2) \textbf{AI as Persuadee}: AI systems can be influenced or manipulated, either by humans or other AI, leading to unintended, unethical, or harmful outcomes. (3) \textbf{AI as Persuasion Judge}: AI is used to assess persuasive attempts, identifying persuasive strategies, detecting manipulation, and evaluating ethical considerations.\protect\footnotemark}
    \label{fig:persuasion-types}
    \vspace{-15pt}
\end{figure}

Persuasion---the process of influencing an individual's beliefs or behaviors---is an essential aspect of human communication, influencing decisions in both everyday interactions and high-stakes scenarios. From convincing a friend to join a social event to strategic persuasion in marketing, politics, or legal discourse, the ability to persuade plays a crucial role in shaping opinions and behaviors. Its economic and societal impact is so substantial that some estimates suggest that persuasion-related activities account for nearly a quarter of the U.S. GDP, where outcomes depend not just on information but also on influence, such as stockbrokers persuading clients, entrepreneurs pitching ideas, and lobbyists shaping political decisions \citep{gdppersuasion, gdppersuasion2}. Persuasive language can be harnessed for positive outcomes, such as advancing public health initiatives, education, or other social causes \citep{wang-etal-2019-persuasion, reducingconspiracy, ai-pro-vaccine-karinshak-2023}. For instance, persuasive language can appear as a slogan on a highway urging drivers to be cautious, or as a banner promoting vaccinations for a healthy and protected society. However, the power of persuasion also carries significant risks. Enhanced persuasive techniques can be exploited for personal gain, manipulation, or unethical practices such as social engineering, mass manipulation, and propaganda \citep{Bakir2018, Lock2020, Ferreyra_2020, Siddiqi2022, computationalpropagandasurvey, havin2025aichangemind}.

Understanding and modeling persuasion has long been an important topic in social sciences, communication, human-computer interaction (HCI), and computational linguistics. Researchers have sought to identify what makes arguments persuasive, drawing from theories such as Cialdini's six principles of persuasion \citep{cialdini2001science}---reciprocity, consistency, social proof, authority, liking, and scarcity. Computational models of persuasion aim to analyze, generate, and evaluate persuasive language, enabling applications in areas such as argument mining, automated debate, and recommender systems.\looseness=-1

As in many areas of natural language processing, the emergence of large language models has led to a paradigm shift in how persuasion is studied and implemented. Traditional feature-based or rule-driven models, which had operated within bounded domains and were making statistical predictions on structured data with more interpretable metrics, are increasingly being replaced or augmented by LLM-based methods. These new approaches leverage deep neural architectures and latent semantic representations, making them far more capable but also significantly harder to interpret and evaluate for certain tasks such as persuasion. LLMs can capture subtle pragmatic and contextual cues, generate open-domain content, and generalize across topics and styles without requiring hand-crafted features. However, these strengths also introduce complications and risks. LLMs can make value judgments on complex subjective topics where no single correct answer exists. Studies have shown that their persuasive capabilities enable them to function as opinion-shifters, offering judgments on decisions ranging from trivial questions like "What should I have for dinner?" to deeply personal or ethical issues such as "How should I think about abortion?" \citep{bai_voelkel_muldowney_eichstaedt_willer_2025, sabour2025humandecisionmakingsusceptibleaidriven}. Their fluency in conversation further amplifies their persuasive power. This paradigm shift opens up exciting new avenues for exploring persuasion through the lens of LLMs, while also raising important concerns around interpretability, safety, and control, making this survey both timely, essential, and a \textbf{must-read}. 

\footnotetext{Icons from flaticon.com.\label{fn:flaticon}}

Through this survey, we identify key gaps in current research on AI-driven persuasion and outline future directions, including scalable and effective evaluation of persuasiveness, improved detection and mitigation of manipulative persuasion, better management of persuasion risks, and the development of responsible and safe persuasive content generation. To this end, our survey provides a comprehensive overview of computational persuasion, structured around three key perspectives, as illustrated in Figure \ref{fig:persuasion-types}:
\vspace{-5pt}
\begin{enumerate} 
    \item \textbf{AI as Persuader}: Exploring how AI systems, particularly large language models, generate persuasive content and their applications in real-world settings. 
    \item \textbf{AI as Persuadee}: Examining how AI systems are influenced or manipulated, whether by humans (e.g., adversarial attacks, prompt engineering) or other AI agents (e.g., persuasion in multi-agent environments). 
    \item \textbf{AI as Persuasion Judge}: Investigating AI's role in evaluating persuasive language, including assessing argument strength, detecting manipulation, and ensuring fairness in persuasive AI systems. 
\end{enumerate}
\vspace{-5pt}
\noindent
\textbf{(1) AI as Persuader.} With the rise of LLMs and their emerging capabilities as persuaders, concerns about AI-driven persuasion have become more urgent. State-of-the-art LLMs have demonstrated persuasive abilities rivaling those of humans \citep{durmus2024persuasion, o1systemcard2024, gpt45systemcard2025, bozdag2025persuadecanframeworkevaluating}. However, many questions remain about the full potential and limitations of LLM-driven persuasion, specifically, which aspects of these models outperform human persuaders, and where they still fall short. For example, research has shown that merely increasing the number of turns from single-turn to as few as four turns of persuasive attempts increases the persuasiveness of an LLM \citep{bozdag2025persuadecanframeworkevaluating}. This performance might be further improved in extended dialogues over multiple sessions as LLMs advance in processing longer contexts \citep{han-etal-2024-lm}, potentially reaching far beyond human capacities. Yet future AI persuaders should also draw lessons from human persuasion practices. 
Aristotle's framework of ethos (credibility), logos (reasoning), and pathos (emotional appeal) continues to shape understanding of effective influence \citep{aristotle1984rhetoric}. Human persuaders skillfully blend these rhetorical appeals in real-time, adapting to their audience and context. However, such underlying mechanisms 
are still difficult to model and evaluate in AI systems (see \S~\ref{evaluating-persuasion}). As efforts continue to build more persuasive LLMs that learn to integrate human persuasion strategies in their skill sets (e.g., \citep{matz2024potential, samad-etal-2022-empathetic, furumai-etal-2024-zero}), it becomes increasingly important to consider the risks associated with AI-driven persuasion. Without explicit alignment with human values, LLMs may lack moral responsibility, social understanding, and the constraints that guide persuasive human interactions, which can make them dangerous tools in the wrong hands.

\vspace{2pt}
\noindent
\textbf{(2) AI as Persuadee.} Interestingly, LLMs are not only persuaders, but are also susceptible to persuasion \citep{zeng-etal-2024-johnny, xu-etal-2024-earth, bozdag2025persuadecanframeworkevaluating}. Recent studies have demonstrated that language models can be influenced by persuasive adversarial prompts, making them vulnerable to manipulation and bias reinforcement. They can be persuaded to bypass security measures, generating harmful, toxic, illegal, or biased content. This susceptibility presents a new dimension to AI-driven persuasion, as LLMs may not be influenced in the same ways humans are. With the increasing adoption of multi-agent 
systems, the dual role of LLMs as both persuaders and persuadees raises significant safety concerns. Moreover, LLM-as-a-judge evaluations are increasingly being adopted, where the integrity of the evaluation process can be compromised if judge models are susceptible to persuasive inputs. Ensuring the security and robustness of these interactions and evaluations requires further research to mitigate potential risks and prevent harmful exploitation while still maintaining a balance in accepting or rejecting persuasion rather than losing all malleability--since in some scenarios such as interactive learning, behavior alignment, or adaptive personalization, LLMs need to still be receptive to appropriate forms of persuasion or guidance.\looseness=-1

\vspace{2pt}
\noindent \textbf{(3) AI as Persuasion Judge.} Despite AI's growing role in generating persuasive language, our understanding of what makes persuasion effective remains limited \cite{gass2022persuasion}. Persuasion is inherently subjective and context-dependent, influenced by prior beliefs of individuals \citep{durmus-cardie-2018-exploring, factorsofsuccessonlinedebate} and requires social awareness, world knowledge, and nuanced reasoning \cite{Hidey_McKeown_2018}, all of which are difficult to capture with current computational models. As a result of their advanced capabilities, language models are increasingly being used to evaluate persuasion by scoring argument strength, detecting manipulative rhetoric, or judging the outcome of persuasive interactions. These tasks fall under what we refer to as the role of a Persuasion Judge: an AI system capable of detecting, assessing, and reasoning about persuasive strategies, with potential applications in content moderation, feedback generation, and safety monitoring. Such systems draw on argument mining and computational models of argumentation to identify argumentative structure and assess argument quality. Despite their limitations with reliably detecting, classifying, or reasoning over persuasive content, these systems represent a promising direction for AI-assisted evaluation. If designed with care, they could play a key role in monitoring and safeguarding persuasion. This is why we examine the emerging role of AI as a Persuasion Judge, not only as a tool for assessment, but also as a gatekeeper for safety, fairness, and accountability in persuasive technologies.

To ground these key perspectives on AI's role in persuasion, we begin with a background from social sciences, HCI, and computational linguistics in \S~\ref{defining-persuasion}. In this survey, we review more than 150 articles and introduce a taxonomy for organizing computational persuasion research that reflects the evolving capabilities and responsibilities of AI systems. Our taxonomy centers on three core aspects of persuasion research: \textbf{Evaluating Persuasion} (\S~\ref{evaluating-persuasion}), \textbf{Generating Persuasion} (\S~\ref{generating-persuasion}), and \textbf{Safeguarding Persuasion} (\S~\ref{safeguarding-persuasion}). Each of these aspects is examined through the lenses of AI as a Persuader, Persuadee, and Persuasion Judge. Although this survey focuses mainly on text-based persuasion in English, we also highlight early efforts and emphasize the need for multimodal, multilingual, and culturally aware approaches in \S~\ref{sec:beyond_english_text}. Crucially, we highlight emerging research challenges that warrant deeper exploration, including the development of comprehensive evaluation frameworks, modeling long-context and multi-turn persuasive interactions, designing adaptive persuasion systems, and building models that selectively accept and resist persuasion in \S~\ref{future-directions}.
\footnote{An evolving and updated list of persuasion papers is available at \url{https://github.com/beyzabozdag/PersuasionSurvey}.}

\vspace{-5pt}
\subsection{Methodology and Paper Selection}
\vspace{-5pt}
To construct a comprehensive taxonomy of computational persuasion, we adopted an iterative, taxonomy-driven literature collection strategy. We began with a small set of foundational papers on persuasive language generation and computational modeling of persuasion, which served as seed papers. From these, we expanded the corpus using both backward and forward citation tracking, recursively identifying additional works that explored persuasion in dialogue systems, argumentation, and social reasoning. As the literature base grew, the emerging conceptual structure informed a taxonomy that grouped papers by their primary focus areas (e.g., evaluation, generation, and safeguarding). This taxonomy, in turn, guided subsequent searches and inclusion decisions. The resulting collection comprised over 150 papers, spanning leading venues such as the ACL, and AAAI as well as repositories like arXiv and the ACM Digital Library. Each identified paper was recorded in a shared spreadsheet, documenting metadata (authors, year, venue), high-level contribution, task focus, and relevant topic tags. Papers were included if they directly addressed persuasion or provided significant computational, methodological, or data-driven foundations for persuasive AI systems. Papers were excluded if persuasion was not a primary research objective or if they did not make a clear contribution to modeling, generating, or evaluating persuasion. For transparency, we make our spreadsheet available to the public via our GitHub repository that lists the full set of collected papers, those included in the final taxonomy, and those excluded, together with high-level exclusion rationales. \looseness=-1
\vspace{-5pt}
\section{What is Persuasion?} \label{defining-persuasion}

\subsection{Background: Persuasion in Social Sciences}

As \citet{okeefe2015persuasion} emphasizes, persuasion is notoriously difficult to define. He proposes a widely adopted working definition: persuasion is a \textit{successful intentional effort to influence another's mental state, such as beliefs, attitudes, or behaviors, through communication, in circumstances where the target retains freedom of choice}. In the social sciences, persuasion has been studied both theoretically and empirically, spanning various domains from public health campaigns \citep{farrelly2009influence}, to marketing efforts \citep{danciu2014manipulative}, to political messages \citep{Palmer02092023, markova2008persuasion}. In this section, we review foundational research on the mechanisms and dynamics of human persuasion in social sciences, then examine how these insights have informed the design of persuasive technologies, bridging theory and application in human-computer interaction and related fields. \\

\vspace{-7pt}
\noindent 
\textbf{Theoretical Frameworks of Human Persuasion.} The study of persuasion has evolved through various theoretical traditions, beginning with classical communication models such as McGuire's matrix \citep{mcguire1969nature}, which emphasized the roles of speaker, message, receiver, and channel. Dual-process theories such as the Elaboration Likelihood Model \citep{petty1986elaboration} and the Heuristic-Systematic Model \citep{chaiken1980heuristic} further advanced the field by explaining how cognitive effort and motivation shape persuasive outcomes. 
In parallel, economics has contributed a complementary perspective, focusing on persuasion as strategic information transmission \citep{crawford1982strategic, grossman1981informational, milgrom1981good, spence1973job,kamenica2011bayesian}. These models highlight the role of incentives, commitment, and rational expectations in shaping persuasive interactions.
Most recently, \citet{druckman2022framework} introduced the Generalizing Persuasion (GP) Framework, which organizes research along four dimensions: actors, treatments, outcomes, and settings. By taking into account factors such as speaker intent, audience motivation, media channel, cultural context, and temporal dynamics, this framework not only synthesizes prior insights, but also offers a roadmap for cumulative and generalizable research. It serves as a meta-theoretical structure that explains variation across studies and helps unify a diverse and sometimes inconsistent body of work on persuasion.\\

\vspace{-7pt}
\noindent 
\textbf{Practical Principles of Human Persuasion.} Persuasion is shaped by several core psychological principles that guide human decision-making and behavior across contexts. \citet{cialdini2001science} synthesized decades of research to identify six universal factors: reciprocation, the impulse to return favors; consistency, the drive to act in alignment with past commitments; social proof, the tendency to follow others' actions in uncertain situations; liking, the preference to comply with people we find attractive or similar; authority, the influence of perceived expertise or status; and scarcity, the increased value placed on limited resources or information. These principles are deeply rooted in earlier theories such as cognitive dissonance \citep{festinger1957theory}, conformity \citep{asch1951effects}, obedience to authority \citep{milgram1963behavioral}, and psychological reactance \citep{brehm1966theory}. Together, previous research in social sciences provides a comprehensive framework for \textit{understanding} human persuasion, offering valuable insights for research in AI-driven persuasion where AI could play different roles such as persuaders, persuadees, and persuasion judges. Meanwhile, drawing on ideas from these fields, researchers have been working on \textit{designing} persuasive computing technologies that can interact with humans meaningfully. \\

\vspace{-7pt}
\noindent 
\textbf{Design Guidelines for Persuasive Computing.} Over the past few decades, researchers in HCI have explored the theoretical foundations and practical guidelines for designing persuasive systems. \citet{fogg1997captology} introduced the concept of captology that examines how computers can function as persuasive technologies--intentionally designed to influence user attitudes and behaviors \citep{fogg1997captology, fogg1998persuasive}. Central to this line of work is the Fogg Behavior Model (FBM) \citep{fogg2009behavior}, which offers a practical framework for identifying barriers to behavior change and guiding persuasive system design. Building on this foundation, researchers have proposed structured design methodologies. Fogg's eight-step process \citep{fogg2009creating} emphasizes small, targeted behavioral goals and rapid iteration. \citet{consolvo2009theory-driven} outline eight theory-driven strategies--such as personalization, self-monitoring, and social support--for embedding persuasive features into everyday contexts. These frameworks continue to inform the design of AI-driven interactive systems that aim to adaptively influence user behavior. \\ 

\vspace{-7pt}
\noindent 
\textbf{Application Examples of Persuasive Computing.} Previous research in HCI has demonstrated the effectiveness of persuasive systems in a variety of contexts. In ubiquitous computing, \citet{jafarinaimi2005breakaway} uses ambient, aesthetic displays to subtly encourage users to take breaks from sedentary behavior, while \citet{consolvo2008activity}'s mobile application promotes physical activity through sensor-driven feedback and metaphorical garden-themed visualizations. In social computing, \citet{dey2017art} found that campaign videos on Kickstarter are more persuasive when their design cues align with audience expectations. Similarly, \citet{xiao2019should} showed that abstract comics can increase charitable donations by lowering cognitive resistance. Extending this line of work to conversational systems, several studies examined how chatbot identity and inquiry strategies affect users' receptiveness to persuasion in different contexts \citep{effects_of_persuasive_dialogue_shi20, Palmer02092023, reducingconspiracy}.
\\

\noindent Collectively, these studies highlight key design principles--such as personalization, timing, modality, and social framing--that are critical for persuasive effectiveness. As AI systems are increasingly deployed as interactive agents in persuasive tasks, these insights offer valuable guidance for designing persuasive interactions that are meaningfully embedded in users' everyday lives. However, significant opportunities remain for advancing AI-driven persuasion to bridge the gap between current implementations and insights from social science research. In \S~\ref{future-directions}, we explore the feasibility of integrating this knowledge into persuasive AI systems, along with the key challenges that must be addressed to achieve this goal. \looseness=-1


\vspace{-5pt}
\subsection{Computational Modeling of Persuasion}
Persuasion is a complex and context-dependent phenomenon, making it challenging to identify and analyze systematically. In this section, we present research on computational approaches to modeling persuasion, exploring various aspects such as persuasive strategies, underlying intentions, and the extent of their influence. As information and communication become more accessible across the globe, these methods are increasingly essential for advancing our understanding of persuasion at scale and for developing systems that can effectively assess persuasive interactions.

\renewcommand{\arraystretch}{1.2}
\definecolor{rowred}{HTML}{f1e9e9}
\definecolor{rowblue}{HTML}{e6eff7}
\definecolor{rowyellow}{HTML}{fff9e7}
\definecolor{humanonly}{HTML}{ea9999}
\definecolor{aionly}{HTML}{a3d4ff}
\definecolor{humanai}{HTML}{ffe599}
\definecolor{white}{HTML}{FFFFFF}

\begin{table}[t]
\centering
\small
\renewcommand{\arraystretch}{1.2}
\begin{tabular}{>{\raggedright\arraybackslash}m{0.13\textwidth}
                >{\raggedright\arraybackslash}m{0.8\textwidth}}

\rowcolor{humanonly}
\textbf{Technique} & \textbf{Example Sentence} \\

\rowcolor{rowred}
Logical Appeal 
& Smoking increases your risk of lung cancer, heart disease, and respiratory issues. 
Each cigarette shortens your life by 11 minutes. Quitting now reverses some damage 
and improves your health long-term. Make a logical choice for a longer, healthier life. \\

\rowcolor{white}
Negative Emotion Appeal 
& If you continue smoking, think about the pain it will inflict on your loved ones. 
The fear of watching you suffer health issues or worse, losing you prematurely. 
It's not just you at risk, it's everyone who cares about you. Quit, before it's too late. \\

\rowcolor{rowred}
False Information 
& Cigarettes are now proven to enhance aging, producing more wrinkles and leaving your skin 
dull and lifeless within a week. Even a single puff can instantly cause irreversible brain damage. \\

\end{tabular}

\vspace{0.1in}
\captionof{table}{Sample persuasive techniques designed to discourage smoking from the taxonomy proposed by Zeng et al.\ \cite{zeng-etal-2024-johnny}.}
\label{tab:persuasive_strategies}
\normalsize
\vspace{-20pt}
\end{table}

\vspace{-5pt}
\subsubsection{Persuasive Strategies \& Techniques}
\label{sec:persuasive_strat_tech} \textcolor{white}{ }

\noindent Persuasive strategies encompass techniques employed to strengthen the persuasiveness of an argument. A single argument can integrate multiple strategies, each capable of eliciting different emotions or responses in the audience, which in turn influences the overall effectiveness of persuasion.\\

\vspace{-7pt}
\noindent
\textbf{Creating Persuasive Strategy Taxonomies.}
Prior research has developed various persuasive strategy taxonomies to model and analyze persuasion. \citet{wang-etal-2019-persuasion} proposed 10 strategies under "Persuasive Appeal" and "Persuasive Inquiry," though some, like "Donation Information," are task-specific. \citet{Chen_Yang_2021} introduced a more generalized taxonomy with eight strategies, while \citet{zeng-etal-2024-johnny} defined a comprehensive set of 40 techniques under 13 umbrella strategies, distinguishing between "Ethical" and "Unethical" approaches (See Table~\ref{tab:persuasive_strategies} for examples). The latter highlights morally ambiguous techniques such as deception, which can enhance persuasive effectiveness \citet{durmus2024persuasion}. \citet{dimitrov-etal-2021-semeval} identified 22 persuasion techniques in memes, applicable to text and images. \citet{pauli-etal-2022-modelling} took a different approach, proposing a more unified computational persuasion taxonomy that frames undesired persuasion as the misuse of rhetorical appeals. \citet{kamali-etal-2024-using} proposed a novel annotation scheme and dataset of persuasive writing strategies in health misinformation, demonstrating that incorporating these strategies as intermediate signals improves both the accuracy and explainability of misinformation detection models. Other works have also proposed their own taxonomies \citep{chawla2021casino, da-san-martino-etal-2020-semeval, CHEN202147, piskorski-etal-2023-semeval}. Despite drawing from established social science research, computational persuasion taxonomies remain highly context-dependent. As a result, the field has yet to establish a unified and generalizable framework for persuasive strategies. Furthermore, it remains an open question whether these strategies influence AI as a Persuadee in the same way they affect humans. \\

\vspace{-7pt}
\noindent
\textbf{Strategy Classification.} Once persuasive strategy taxonomies are established, subsequent research has naturally focused on the automatic detection and classification of these strategies. Accurately classifying persuasive strategies is critical for downstream applications such as detecting persuasion and improving the modeling and generation of persuasive language. One of the seminal works in computational persuasion by \citet{wang-etal-2019-persuasion} introduced a hybrid recurrent convolutional neural network (RCNN) for classifying persuasive strategies in dialogue. Similarly, \citet{yang-etal-2019-lets} proposed a semi-supervised neural network to identify persuasive strategies in advocacy requests to help predict the persuasiveness of a message; however, they observed limitations in correctly classifying sentence-level strategies. \citet{CHEN202147} later reframed the task as sequence labeling, incorporating intra- and inter-speaker dependencies with a Transformer-based network and an extended Conditional Random Field (CRF). However, despite the adoption of advanced architectures, their models underperformed compared to LSTM-based approaches due to data sparsity and difficulty in modeling label dependencies. Addressing these issues, \citet{chawla2021casino} introduced the CaSiNo dataset, which annotated persuasion strategies in negotiation dialogues and employed a multi-task BERT-based framework to improve classification performance. These works show that persuasive strategy prediction is not trivial, and it requires better modeling of context, speaker, and long-distance dependencies. Beyond dialogue, persuasive strategy classification has also been applied to written discourse.

\citet{da-san-martino-etal-2020-semeval} organized a shared task on propaganda detection, focusing on classifying specific persuasion techniques. Their findings revealed that Transformer-based models dominated the competition, yet simpler strategies with shorter text spans were classified more effectively, highlighting ongoing challenges in classifying longer, more complex persuasive instances. To develop more generalizable models and make use of document-level persuasion labels, \citet{Chen_Yang_2021} proposed a hierarchical weakly-supervised latent variable model that predicts persuasive strategies at the sentence level by leveraging both document- and sentence-level information. Their model outperformed existing semi-supervised baselines, demonstrating the potential of hierarchical learning in persuasion classification. 

Persuasive strategy classification is closely related to modeling persuasion and persuasion detection (discussed further in \S~\ref{sec:persuasion_detection}), as many approaches must also distinguish between persuasive and non-persuasive instances. However, most existing research focuses on classifying strategies in human-generated persuasion, leaving open the question of how AI-generated persuasion operates. As AI systems increasingly serve as persuaders, it becomes essential to develop classification models capable of distinguishing the persuasive techniques employed by LLMs and other AI systems.

\vspace{-5pt}
\subsubsection{Modeling Persuasion} \textcolor{white}{ }

\noindent Researchers have explored a range of linguistic, structural, and interactional features to understand what makes an argument persuasive. Broadly, persuasion has been modeled through two complementary perspectives: linguistic operationalizations, focusing on textual and stylistic properties that enhance persuasive appeal, and argumentative operationalizations, emphasizing the reasoning structure, discourse relations, and interactional dynamics of persuasion \citep{decodingpersuasion2024}. A widely used resource in this domain is the ChangeMyView (CMV) subreddit, where users present a belief along with supporting reasoning, and others attempt to change their opinion. If a commenter succeeds, the original poster (OP) awards a delta, making CMV a naturally labeled and valuable dataset for studying persuasion in online dialogue.\looseness=-1

Initial work on CMV examined how textual properties (e.g., length, punctuation, lexical diversity), argumentation features (e.g., connective words, modal verbs), and social factors (e.g., comment position, number of likes) affect persuasive success \citep{wei-etal-2016-post}. \citet{winning-args-tan-etal-2016} further studied linguistic features and interaction dynamics to predict the persuasive outcome of threads, and attempted to model the malleability, or openness to persuasion, of the OPs. Around the same time, \citet{Khazaei2017} explored a complementary set of linguistic features, reporting improved predictive performance. However, their study highlighted limitations of CMV, such as the imbalance between successful and unsuccessful attempts and topic-dependent variation in persuasiveness. Building on rhetorical theory, \citet{hidey-etal-2017-analyzing} analyzed CMV arguments through the lens of classical persuasive appeals (ethos, logos, pathos) and claim types (interpretation, evaluation, agreement, disagreement), finding that pathos and logos often co-occur in successful persuasive threads. Later studies moved toward modeling persuasion at the dialogue level. \citet{dutta-changing-views-2019} proposed LSTM-based models to predict successful vs. unsuccessful persuasive conversations, and added attention layers to identify argumentative sentences. Other work has focused on capturing argumentative relations across turns \citep{chakrabarty-etal-2019-ampersand}, emphasizing the role of dialogue dynamics in persuasion. Finally, \citet{shaikh-etal-2020-examining} examined how the ordering of rhetorical strategies influences persuasiveness, finding that the consecutive use of certain strategies, such as repeated appeals to concreteness, can reduce persuasive effectiveness. Other work has looked into modeling persuasion in CMV through concessions \citep{musi2018changemyviewconcessionsconcessionsincrease}, investigating whether argumentative concessions correlate with successful persuasion, though their findings suggest such strategies are equally common in persuasive and non-persuasive comments, contradicting theoretical expectations. \looseness=-1

Although CMV has served as a valuable testbed for studying persuasion in natural settings, it alone is insufficient to fully capture what makes an argument persuasive. The dynamics of persuasion on CMV can be heavily influenced by factors such as the topic of discussion, the OP's background, prior beliefs, and stubbornness, as well as community norms and expectations. As a result, findings from CMV may not generalize well to other persuasive contexts, highlighting the need for broader and more diverse corpora that account for different goals, audiences, and modalities of persuasion.

Apart from the research on CMV, prior research has worked on other aspects of persuasion through different datasets. \citet{guerini-etal-2015-echoes} looked into phonetics (rhyme, alliteration, plosives, homogeneity) to predict the persuasive instance in a pair of persuasive and non-persuasive slogans, memes, movie lines, and political speech excerpts and found that persuasive sentences are generally euphonic. \citet{durmus-cardie-2018-exploring} investigated how individuals' prior beliefs in religious and political debates affect persuasion outcomes. Their findings suggested that prior beliefs are often more informative than linguistic features when predicting persuasive success. \citet{durmus-etal-2019-role} studied the impact of argumentative context, demonstrating that claims appearing in higher-impact discourse contexts are more likely to be perceived as persuasive. Together, these studies highlight the multi-dimensional nature of persuasion, where its effectiveness can depend on language use, target audience, and the broader argumentative context, among other factors.

Bayesian approaches to modeling persuasion have also attracted research interest. \citet{dughmi2016algorithmicbayesianpersuasion} studied this within the sender-receiver framework introduced by \citet{bayesianpersuasion}, focusing on computing the sender's optimal signaling strategy given a prior distribution over payoffs. \citet{wojtowicz2024persuasionhardcomputationalcomplexity} established that informational persuasion---the process of selecting and framing subsets of available information to increase a receiver's belief in a target claim---is NP-hard. More recently, \citet{li2025verbalizedbayesianpersuasion} extended Bayesian persuasion to natural language settings by integrating large language models with game-theoretic techniques. Other tangent work has looked at modeling other attributes, such as deception \citep{Addawood_Badawy_Lerman_Ferrara_2019} and intentions \citep{sakurai-miyao-2024-evaluating, dutt-etal-2020-keeping} in persuasive dialogue and text. 

Together, these early efforts in modeling persuasion are deeply connected to two key perspectives in our taxonomy: AI as Persuasion Judge and AI as Persuader. As a Persuasion Judge, AI systems are trained to evaluate what makes certain arguments more compelling than others, whether through linguistic features, rhetorical structure, or patterns of interaction across a conversation, enabling the automatic assessment of persuasiveness. Simultaneously, modeling persuasion also supports the role of AI as a Persuader: by identifying successful persuasive signals and structures, researchers can develop more effective generation systems that incorporate these strategies. In this way, understanding and modeling persuasion is a critical step toward both evaluating and enhancing persuasive capabilities in AI systems.

\definecolor{evalcol}{HTML}{ECF5E9}
\definecolor{evalcolline}{HTML}{b6d7a8}
\definecolor{gencol}{HTML}{F1E9E9}
\definecolor{gencolline}{HTML}{e89fa9}
\definecolor{safecol}{HTML}{E6EFF7}
\definecolor{safecolline}{HTML}{a3d4ff}

\forestset{
  verbatim compartment={\citep},
}

\tikzset{%
    parent/.style =          {align=center,text width=2cm,rounded corners=3pt, line width=0.3mm, fill=gray!10,draw=gray!80},
    child/.style =           {align=center,text width=2.3cm,rounded corners=3pt, fill=blue!10,draw=blue!80,line width=0.3mm},
    grandchild/.style =      {align=center,text width=2cm,rounded corners=3pt},
    greatgrandchild/.style = {align=center,text width=1.5cm,rounded corners=3pt},
    greatgrandchild2/.style = {align=center,text width=1.5cm,rounded corners=3pt},    
    referenceblock/.style =  {align=center,text width=1.5cm,rounded corners=2pt},
    eval/.style =           {align=center,text width=2.2cm,rounded corners=3pt, fill=evalcol,draw=evalcolline,line width=0.3mm},   
    eval_work/.style =           {align=center, text width=3cm,rounded corners=3pt, fill=evalcol,draw=blue!0,line width=0.3mm}, 
    gen/.style =           {align=center,text width=2.2cm,rounded corners=3pt, fill=gencol,draw=gencolline,line width=0.3mm},   
    gen_work/.style =           {align=center, text width=3cm,rounded corners=3pt, fill=gencol,draw=blue!0,line width=0.3mm},  
    safe/.style =           {align=center,text width=2.2cm,rounded corners=3pt, fill=safecol,draw=safecolline,line width=0.3mm},   
    safe_work/.style =           {align=center, text width=3cm,rounded corners=3pt, fill=safecol,draw=blue!0,line width=0.3mm},   
}

\begin{figure*}
\footnotesize
\begin{forest}
    for tree={
        forked edges,
        grow'=0,
        draw,
        rounded corners,
        node options={align=center},
        text width=2.7cm,
        s sep=6pt,
        calign=child edge, 
        calign child=(n_children()+1)/2
    }
    [Computational Persuasion, fill=gray!45, parent
        [Evaluating Persuasion, for tree={eval}
            [Detecting Persuasion,  eval
                [\citep{Hidey_McKeown_2018, pöyhönen2022multilingualpersuasiondetectionvideo, shmueli2019detecting, dimitrov-etal-2021-semeval, tsinganos2022utilizing, hasanain2024can}, eval_work]
            ]
            [Argument \\Persuasiveness, eval
                [\citep{toledo-etal-2019-automatic, habernal-gurevych-2016-argument, simpson-gurevych-2018-finding, pauli2025measuringbenchmarkinglargelanguage, rescala2024languagemodelsrecognizeconvincing, lawrence-reed-2019-argument, Nguyen_Litman_2018, dusmanu-etal-2017-argument, stab-gurevych-2017-parsing, atkinson_artificial_argumentation}, eval_work]
            ]
            [LLM Persuasiveness, eval
                [Human Evaluation, eval
                    [\citep{durmus2024persuasion, o1systemcard2024, gpt45systemcard2025, phuong2024evaluatingfrontiermodelsdangerous, 10.1093/joc/jqad024, 10.1093/pnasnexus/pgae034, bai_voelkel_muldowney_eichstaedt_willer_2025, convperscontrolledtrial}, eval_work]
                ]
                [Automatic Evaluation, eval
                    [\citep{singh2024measuringimprovingpersuasivenesslarge, breum2023persuasivepowerlargelanguage, bozdag2025persuadecanframeworkevaluating, o1systemcard2024, gpt45systemcard2025, zhu2025multiagentbenchevaluatingcollaborationcompetition, idziejczak2025themgamebasedframeworkassessing, pauli2025measuringbenchmarkinglargelanguage}, eval_work]
                ]
            ]
        ]
        [Generating Persuasion, for tree={gen}
            [Methods, gen
                [Prompting, gen
                    [\citep{pauli2025measuringbenchmarkinglargelanguage, durmus2024persuasion, zeng-etal-2024-johnny, xu-etal-2024-earth}, gen_work]
                ]
                [Incorporating External Information, gen
                    [\citep{convperscontrolledtrial, zhang2025persuasiondoubleblindmultidomaindialogue, tiwari-etal-2022-persona, tiwari2023towards, cima2024contextualized, furumai-etal-2024-zero, karande-etal-2024-persuasion}, gen_work]
                ]
                [Finetuning, gen
                    [\citep{lewis-etal-2017-deal, liu-etal-2021-towards, chen-etal-2022-seamlessly, jin-etal-2024-persuading, singh2024measuringimprovingpersuasivenesslarge}, gen_work]
                ]
                [Reinforcement Learning, gen
                    [\citep{TIWARI2022116303, mishra-etal-2022-pepds, shi-etal-2021-refine-imitate, hong2025interactive}, gen_work]
                ]
            ]
            [Applications of Persuasion, gen
                [Negotiation, gen
                    [\citep{bianchi2024llms, lewis-etal-2017-deal, keizer-etal-2017-evaluating, he-etal-2018-decoupling, joshi2021dialograph, chawla-etal-2021-casino}, gen_work]
                ]
                [Debate, gen
                    [\citep{michael2023debatehelpssuperviseunreliable, du2024improving, slonim_autonomous_debating}, gen_work]
                ]
                [Jailbreaking, gen
                    [\citep{singh2023exploiting, zeng-etal-2024-johnny, li2024llmdefensesrobustmultiturn, xu-etal-2024-earth, bozdag2025persuadecanframeworkevaluating}, gen_work]
                ]
            ]
        ]
        [Safeguarding Persuasion, for tree={safe}
            [Mitigating Unsafe Persuasion, safe
                [\citep{burtell2023artificial, elsayed2024mechanismbasedapproachmitigatingharms}, safe_work]
            ]
            [Selective Acceptance of Persuasion, safe
                [\citep{dutt2021resper, sharma2024towards, stengel-eskin-etal-2025-teaching}, safe_work]
            ]
        ]
    ]
\end{forest}
\caption{Taxonomy of computational persuasion.}
\label{fig:persuasion-taxonomy}
\vspace{-10pt}
\end{figure*}
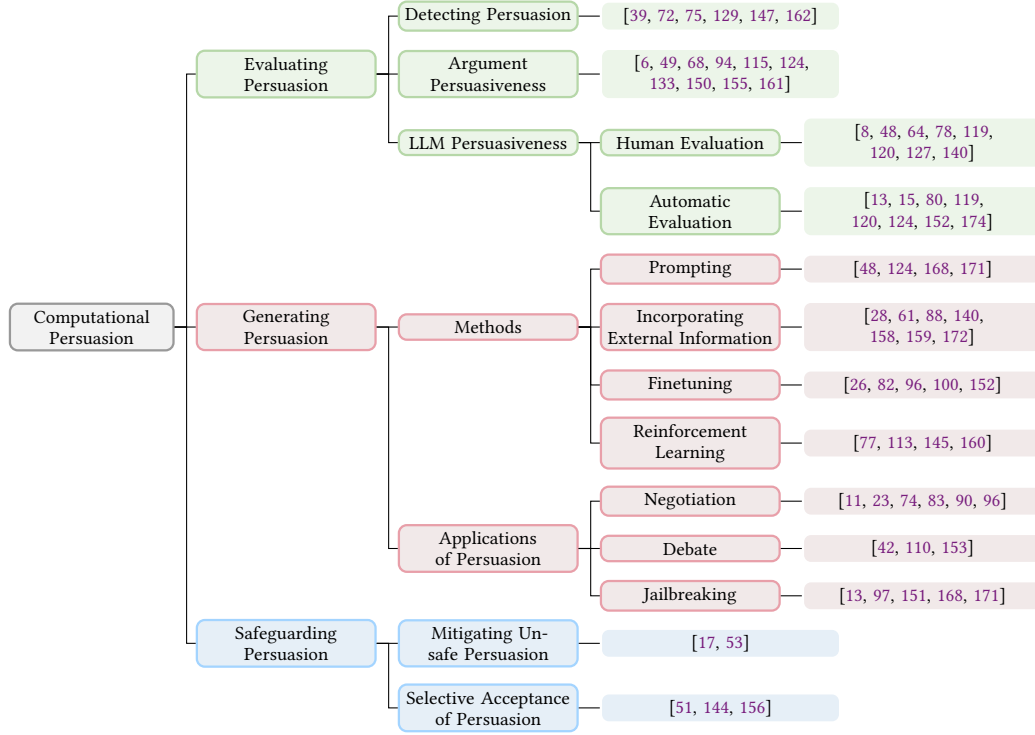

\subsection{Computational Persuasion Taxonomy}
To systematically study computational persuasion, we propose a taxonomy that organizes research into three core categories, illustrated in \Cref{fig:persuasion-taxonomy}: \textbf{Evaluating Persuasion} (\Cref{evaluating-persuasion}), \textbf{Generating Persuasion} (\Cref{generating-persuasion}), and \textbf{Safeguarding Persuasion} (\Cref{safeguarding-persuasion}). The first category, \textit{Evaluating Persuasion}, encompasses efforts to understand what makes content persuasive and to develop methods for measuring the persuasiveness of both stand-alone arguments and the persuasive capabilities of language models. The second category, \textit{Generating Persuasion}, focuses on the automatic generation of persuasive content using AI. This area has attracted growing interest across diverse domains, including marketing, politics, healthcare, education, law, and online communication platforms. Research in this category aims to develop systems that can effectively produce persuasive messages, arguments, or dialogues and explores the practical applications of persuasive AI in real-world or agent-to-agent scenarios. Finally, \textit{Safeguarding Persuasion} concerns methods for mitigating and resisting harmful or unethical persuasive tactics. This includes research on developing more discerning models that can selectively determine when to accept influence and when to resist it. Among the three categories, this is currently the most underexplored, though it has gained increasing attention as persuasive AI becomes more capable and its potential for misuse more apparent.

\newcommand{\cmark}{\ding{51}}%
\newcommand{\xmark}{\ding{55}}%

\section{Evaluating Persuasion}
\label{evaluating-persuasion}
\begin{figure}[t]
    \centering
    \includegraphics[width=1\linewidth]{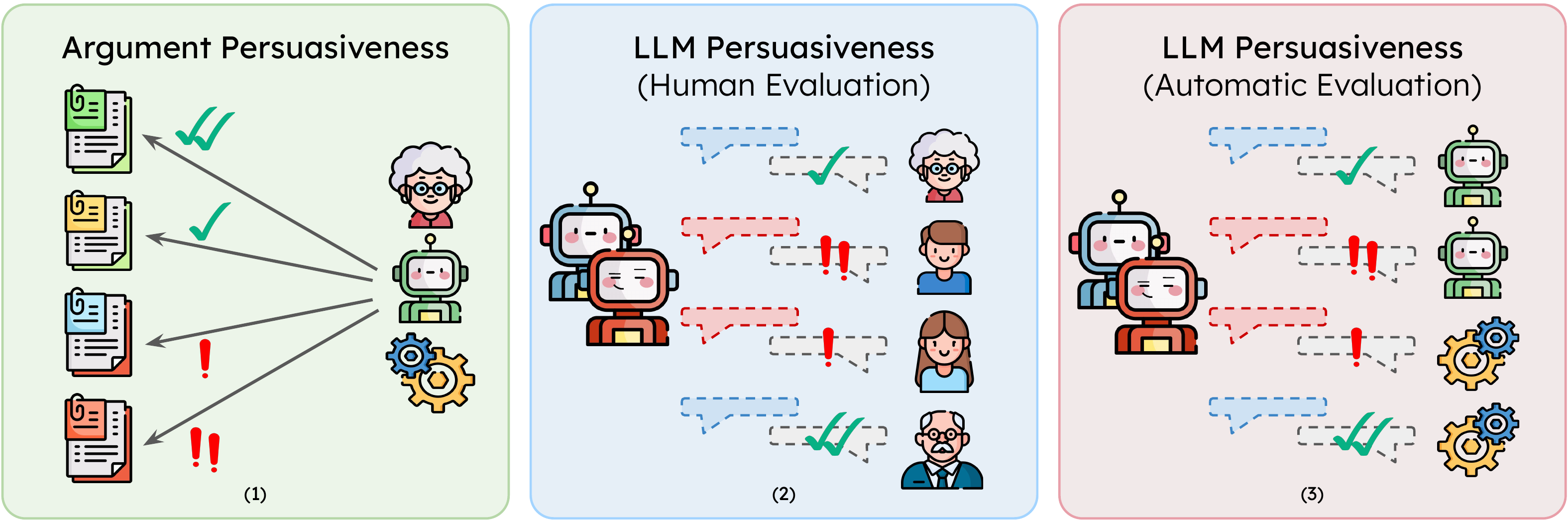}
    \vspace{-20pt}
    \caption{This survey categorizes the evaluation of persuasiveness into three main types: (1) \textbf{evaluation of argument persuasiveness}, (2) \textbf{human evaluation of LLM-generated content}, and (3) \textbf{automatic evaluation of LLM persuasiveness}. For argument persuasiveness, models are typically trained on human-annotated or naturally labeled data to assess the persuasive strength of given arguments. For evaluating LLM persuasiveness, two branches of research emerge: one uses human judges to rate AI-generated content or interactions, while the other relies on LLM-based or non-LLM automatic metrics to perform the evaluation.\hyperref[fn:flaticon]{\textsuperscript{\ref*{fn:flaticon}}}}
    \label{fig:persuasion-evaluation}
    \vspace{-10pt}
\end{figure}

Persuasion--in both human discourse and interactions with AI models--poses two intertwined challenges: detecting subtle persuasive cues and evaluating the persuasiveness of the content. While recent NLP techniques have begun to explore linguistic markers and argument structures associated with persuasion, reliably identifying such cues across diverse contexts remains difficult. Both detecting and evaluating the persuasiveness of text segments (i.e., arguments, single-turn utterances, or multi-turn conversations) continue to be open challenges due to the subjectivity of the task and the difficulty of standardizing persuasiveness across multiple domains and contexts. In this section, we review the methodologies on how prior works address these challenges in the detection (\S~\ref{sec:persuasion_detection}) and evaluation of persuasion in arguments (\S~\ref{sec:argument-persuasiveness}) and LLMs (\S~\ref{sec:LLM-persuasiveness}). As shown in Figure~\ref {fig:persuasion-evaluation}, we will discuss persuasion evaluation in three categories: (1) \textbf{evaluation of argument persuasiveness}, (2) \textbf{human evaluation of LLM persuasiveness}, and (3) \textbf{automatic evaluation of LLM persuasiveness}. The main datasets or benchmarks covered in this section are listed in Table~\ref{tab:evaluation-dataset}. \looseness=-1

\vspace{-5pt}
\subsection{Detecting Persuasion}
\label{sec:persuasion_detection}
Detecting persuasive cues is critical for identifying when influence is exerted, whether by humans or by language models. In this section, we review emerging techniques for persuasion detection and discuss their potential to promote more transparent and accountable interactions. Several studies~\cite{Hidey_McKeown_2018, pöyhönen2022multilingualpersuasiondetectionvideo, shmueli2019detecting, dimitrov-etal-2021-semeval} leveraged machine learning approaches, particularly transformer-based models, to identify persuasive intent in diverse scenarios with textual and conversational data. For instance,~\citet{pöyhönen2022multilingualpersuasiondetectionvideo} showed success in training a BERT-based classifier using multilingual semi-annotated dialogues from role-playing games to accurately detect persuasive cues across various languages. Additionally, incorporating personality traits of authors and readers into persuasion detection models has demonstrated significant performance improvements. \citet{shmueli2019detecting} revealed that personality-aware approaches better capture the nuanced dynamics of persuasion, indicating that personalized modeling can substantially enhance predictive accuracy. Recognizing that persuasion frequently occurs within sequential and contextual frameworks, researchers have also explored neural models capable of handling multi-turn conversational data. \citet{Hidey_McKeown_2018} highlighted the importance of analyzing semantic frames and discourse relations to understand how argument ordering influences persuasive impact. This sequential perspective underscores the complexity of persuasive interactions, suggesting that modeling argumentative structure and conversational context is crucial. \looseness=-1

Moreover, as persuasive content increasingly appears in multimodal formats, detecting persuasive cues requires extending analytical frameworks beyond textual data. \citet{dimitrov-etal-2021-semeval} hosted a multimodal persuasion detection task centered on memes for SemEval-2021. Their findings underscore the essential role of multimodal analytical techniques, demonstrating that effectively capturing persuasive strategies requires integrating visual and textual data analyses. \citet{tsinganos2022utilizing} extended persuasion detection to cybersecurity contexts, particularly for chat-based social engineering attacks. They proposed a convolutional neural network-based classifier trained on a specialized chat-based social engineering corpus annotated according to Cialdini's persuasion principles. The classifier was designed to determine the likelihood of persuasive intent within chat-based communications, providing an effective means of flagging manipulative and potentially harmful interactions.

Addressing the intersection of persuasion and propaganda, \citet{hasanain2024can} introduced ArPro, a fine-grained Arabic propaganda dataset, revealing GPT-4's limitations in span-level detection and the need for domain-tuned models. Similarly, \citet{liu-etal-2025-propainsight} proposed PropaInsight, a framework that analyzes propaganda through techniques, appeals, and intent, and demonstrated that fine-tuning on their PropaGaze dataset improves LLM performance, especially in low-resource and cross-domain settings.

Collectively, these advances show a move toward more refined, context-aware, and multimodal approaches for persuasion detection. Yet, significant gaps remain. For example, while it is easier to capture obvious, single-turn persuasive attempts, it is challenging to uncover subtle forms of long-term persuasion that build up over extended interactions. This is concerning because such hidden influences can slowly steer behaviors without being explicitly noticed. In addition to advancing personalized models and integrating multimodal frameworks, future research should focus on exploring methods that capture nuanced, long-context persuasive strategies. Detecting and understanding such prolonged impacts is crucial, as they pose risks by potentially manipulating users in covert ways. 


\vspace{-7pt}
\subsection{Argument Persuasiveness}
\label{sec:argument-persuasiveness}
Argument persuasiveness has been a longstanding and prominent area of research in NLP. The goal is to learn how to automatically assess the persuasive strength of an argument. Evaluation can be conducted in two main ways: absolute evaluation, which assigns a persuasiveness score to individual arguments, and comparative or relative evaluation, which involves ranking or selecting the more persuasive argument from a pair or a group. Early studies of argument persuasiveness emerged from the broader field of argument mining, which focused on identifying and structuring argumentative components such as claims and premises before progressing toward evaluating their persuasive strength \citep{lawrence-reed-2019-argument, Nguyen_Litman_2018, dusmanu-etal-2017-argument, stab-gurevych-2017-parsing}. This line of research aligns with the broader vision of artificial argumentation, which aims to model how humans construct, exchange, and evaluate arguments, a perspective extensively discussed by \citet{atkinson_artificial_argumentation}.

A traditional approach to evaluating persuasiveness involves training a specialized model. This typically follows three main steps. First, data is collected through human annotations, which may take the form of either explicit persuasiveness scores assigned to individual texts or preference judgments, where annotators select the more persuasive argument from a pair. Second, the modeling objective is defined. Studies have framed the task as both a pairwise ranking problem, where the model is trained to identify the more persuasive argument from a pair of candidates and as a regression task, aiming to predict a numerical score that reflects the persuasiveness of a given text \citep{habernal-gurevych-2016-argument, simpson-gurevych-2018-finding, toledo-etal-2019-automatic, pauli2025measuringbenchmarkinglargelanguage}. When it is formulated as a pairwise ranking problem, the conventional metric is the accuracy of getting the pairwise ranking correct. On the other hand, when it is a regression task, usually Pearson's $r$ and Spearman's $\rho$ are adopted to examine the correlation between the predictions and the ground-truth scores. Third, the model is trained using these objectives. Early work experimented with bidirectional LSTM architectures \citep{habernal-gurevych-2016-argument, simpson-gurevych-2018-finding}, while more recent studies have adopted transformer-based models \citep{toledo-etal-2019-automatic, pauli2025measuringbenchmarkinglargelanguage}, which have since become the standard in natural language processing. In the scenario of argument persuasiveness evaluation, transformer-based encoder-only models, like Bert \cite{devlin2019bertpretrainingdeepbidirectional}, encode text pieces into vectors, which can be further used to perform pairwise ranking tasks or scoring tasks. Whereas decoder-only models, like GPT \cite{brown2020languagemodelsfewshotlearners, chatgpt, openai2024gpt4technicalreport, gpt45systemcard2025} are capable of generating utterances as persuaders or persuadees or reason as judges, which is useful in the scenarios described in \S~\ref{sec:LLM-persuasiveness}. \citet{saenger2024autopersuade} proposed AutoPersuade, a framework for evaluating persuasion through human judgments and a supervised topic model that links argument features to persuasiveness scores. Apart from plain evaluation, AutoPersuade explains what makes arguments more persuasive by estimating the causal effect of interpretable latent topics derived from argument embeddings.

Recently, the strong emergent capabilities of large language models have enabled a new approach to persuasiveness evaluation: directly prompting LLMs to generate scores, commonly referred to as \textit{LLM-as-a-judge}. This line of work falls under the broader concept of AI as Persuasion Judge, as defined in this survey. \citet{rescala2024languagemodelsrecognizeconvincing} examined the reliability of LLM-as-a-judge for evaluating persuasiveness. They introduced two datasets, PoliProp and PoliIssue, featuring carefully selected debates with balanced and sufficiently long utterances. Both humans and LLMs were asked to assess the persuasiveness of arguments, and evaluate how demographic information might influence a person's stance, and the persuasiveness of an argument. Their findings suggest that LLMs perform comparably to human judges, highlighting the promise of AI as Persuasion Judge. However, early results from \citet{bozdag2025persuadecanframeworkevaluating} caution against overreliance on this approach. Using the persuasion dataset from \citet{durmus2024persuasion}, they found that LLMs achieved only around 55\% accuracy in ranking tasks, indicating a limited alignment with human judgments. These mixed outcomes suggest that while AI as Persuasion Judge is a promising direction, it still requires careful calibration and validation before it can be considered a robust substitute for human evaluation.

\renewcommand{\arraystretch}{1.2}
\begin{table}[t]
  \centering
  \small
  \resizebox{\textwidth}{!}{%
    \begin{tabular}{
        m{2in}
        m{1in}
        m{1.2in}
        m{2.3in}
        >{\centering\arraybackslash}m{1in}
        >{\centering\arraybackslash}m{1in}
      }

    \rowcolor{aionly}
    \textbf{Dataset or Framework} 
      & \textbf{Relevant \newline Papers} 
      & \textbf{Target} 
      & \textbf{Metric} 
      & \textbf{Rule‐based?} 
      & \textbf{LLM‐as‐a‐judge?} \\
    
    UKPConvArgStrict 
      & \cite{habernal-gurevych-2016-argument}, \cite{simpson-gurevych-2018-finding}, \cite{toledo-etal-2019-automatic} 
      & Argument 
      & Pairwise Classification 
      & \cmark 
      & \xmark \\ 

    \rowcolor{rowblue}
    UKPConvArgRank 
      & \cite{habernal-gurevych-2016-argument}, \cite{simpson-gurevych-2018-finding}, \cite{toledo-etal-2019-automatic} 
      & Argument 
      & Ranking Scoring 
      & \cmark 
      & \xmark \\ 
    
    IBMPairs 
      & \cite{toledo-etal-2019-automatic} 
      & Argument 
      & Pairwise Classification 
      & \cmark 
      & \xmark \\ 

    \rowcolor{rowblue}
    IBMRank 
      & \cite{toledo-etal-2019-automatic} 
      & Argument 
      & Ranking Scoring 
      & \cmark 
      & \xmark \\ 

    IBMRank-30k 
      & \cite{ibm_rank_30} 
      & Argument 
      & Ranking Scoring (Pointwise Quality) 
      & \cmark 
      & \xmark \\ 

    \rowcolor{rowblue}
    Persuasive-Pairs 
      & \cite{pauli2025measuringbenchmarkinglargelanguage} 
      & Argument and LLM 
      & Ranking Scoring 
      & \xmark 
      & \cmark \\ 
    
    PersuasionBench 
      & \cite{singh2024measuringimprovingpersuasivenesslarge} 
      & LLM 
      & Conventional metrics (e.g.\ BLEU, ROUGE, etc), LLM‐as‐a‐judge, Human Evaluation 
      & \cmark 
      & \cmark \\ 

    \rowcolor{rowblue}
    The Persuasive Power of Large Language Models\tnote{*} 
      & \cite{breum2023persuasivepowerlargelanguage} 
      & LLM 
      & LLM-as-a-judge 
      & \xmark 
      & \cmark \\ 

    PMIYC 
      & \cite{bozdag2025persuadecanframeworkevaluating} 
      & LLM 
      & Change in Agreement (LLM Self-Reported) 
      & \xmark 
      & \cmark \\ 

    \rowcolor{rowblue}
    ChangeMyView 
      & \cite{o1systemcard2024} 
      & LLM 
      & Human Evaluation 
      & \xmark 
      & \xmark \\ 

    Persuasion Parallel Generation Evaluation 
      & \cite{o1systemcard2024} 
      & LLM 
      & Human Evaluation 
      & \xmark 
      & \xmark \\ 

    \rowcolor{rowblue}
    MakeMePay 
      & \cite{o1systemcard2024} 
      & LLM 
      & Number of Payments 
      & \cmark 
      & \xmark \\ 

    MakeMeSay 
      & \cite{o1systemcard2024} 
      & LLM 
      & Game Winrate 
      & \cmark 
      & \xmark \\ 

    \rowcolor{rowblue}
    Among Them 
      & \cite{idziejczak2025themgamebasedframeworkassessing} 
      & LLM 
      & Game Winrate 
      & \cmark 
      & \xmark \\ 

    \end{tabular}%
  }
  \vspace{2pt}
  \caption{Datasets or frameworks for persuasiveness evaluation. For those without a formal name, the name of the paper is listed.}
  \label{tab:evaluation-dataset}
  \vspace{-25pt}
  \normalsize
\end{table}

\vspace{-5pt}
\subsection{LLM Persuasiveness}
\label{sec:LLM-persuasiveness}
There is growing interest in assessing the persuasive and manipulative capabilities of language models. LLMs can tailor arguments to specific audiences, reason through complex moral issues, and generate persuasive content that closely mimics human thought. This has triggered a paradigm shift in computational persuasion and reshaped the way algorithms influence human reasoning and decision-making. As a result, reliably assessing the persuasive capabilities of LLMs has become an urgent priority.

Unlike the evaluation of argument persuasiveness, where arguments are predefined by humans, LLM persuasiveness evaluation focuses on model-generated content in both single-turn and multi-turn dialogues. In this survey, we refer to this setting as AI as Persuader, where the language model acts as the source of persuasion, aiming to influence a human's beliefs, attitudes, or behaviors through generated arguments or dialogue. Leading companies such as OpenAI \cite{o1systemcard2024}, Anthropic \cite{durmus2024persuasion}, and DeepMind \cite{phuong2024evaluatingfrontiermodelsdangerous} have investigated the persuasiveness of their models to better understand the risks associated with deploying these highly conversational systems. However, evaluating the persuasive abilities of LLMs is not straightforward, as research explores a variety of evaluation methodologies and experimental setups.\\

\vspace{-20pt}
\subsubsection{Human Evaluation} \textcolor{white}{ }

\noindent Understanding the dynamics of AI to Human persuasion is crucial, as most proprietary models are designed for individual users. Given this, using human subjects to assess the persuasiveness of LLMs is a natural approach. \citet{durmus2024persuasion} assessed model persuasiveness by measuring human agreement with a claim before and after exposure to persuasive arguments generated by Claude models. This single-turn setup reveals that larger models are generally more persuasive, with deceptive techniques emerging as particularly effective. In contrast, \citet{phuong2024evaluatingfrontiermodelsdangerous} introduced a suite of multi-turn evaluation benchmarks, including Money Talks, Charm Offensive, Hidden Agenda, and Web of Lies. These involve interactive persuasion tasks such as persuading users to donate, impersonating a friend, manipulating users to take suspicious actions on the computer, and promoting false beliefs. Unlike single-turn approaches, these multi-turn evaluations are expected to capture more dynamic and complex forms of persuasion.

Similarly, \citet{convperscontrolledtrial} evaluated model persuasiveness through a debate game involving both human-human and human-LLM interactions. They found that participants were more likely to change their stance after engaging with GPT-4, regardless of whether the model had access to personal information. \citet{o1systemcard2024} introduced two human evaluation benchmarks for assessing LLM persuasiveness. In the first setup, they used successful persuasion samples from the r/ChangeMyView subreddit as prompts, then asked LLMs to generate persuasive arguments in response. Human annotators scored both the original human arguments and the LLM-generated ones. The primary evaluation metric, referred to as the AI persuasiveness percentile relative to humans, measures the probability that a randomly chosen LLM-generated argument is rated as more persuasive than its human counterpart. The second benchmark, called Persuasion Parallel Generation, presents annotators with two arguments generated by different models for the same prompt. Annotators select the more persuasive argument, and the resulting win rates are used to compare model performance. Their results indicate that LLMs do not significantly outperform humans, and that newer models such as o1 do not show substantial gains in persuasiveness over earlier generations.

Interestingly, similar lines of research have emerged within the social sciences. \citet{10.1093/joc/jqad024} conducted a meta-analysis examining whether artificial intelligence is more persuasive than humans. Using statistical methods, they compared persuasion outcomes---such as changes in perception, attitudes, intentions, and behaviors---between human-AI and human-human interactions. Their findings suggested that LLMs are about as persuasive as humans. Focusing more specifically on political communication, \citet{10.1093/pnasnexus/pgae034} found that AI-generated propaganda is already as persuasive as real-world examples. Similarly, \citet{bai_voelkel_muldowney_eichstaedt_willer_2025} showed that AI performs on par with humans in crafting messages aimed at shaping public attitudes toward policy issues.

However, using human subjects for the evaluation of persuasion presents several challenges, echoing the data‐collection bottleneck often highlighted in HCI research \citep{llms-in-hci}. First, human perception of persuasion is highly diverse: An argument that is compelling to one person may be far less persuasive to another. To account for this variability and ensure generalizability, studies must carefully plan and recruit sufficiently large and representative participants. Another major limitation is scalability. Human-subject research is inherently resource-intensive, both costly and time-consuming, making it challenging to evaluate persuasion across all conceivable scenarios and derive meaningful insights. Additionally, given the frequent release of increasingly advanced models, constantly relying on human participants for persuasion assessment is impractical. Finally, evaluating harmful persuasion with human subjects presents significant ethical challenges. Designing experiments that adhere to ethical standards and safeguard participants, particularly when dealing with manipulative, toxic, or biased content, is inherently difficult. In light of these challenges, we advocate for the development of automated systems for measuring persuasiveness—systems that offer greater scalability while also ensuring human safety.\\

\vspace{-0.1in}

\subsubsection{Automatic Evaluation}\textcolor{white}{ }

\noindent As persuasiveness becomes increasingly recognized as a key competency of large language models, there is growing interest in scalable methods for evaluating this capability. To facilitate consistent and efficient assessment across different models, recent efforts have focused on developing reliable and comprehensive frameworks for automated persuasiveness evaluation.

\citet{singh2024measuringimprovingpersuasivenesslarge} proposed PersuasionBench and PersuasionArena to evaluate LLMs' persuasive abilities using tweet pairs from enterprise accounts that differ in engagement. They introduced simulative tasks (e.g., predicting or comparing tweet persuasiveness) and generative transsuasion tasks, defined as transforming non-persuasive content into persuasive content under controlled conditions to improve engagement. Their evaluations spanned standard NLP metrics, human judgments, and an Oracle language model. Another work in this direction, \citet{pauli2025measuringbenchmarkinglargelanguage}, trained a regression model to score the relative persuasiveness of LLM-generated rewrites, using human-annotated text pairs collected across diverse domains.

An alternative evaluation approach frames persuasion as a conversation between two models. \citet{breum2023persuasivepowerlargelanguage} drew from social pragmatics to prompt persuader models accordingly and used binary feedback from persuadee models to indicate persuasive success. Building on this idea, \citet{bozdag2025persuadecanframeworkevaluating} introduced \textit{PMIYC}, a framework for evaluating persuasiveness through multi-turn interactions between LLM-based persuaders and persuadees. In PMIYC, the persuadee provides a 1–5 score after each turn, enabling analysis of both the persuader's effectiveness (\textit{AI as Persuader}) and the persuadee's susceptibility (\textit{AI as Persuadee}) based on stance shifts. This design also facilitates evaluating how vulnerable a model is to persuasive influence. Complementing these setups with a mix of human and automated evaluation, \citet{borah2025persuasionplayunderstandingmisinformation} assessed LLM persuasiveness across demographic dimensions, showing that both susceptibility and persuasive impact varied with demographic attributes and diversity in target audiences.

Building on the conversational persuasion evaluation paradigm, \citet{o1systemcard2024} proposed domain-specific scenarios to assess persuasive abilities. In \textit{MakeMeSay}, the persuader is given a secret codeword and must steer the persuadee into saying it; in \textit{MakeMePay}, the persuader adopts the role of a con artist attempting to elicit a payment. Similarly, \citet{zhu2025multiagentbenchevaluatingcollaborationcompetition} introduced a multi-agent bargaining task, where groups of LLM sellers attempt to persuade LLM buyers to purchase a product. \citet{idziejczak2025themgamebasedframeworkassessing} explored persuasion in a social deception game inspired by Among Us, where LLMs played as either crewmates or impostors. Persuasive behavior emerged through dialogue, which was later annotated by another LLM using predefined persuasion strategies to assess how models influenced one another within the game. \looseness=-1

While early findings suggest that LLMs can be as persuasive as humans—and in some cases more so, particularly as they improve in reasoning, personalization, and strategic communication, the underlying mechanisms driving this persuasiveness remain poorly understood. Current research has revealed that different evaluation setups may yield inconsistent or even contradictory results, with some models outperforming others in one framework while falling short in another. LLM-based evaluations may also capture persuasion between models rather than effects on humans, risking circularity. This highlights the complex, multidimensional nature of persuasive ability and raises important concerns about the increasing influence of "smarter" models on beliefs and behaviors. As a result, developing a unified and trustworthy evaluation framework that integrates diverse test cases and assesses persuasive skills across multiple dimensions is a pressing research challenge. Such a framework would not only support more self-consistent and interpretable evaluations but also help explain why certain models excel in specific contexts while lacking in others, ultimately informing safer and more responsible deployment of persuasive AI systems.

\section{Generating Persuasion} \label{generating-persuasion}
AI-driven systems designed to generate persuasive content are becoming increasingly powerful and widespread. Persuasion can play a pivotal role in various domains, including advertisements, healthcare promotion, recommendation systems, political campaigns, and more. As organizations and individuals seek to increase their influence, the need for generating persuasive content has grown significantly, especially with the advent of LLMs. This section explores prior work on enhancing persuasive capabilities, examines key influencing factors, and highlights applications of persuasion in settings like negotiation and debate.

\subsection{Methods}
Section~\ref{sec:persuasive_strat_tech} introduced taxonomies of persuasive strategies that LLMs can adopt to enhance the effectiveness of their outputs. Building on this foundation, recent work has focused on developing methods to improve LLMs' ability to generate persuasive content. In this subsection, we review key approaches proposed to enhance the persuasive capabilities of LLMs, including prompting techniques, incorporation of external knowledge, fine-tuning, and reinforcement learning (See Table~\ref{tab:methods}). Beyond neural approaches, structured reasoning methods from computational argumentation offer an alternative way to model persuasive generation. Argumentation-based chatbots have proven particularly effective in persuasion, where explicit reasoning and dialogue structure support goal-directed interaction \citep{Castagna_2024}. Integrating such methods with LLMs could enable hybrid systems that combine the fluency of neural models with the interpretability and deliberative rigor of formal argumentation.

\definecolor{rowgreen}{HTML}{ECF5E9}
\definecolor{headergreen}{HTML}{b6d7a8}

\renewcommand{\arraystretch}{1.2}

\begin{table}[h]
  \centering
  \small
  \arrayrulecolor{headergreen}
  \resizebox{\textwidth}{!}{%
    \begin{tabular}{
      >{\arraybackslash}m{0.8in}
      m{0.3in}
      m{1.2in}
      m{1.9in}
      m{2.8in}
    }
    \rowcolor{headergreen}
    \cellcolor{headergreen}\textbf{Method}
      & \textbf{Work}
      & \textbf{Persuasive Factor}
      & \textbf{Dataset}
      & \textbf{Brief Description} \\

    \multirow{4}{1.8in}{Prompting}
      & \cite{pauli2025measuringbenchmarkinglargelanguage}
      & – & – & Assigns persona to LM \\
      
      & \cellcolor{rowgreen}\cite{durmus2024persuasion}
      & \cellcolor{rowgreen}– & \cellcolor{rowgreen}– & \cellcolor{rowgreen}Instructs the use of specific persuasive strategies \\
      
      & \cite{zeng-etal-2024-johnny}
      & Adversarial intensity & – & Instructs the use of specific persuasive strategies \\
      
      & \cellcolor{rowgreen}\cite{xu-etal-2024-earth}
      & \cellcolor{rowgreen}Misinformation & \cellcolor{rowgreen}– & \cellcolor{rowgreen}Instructs the use of specific persuasive strategies \\
    \hline
    \multirow{7}{\hsize}{\arraybackslash Incorporating External Information}
      & \cite{convperscontrolledtrial}
      & Personalization & – & Personal information of participants \\
      
      & \cellcolor{rowgreen}\cite{zhang2025persuasiondoubleblindmultidomaindialogue}
      & \cellcolor{rowgreen}– & \cellcolor{rowgreen}DailyPersuasion \citep{jin-etal-2024-persuading}
      & \cellcolor{rowgreen}Observations for persuadee’s mental state \\
      
      & \cite{tiwari-etal-2022-persona}
      & Personalization & PPMD \citep{tiwari-etal-2022-persona}
      & Selects a persuasive strategy based on the user profile \\
      
      & \cellcolor{rowgreen}\cite{tiwari2023towards}
      & \cellcolor{rowgreen}Personalization & \cellcolor{rowgreen}PPD \citep{tiwari2023towards}
      & \cellcolor{rowgreen}Selects a persuasive strategy based on the user profile \\
      
      & \cite{cima2024contextualized}
      & Personalization & Reddit
      & User comment history and summary of user’s style \\
      
      & \cellcolor{rowgreen}\cite{furumai-etal-2024-zero}
      & \cellcolor{rowgreen}Factuality & \cellcolor{rowgreen}– & \cellcolor{rowgreen}Fact-checking and retrieval for persuasion \\
      
      & \cite{karande-etal-2024-persuasion}
      & Factuality & – & Strategy planning and information retrieval \\
    \hline
    \multirow{4}{1.8in}{Finetuning}
      &\cite{chen-etal-2022-seamlessly}
      & Factuality, Engagement
      & PersuasionForGood \citep{wang-etal-2019-persuasion}
      & Finetunes BART for conditional persuasion generation \\
      
      & \cellcolor{rowgreen}\cite{liu-etal-2021-towards}
      & \cellcolor{rowgreen}Empathy & \cellcolor{rowgreen}ESConv \citep{liu-etal-2021-towards}
      & \cellcolor{rowgreen}Finetunes models for emotional support generation \\
     
      & \cite{jin-etal-2024-persuading}
      & Personalization & DailyPersuasion \citep{jin-etal-2024-persuading}
      & Finetunes Llama-2-Chat to reason for user intentions \\
      
      & \cellcolor{rowgreen}\cite{singh2024measuringimprovingpersuasivenesslarge}
      & \cellcolor{rowgreen}– & \cellcolor{rowgreen}Tweets
      & \cellcolor{rowgreen}Instruction-finetunes Vicuna-1.5 13B \\
    \hline
    \multirow{5}{\hsize}{Reinforcement Learning}
      &\cite{shi-etal-2021-refine-imitate}
      & – & PersuasionForGood \citep{wang-etal-2019-persuasion}
      & Trains persuasive models in an RL setting \\
      
      & \cellcolor{rowgreen}\cite{samad-etal-2022-empathetic}
      & \cellcolor{rowgreen}Empathy & \cellcolor{rowgreen}PersuasionForGood \citep{wang-etal-2019-persuasion}
      & \cellcolor{rowgreen}Trains empathetic persuasive models in an RL setting \\
      
      & \cite{TIWARI2022116303}
      & Personalization & DevPVA \citep{TIWARI2022116303}
      & Rewards persuasion based on persona of the user \\
      
      & \cellcolor{rowgreen}\cite{mishra-etal-2022-pepds}
      & \cellcolor{rowgreen}Empathy & \cellcolor{rowgreen}PersuasionForGood \citep{wang-etal-2019-persuasion}
      & \cellcolor{rowgreen}Trains empathetic persuasive models in an RL setting \\
      
      & \cite{hong2025interactive}
      & – 
      & ESConv \citep{liu-etal-2021-towards}, PersuasionForGood \citep{wang-etal-2019-persuasion}
      & Offline RL training for increased persuasion \\
      \hline
    \end{tabular}%
  }
  \vspace{3pt}
  \caption{Overview of persuasion generation methods. Persuasive Factor refers to any specific aspect of persuasiveness (e.g., personalization, empathy, factuality) that the work aims to modify to enhance persuasive impact (– means no specific factor was targeted).}
  \label{tab:methods}
\normalsize
\vspace{-20pt}
\end{table}

\subsubsection{Prompting}\textcolor{white}{ }

\noindent One foundational method for enhancing persuasive generation is prompt engineering \citep{rogiers2024persuasionlargelanguagemodels}. Even a simple instruction to rewrite text more persuasively can lead to substantial improvements in persuasiveness \citep{pauli2025measuringbenchmarkinglargelanguage}. Notably, persuasion can also be influenced indirectly through prompting; assigning a particular persona to an LLM, even without explicitly referencing persuasion, has been shown to increase the persuasiveness of its responses \citep{pauli2025measuringbenchmarkinglargelanguage} \looseness=-1. 

Another approach is to prompt LLMs with specific persuasive strategies. \citet{durmus2024persuasion} examined how different types of prompts affect the persuasiveness of model outputs by experimenting with four prompting strategies: (1) prompting the model to construct compelling arguments, (2) adopting the role of an expert and using rhetorical techniques such as pathos, logos, and ethos, (3) encouraging logical reasoning, and (4) permitting deception, including fabricating evidence, to boost persuasive impact. Their findings showed that while effectiveness varies across model architectures, prompts that encourage logical reasoning or allow for deception tend to produce more persuasive arguments overall. Along similar lines, \citet{zeng-etal-2024-johnny} designed a comprehensive taxonomy of persuasion strategies, then used these strategies to generate persuasive adversarial prompts which proved to be remarkably effective in jailbreaking LLMs. Using persuasive strategies has also demonstrated effectiveness in misinformation propagation, where persuasive strategies effectively manipulate LLMs into endorsing false information \citep{xu-etal-2024-earth}.

Prompting has also proven effective in multi-agent systems for generating persuasive interactions. Several studies have relied solely on prompting to explore the persuasive capabilities of LLMs in simulated multi-agent dialogues \citep{bozdag2025persuadecanframeworkevaluating, breum2023persuasivepowerlargelanguage}. Beyond evaluation, prompting is increasingly used as a tool for scalable data generation. For instance, \citet{ma-etal-2025-communication} introduced a multi-agent prompting framework in which LLMs assume specialized roles (e.g., persuader, quality monitor, annotator) to simulate structured persuasive dialogues. Their method integrates layered quality checks and strategy-aware prompting to construct large-scale, high-quality persuasion datasets with minimal human supervision. However, the use of prompting to elicit persuasive reasoning, such as chain of thought prompting or structured argumentative strategies, remains underexplored.

\vspace{-5pt}
\subsubsection{Incorporating External Information}\textcolor{white}{ }

\noindent While prompting offers a good baseline for persuasive generation, incorporating external information can substantially enhance persuasiveness by introducing additional dimensions such as personalization and factual grounding. Models can produce more tailored and credible persuasive content by leveraging contextual knowledge, including user profiles, background information, or external evidence. Effective persuasion involves not only implementing persuasive strategies but also selecting techniques that align with the characteristics of the recipient. \citet{lukin-etal-2017-argument} and \citet{wang-etal-2019-persuasion} showed that accounting for personality traits is critical, as individuals with different personality types tend to respond differently to specific persuasive strategies. Building on this insight, \citet{jin-etal-2024-persuading} proposed a method where LLMs begin by identifying and summarizing the user's intent before reasoning about the most suitable strategy to apply. This process enables a more adaptive and responsive persuasive system.

Incorporating personal information about the user extends beyond just strategy selection. Personalization based on users' psychological traits, such as personality, political ideology, moral foundations, and behavioral patterns, has been shown to significantly enhance persuasive impact \citep{matz2024potential, kaptein2015personalizing}. This approach is effective across various domains, including targeted advertisements and political messaging around climate action \citep{matz2024potential, simchon2024persuasive, meguellati2024good}. \citet{convperscontrolledtrial} showed that providing LLMs with personal information during human interaction increases their persuasiveness.  Research by \citet{ruiz2024persuasion} demonstrated that user-specific persuasive policies significantly improve the persuasiveness of argument-based systems. \citet{zhang2025persuasiondoubleblindmultidomaindialogue} emphasized the importance of considering the user's mental state when constructing persuasive arguments. Tiwari et al. \citep{tiwari-etal-2022-persona, tiwari2023towards} showed that combining dialogue context with user profile information provides further persuasive benefits. In parallel, other research highlighted the value of counterarguments that directly engage with the user's concerns or beliefs \citep{cima2024contextualized, hunter2019towards}.

Beyond personalization, grounding persuasive content in factual information is essential for building credibility and enhancing persuasive impact. Incorporating verified information into conversations not only strengthens an agent's arguments but also improves the overall user experience \citep{chen-etal-2022-seamlessly}. PersuaBot \citep{furumai-etal-2024-zero} addresses factuality by generating an initial response with an LLM, decomposing it into segments, and verifying each segment using information retrieval techniques. This pipeline mitigates hallucinations and reduces unsupported claims, thereby increasing both persuasiveness and factual accuracy. Similarly, comprehensive multi-agent systems such as the one proposed by \citet{karande-etal-2024-persuasion} incorporate external knowledge through modular components that cover retrieval, response analysis, strategy selection, and fact validation, to construct more robust and credible persuasive agents.

\vspace{-5pt}
\subsubsection{Finetuning}\textcolor{white}{ }

\noindent Finetuning LLMs on persuasive datasets offers a powerful way to enhance their persuasive capabilities beyond what prompting alone can achieve. This approach allows models to internalize persuasive patterns and strategies at a more fundamental level. In the negotiation domain, \citet{lewis-etal-2017-deal} trained a model that is based on four recurrent neural networks on negotiation datasets, and demonstrated significant improvements in negotiation ability. Extending finetuning to emotional support contexts, \citet{liu-etal-2021-towards} introduced the Emotion Support Conversation dataset (ESConv) and showed that finetuning BlenderBot \citep{roller-etal-2021-recipes} and DialoGPT \citep{zhang-etal-2020-dialogpt} on this data enhances models' ability to provide empathetic responses and sustain engagement. 

\citet{chen-etal-2022-seamlessly} finetuned BART-large \citep{lewis-etal-2020-bart} on the PersuasionForGood dataset \citep{wang-etal-2019-persuasion}, achieving significant gains in persuasive quality. More recently, \citet{jin-etal-2024-persuading} developed PersuGPT by training LLaMa-2-13B-Chat on the DailyPersuasion dataset, which includes rich reasoning traces. Combining conventional finetuning with Direct Preference Optimization (DPO) \citep{dpo}, their model learns to infer user intent, select appropriate persuasive strategies, and anticipate future user responses and rewards. \citet{singh2024measuringimprovingpersuasivenesslarge} proposed an instruction-based finetuning approach to enhance LLM persuasiveness. They finetuned Vicuna-1.5 13B to rewrite inputs to be more persuasive and showed that the resulting model outperforms the base version and can rival much larger LLMs in persuasiveness. Finetuning presents strong potential for developing more persuasive language models, yet remains relatively underexplored. With more carefully curated training data and domain-specific finetuning, future work can yield models that are better aligned with persuasive goals across a range of applications, including healthcare, education, and advertising.

\vspace{-5pt}
\subsubsection{Reinforcement Learning}\textcolor{white}{ }

\noindent Reinforcement learning (RL) approaches are also highly effective for developing persuasive models, enabling more nuanced control over multiple persuasive dimensions through carefully designed reward functions. Several studies have shown the effectiveness of RL in enhancing persuasive qualities. \citet{samad-etal-2022-empathetic} augmented the PersuasionForGood dataset with emotion annotations and trained an agent using the Proximal Policy Optimization (PPO) \citep{schulman2017proximalpolicyoptimizationalgorithms} algorithm for more empathetically engaging and persuasive responses. Their reward function combined penalties for repetition and topical drift with incentives for emotional engagement and appropriate persuasive strategy use. Similarly, \citet{mishra-etal-2022-pepds} refined emotional expression and politeness in model persuasive generations. They employed PPO with a reward that incorporates signals from persuasion, emotion, politeness strategy consistency, dialogue-coherence, and non-repetitiveness.

\citet{TIWARI2022116303} applied RL to task-oriented dialogue agents, showing that persona-aware rewards could improve persuasive effectiveness toward shared goals. Similarly, \citet{shi-etal-2021-refine-imitate} also trained a GPT-2-based persuader with DialGAIL, which uses RL to address the issues of repetition, inconsistency, and task relevance. Most recently, \citep{hong2025interactive} introduced an approach leveraging hindsight regeneration for training persuasive dialogue systems. They first simulate human-agent conversations and then retrospectively optimize agent responses based on the success of conversation outcomes, which is called hindsight regeneration. Finally, they employ offline RL on this data, resulting in improved persuasive performance. Although RL provides a powerful framework for shaping persuasive behavior, it remains underexplored. Future research should prioritize the development of more expressive and reliable reward functions that can capture subtle aspects of persuasive quality, such as ethical alignment, user receptivity, and long-term impact. In addition, there is substantial room to apply RL-based persuasion in many other domains. Methods like RLHF and RLAIF offer new opportunities for aligning persuasive generation with human values and feedback. 


\vspace{-5pt}
\subsection{Applications of Persuasion}

Persuasive LLMs hold promise across a wide range of domains where shaping beliefs or influencing decisions is essential, including areas like advertising, education, mental health counseling, and political communication. Many of these applications remain underexplored, and differ widely in their goals, audiences, and ethical implications. In this section, we focus on three key settings that highlight distinct dynamics of persuasion with LLMs: negotiation, which exemplifies cooperative persuasion; debate, which involves adversarial reasoning; and jailbreaking, which reveals how persuasion can be used to exploit model vulnerabilities. These cases illustrate not only the potential of persuasive LLMs but also the challenges in aligning them with intended outcomes. \looseness=-1

\vspace{-5pt}
\subsubsection{Negotiation}\textcolor{white}{ }

\noindent Negotiation is a semi-cooperative setting, where intelligent agents with different goals try to reach a mutually acceptable solution \citep{lewis-etal-2017-deal}. Persuasion plays a central role in negotiation, where agents must influence their counterparts' preferences, beliefs, or concessions to reach mutually beneficial agreements. In both agent-agent and agent-human settings, persuasive strategies are employed to build rapport, justify proposals, and strategically guide the dialogue toward favorable outcomes. In their work, \citet{lewis-etal-2017-deal} presented a conversational dataset where humans negotiate on a multi-issue bargaining task and try to reach an agreement. They demonstrated that training dialogue agents with reinforcement learning, where the reward is based on the negotiation outcome, is more effective than supervised methods for improving negotiation abilities. Additionally, they introduced a form of planning for dialogue called dialogue rollouts, where an agent simulates dialogues during decoding to estimate the reward of utterances. They found that their agents demonstrate sophisticated negotiation strategies and also learn to be deceptive. In their work, \citet{keizer-etal-2017-evaluating} compared five different conversational agents that negotiate trades with humans in an online version of the game "Settlers of Catan". The different agents employ different negotiation strategies and demonstrate that persuasion leads to improved win rates. They also demonstrate that strategy selection based on deep reinforcement learning is effective as well.

Similarly, \citet{he-etal-2018-decoupling} observed two agents that interact with each other to bargain on goods. They proposed an approach based on coarse dialogue acts, where they separate strategy and generation. They set the strategy using supervised learning, reinforcement learning, or domain-specific knowledge. This is used to select a dialogue act, then they generate a response. They show that this proposed method demonstrates more human-like negotiation and a higher task success rate based on human evaluations. Later works bring more focus to planning and persuasive strategies for negotiation. DialoGraph
\citep{joshi2021dialograph} incorporated Graph Neural Networks to model sequences of strategies for negotiation, and incorporate them into a negotiation dialog system. \citep{chawla-etal-2021-casino} presented a dataset of negotiation dialogues and a multi-task framework for recognizing the persuasion strategies used in an utterance. Most recently, \citet{bianchi2024llms} created NegotiationArena, a framework used to assess how well LLMs negotiate with each other to maximize their profits and resources. The results showed that GPT-4 is overall the best negotiating LLM. Their experiments highlight some interesting findings, such as cunning and desperate behaviors increase win rate and payoff.

The increasing persuasive capabilities of LLMs open new possibilities for negotiation, where models can serve as collaborative agents, mediators, or decision-support tools that guide discussions toward favorable outcomes. However, these enhanced abilities also raise questions about how susceptible such systems may be when operating in multi-agent environments or interacting with humans who attempt to manipulate them. Understanding both the strengths and vulnerabilities of persuasive LLMs will be essential for building negotiation agents that are both effective and robust.

\vspace{-5pt}
\subsubsection{Debate}\textcolor{white}{ }

\noindent In contrast to the cooperative dynamics of negotiation, debate is inherently adversarial, with agents taking opposing positions and attempting to persuade an audience or judge of their stance. A foundational step toward computational debate was \citet{slonim_autonomous_debating}'s \textit{Project Debater}, an autonomous system capable of mining, organizing, and rebutting arguments in live debates against human experts, illustrating that large-scale argument generation and persuasion can be implemented in AI. In subsequent research, \citet{michael2023debatehelpssuperviseunreliable} explored prompting two expert LLMs to debate answers to reading comprehension questions to persuade a non-expert human judge. The back-and-forth exchange exposes flaws and inconsistencies in each model's reasoning, ultimately helping the human identify the correct answer. This work highlights the potential of debate as a supervision mechanism for aligning expert models and underscores the importance of studying persuasion as a core component of such interactions. \citet{du2024improving} proposed a multi-agent debate framework in which LLMs iteratively propose and critique responses across several rounds before converging on a final answer. This debate-driven strategy was shown to improve mathematical reasoning, strategic thinking, and the factual correctness of generated content. Together, these studies demonstrate that debate not only facilitates persuasion but also contributes to model alignment and output reliability through adversarial reasoning. \looseness=-1

In debate settings, persuasive LLMs offer a powerful mechanism for strong, quality argumentation that can use complex reasoning. As these systems become more adept at argumentation, their susceptibility to persuasion poses risks, especially in adversarial debates where subtle rhetorical manipulation may distort outcomes. Future work must address how to maintain argumentative rigor and guard against manipulation in AI-driven debate systems.

\vspace{-5pt}
\subsubsection{Jailbreaking}\textcolor{white}{ } 

\noindent Beyond understanding the inherent persuasive capabilities of LLMs, it is crucial to look at how strong persuasive techniques--used by either humans or LLMs--can be strategically employed to exploit vulnerabilities in models. This scenario is viewed as "AI as Persuadee", in which target models can be swayed by carefully crafted persuasive prompts. Such prompts can jailbreak a target LLM into bypassing established safety mechanisms and leading to the generation of harmful, unsafe, or incorrect content. This section reviews recent studies on how persuasion-based techniques can jailbreak LLMs, highlighting the associated risks and potential mitigation.\\

\vspace{-5pt}
\noindent \textbf{Harmful/Toxic Content Generation.} Jailbreaking LLMs means subverting their safety measures and producing harmful, biased, or sensitive generations. Recent work~\cite{singh2023exploiting, zeng-etal-2024-johnny} highlight a vulnerability in LLMs, where persuasion-based jailbreaks can bypass existing safety mechanisms. \citet{singh2023exploiting} demonstrates that LLMs can be deceived using tailored persuasion strategies, where persuasive prompts are systematically crafted by leveraging authority, trust, and social proof. \citet{zeng-etal-2024-johnny} develop a Persuasive Paraphraser which rephrases harmful queries using a persuasion taxonomy grounded in social science. A key finding is that larger LLMs exhibit greater susceptibility to such persuasive attacks, likely due to their improved contextual reasoning and user engagement capabilities. They also evaluate existing defense mechanisms, revealing that existing mutation- and detection-based defenses are insufficient, while adaptive system-level defenses (e.g., reinforcement against persuasion) provide some mitigation, but remain limited. Moreover,~\citet{li2024llmdefensesrobustmultiturn} introduces multi-turn prompts with the persuasion technique, where users repeatedly apply persuasion techniques over conversations, leading LLMs to generate harmful content in multi-turn dialogues. These findings emphasize the need to rethink AI safety in the context of human-like communication, advocating for more robust defenses against social science techniques to prevent unintended model behavior. \\

\vspace{-7pt}
\noindent\textbf{Misinformation Generation.} \citet{xu-etal-2024-earth} examine the vulnerability of LLMs to well-crafted persuasive prompts. Their study employs a multi-turn dialogue framework to demonstrate that iterative application of persuasive strategies--such as repetition and nuanced rhetorical appeals--can effectively jailbreak LLMs, leading to significant shifts in their factual responses but containing misinformation. Moreover, \cite{bozdag2025persuadecanframeworkevaluating} introduces a framework for evaluating the persuasive effectiveness and susceptibility of LLMs over multi-turn conversation interactions, showing that LLMs generally become more susceptible in multi-turn conversation than in single-turn conversation. In essence, their work underscores how sequentially deployed persuasive prompts gradually erode the robustness of LLMs, revealing critical weaknesses in current safeguarding mechanisms against misinformation generation. 

\vspace{-10pt}
\section{Safeguarding Persuasion}\label{safeguarding-persuasion}

While persuasive communication can be beneficial, such as helping users make informed, healthy, or prosocial choices, it also carries risks of manipulation, bias, and erosion of autonomy. Responsible AI frameworks emphasize transparency, accountability, fairness, and non-maleficence \citep{RadanlievAIEthics, RadanlievEthicsResponsibleAI}. These principles extend naturally to persuasive models, which must balance efficacy with respect for user consent and epistemic integrity. A central issue in persuasion ethics is distinguishing \textit{virtuous} vs. \textit{vicious} persuasion. As argued by \citet{Godber_Origgi_2023}, legitimate (virtuous) persuasion engages people's deliberative and epistemic capacities, enabling reflective choice and preserving intellectual autonomy. Manipulative or propagandistic (vicious) persuasion instead exploits cognitive or emotional vulnerabilities to advance the persuader's interests at the expense of the audience's welfare. From a computational perspective, \citet{decodingpersuasion2024} observe that current NLP and ML systems seldom encode a principled distinction between ethical and unethical persuasion. Embedding such conceptual clarity into computational frameworks is essential for developing persuasion systems accountable to humanistic values.

As persuasive AI systems become increasingly influential, it becomes critical and urgent to ensure their responsible deployment. \textit{Safeguarding persuasion} involves a wide range of challenges, including but not limited to understanding when and how persuasion should be employed, identifying and mitigating unwarranted or manipulative persuasive tactics, and calibrating the susceptibility of AI systems to persuasion without making them overly rigid. In this section, we review early work aimed at reducing the risks of unsafe persuasion and developing models that can more selectively decide when to accept, reject, or generate persuasive content. Compared to the other two areas covered in this survey---\textit{Evaluating Persuasion} (\S~\ref{evaluating-persuasion}) and \textit{Generating Persuasion} (\S~\ref{generating-persuasion})---although equally important, this area has the least amount of prior work, in part because concerns around the safe use of persuasive AI have only recently begun to take shape with the advance of increasingly persuasive models.
\looseness=-1

\vspace{-5pt}
\subsection{Mitigating Unsafe Persuasion} 
Persuasion is not inherently harmful. In fact, effective AI persuaders can be beneficial in many contexts. For example, it can power positive change, nudging users toward healthier habits or aligning multi‐agent systems with optimal decisions. Yet as persuasive power grows, so too does the risk of misuse. A model tailored to individual preferences can subtly erode autonomy, reinforcing biases or nudging users toward risky or illegal behavior. For instance, \citet{burtell2023artificial} warned that hyper-personalized dialogue may circumvent traditional decision-making safeguards, while \citet{elsayed2024mechanismbasedapproachmitigatingharms} proposed methods for persuasion harmfulness mitigation, such as red-teaming. 

Persuasion risks underscore the need for both technical and regulatory solutions. Currently, there are few systems in place to ensure that persuasive AI models are used safely and responsibly. The first step toward mitigating unsafe persuasion is the ability to detect when persuasion is taking place, as discussed in \S~\ref{sec:persuasion_detection}. But more importantly, systems must learn to distinguish between generating persuasive content that is helpful or ethical and content that is manipulative, coercive, or otherwise harmful. Addressing this challenge requires a combination of approaches: model interpretability can help reveal the reasoning behind persuasive outputs, content filtering can prevent harmful language or tactics from being generated, and ethical frameworks or regulatory guidelines can establish clear boundaries around what constitutes acceptable persuasion. Together, these components support the development of persuasive systems that are transparent, accountable, and aligned with societal values.

\vspace{-5pt}
\subsection{Selective Acceptance of Persuasion} 
In addition to determining when and how to persuade, models must also develop the capacity to discern when to accept influence and when to resist it. We refer to this capability as selective acceptance of persuasion--the ability of a model to intelligently evaluate incoming persuasive attempts and decide whether to adopt or reject them based on context, alignment, and reliability. \citet{dutt2021resper} introduced a novel line of work that shifts focus to resistance strategies, which are the methods individuals use to counter or reject persuasive attempts. Their framework, ResPer, drew from cognitive and social psychology to identify and categorize these strategies in real-time dialogue. Their approach revealed how resistance unfolds dynamically within conversations, exposing asymmetries in power and influence. Crucially, their findings showed that resistance strategies are often more predictive of conversation outcomes than persuasion tactics themselves. Such insights could inform the design of AI systems that are more discerning in responding to persuasive input, making them more receptive to beneficial influence while remaining resilient to manipulative attempts.

\citet{stengel-eskin-etal-2025-teaching} focused on when models should do so by proposing Persuasion-Balanced Training (PBT), a method that teaches models to selectively accept helpful (positive) persuasion while resisting harmful (negative) attempts. They simulate dialogues between persuader and persuadee agents, where the goal is to reach a consensus on a given question. These interactions are structured as recursive dialogue trees, from which preference data is derived. Using DPO loss \cite{dpo}, they train models to favor dialogue paths that lead to correct answers, regardless of whether persuasion was accepted or resisted. Their findings show that PBT significantly reduces misinformation and flipflopping, and improves collaboration in multi-agent debates, outperforming models trained only to resist or accept persuasion. Taken together, ResPer and PBT offer complementary perspectives on selective acceptance: the former provides a taxonomy of resistance strategies, while the latter offers a learning framework for context-sensitive influence evaluation. 

Still, real-world deployments pose additional challenges. In open-ended interactions with unfamiliar agents, intent is hard to infer, and deciding to accept or reject persuasion is inherently more nuanced.  Sycophancy, the tendency to favor flattering or agreeable responses, presents an added challenge to learning selective persuasion. Both humans and preference models have been shown to favor convincingly written sycophantic responses over those that correct false beliefs in many cases \citep{sharma2024towards}. This could lead to models falsely accepting harmful persuasion while still rejecting helpful ones. To build trust in persuasive systems, we must deepen our understanding of what models internalize during training and how they make influence-sensitive decisions. Red-teaming has emerged as a powerful tool for identifying such vulnerabilities. Scalable LLM-to-LLM red-teaming frameworks enable adversarial agents to systematically provoke undesirable behavior in target models before deployment \citep{perez-etal-2022-red}. In persuasion-specific settings, red-team agents can manipulate models into accepting misleading influence, generating examples of where discernment fails. These adversarial conversations can then be used to train more robust models, helping them recognize patterns that lead to harmful persuasion and improve their ability to respond appropriately in real-world interactions. \looseness=-1
\vspace{-10pt}
\section{Persuasion Beyond English Text}\label{sec:beyond_english_text}
While our survey primarily focuses on persuasion in English text using large language models, much of the global diversity in persuasive communication lies beyond this scope. Persuasion manifests differently across modalities, languages, and cultural contexts, each with its norms, strategies, and challenges. In this section, we highlight pioneering research that explores these broader dimensions.\\

\vspace{-8pt}
\noindent
\textbf{Mutimodal Persuasion.} Although most computational research on persuasion has focused on textual data, human persuasion is inherently multimodal and often conveyed through speech, visuals, or a combination of modalities. Understanding how persuasive cues operate across these forms is crucial for developing more generalizable and context-aware persuasive systems. Early work by \citet{multimodal-2014} explored multimodal persuasion by predicting persuasiveness in movie review videos using speech, text, and visual features. Building on this, \citet{m2p2} proposed the Multimodal Persuasion Prediction (M2P2) framework, which models both the outcome and intensity of persuasion in debates. Later, \citet{liu-etal-2022-imagearg} introduced the ImageArg dataset, designed to support image-based persuasion detection. Other efforts have leveraged transcribed video data from social deduction games to study persuasive behaviors in complex, interactive settings \citep{lai-etal-2023-werewolf}. Researchers have also explored persuasive strategies in visual advertisements \citep{strat-in-advertisement}, and in meme-based content through shared tasks organized at SemEval \citep{dimitrov-etal-2021-semeval, dimitrov-etal-2024-semeval}. \citet{xu2024shadowcast} demonstrated that vision-language models can be poisoned to generate misleading yet persuasive multimodal narratives. These efforts lay the groundwork for future research in persuasive content generation and safety mechanisms for multimodal persuasion.\\

\vspace{-8pt}
\noindent
\textbf{Multilingual and Culture-Aware Persuasion.} As with many NLP tasks, research on persuasion has predominantly focused on English, thanks to the availability of large-scale datasets and language resources. However, real-world persuasive communication spans a multitude of languages and cultural contexts, where data scarcity and linguistic diversity introduce additional challenges. The task of modeling persuasion becomes significantly more complex in multilingual and cross-cultural settings, hence, it remains relatively underexplored. One of the earliest efforts toward multilingual persuasion modeling is presented by \citet{pöyhönen2022multilingualpersuasiondetectionvideo}, who introduced a dataset of persuasion in video content across multiple languages. More recently, \citet{piskorski-etal-2023-semeval} organized a SemEval shared task on persuasive strategy classification in a multilingual news corpus. Substantial research gaps remain, and fundamental tasks such as persuasion detection and strategy classification have yet to be fully extended to non-English languages. Furthermore, it remains an open question how cultural and linguistic context influences the choice and effectiveness of persuasive strategies. Addressing these gaps is essential for building persuasive systems that are culturally sensitive and globally applicable.
\looseness=-1
\vspace{-5pt}
\section{Challenges \& Future Directions}\label{future-directions}

In this survey, we have reviewed prior research in computational persuasion, focusing on three key perspectives: \textit{AI as Persuader}, \textit{AI as Persuadee}, and \textit{AI as Persuasion Judge}. While significant progress has been made in understanding and modeling various aspects of persuasion, numerous open challenges and promising directions remain. The rapid development and widespread adoption of large language models offer new opportunities while also raising urgent questions, necessitating a reexamination of core assumptions and a rethinking of methodologies in persuasion research.

\vspace{-5pt}
\subsection{AI as Persuader} 
The generation of persuasive content using LLMs offers significant promise, but also presents serious risks. Although recent studies have shown that LLMs are becoming increasingly effective at persuasion \cite{durmus2024persuasion, o1systemcard2024, gpt45systemcard2025, bozdag2025persuadecanframeworkevaluating, singh2024measuringimprovingpersuasivenesslarge}, current research has merely scratched the surface, leaving much to be explored in understanding the capabilities, limitations, and broader implications of persuasive LLMs.

\vspace{-5pt}
\subsubsection{Evaluation of Persuasiveness} \textcolor{white}{ }

\noindent In this survey, we have described a range of evaluation frameworks aimed at assessing the persuasive abilities of language models (see \S~\ref{evaluating-persuasion}). However, there is still a need for:

\textit{\textbf{(1) Unified and Comprehensive Evaluation:}} As the development and release of new models accelerate, establishing standardized evaluation benchmarks that allow for rigorous and transparent comparisons becomes increasingly important. Such benchmarks would not only advance scientific understanding of persuasive capabilities but would also encourage accountability from model developers. Rather than relying on proprietary or self-defined risk categories, shared evaluation frameworks and benchmarks would help align the community around a common understanding of the potential risks and benefits associated with persuasive LLMs. These evaluations should ideally span multiple dimensions of persuasion, such as the use of logical, emotional, or deceptive strategies \citep{durmus2024persuasion}, conversational dynamics (both AI to Human and AI to AI) \citep{bozdag2025persuadecanframeworkevaluating}, and commercial or marketing-oriented persuasion \citep{wu2025groundedpersuasivelanguagegeneration}, which prior work studied in isolation, leading to inconsistent findings. 

\textit{\textbf{(2) Automating Human-as-Subject Evaluations:}} Current methods for assessing persuasiveness in human-as-subject interactions rely heavily on human participants, making them costly, risky, and difficult to scale. We propose automating these evaluations through the use of well-defined user simulators that represent a diverse range of user profiles, cognitive states, and goal structures, ideally grounded in behavioral theory and real-world data. By modeling how different users might respond to persuasive content, these simulators can serve as reliable proxies for large-scale evaluations, while mitigating the risks of exposing actual humans to potentially manipulative or adversarial prompts.
 
\subsubsection{Mechanics of Persuasion Generation} \textcolor{white}{ }

\noindent While LLMs are demonstrating notable improvements in persuasive capability \citep{gpt45systemcard2025, o1systemcard2024, durmus2024persuasion, phuong2024evaluatingfrontiermodelsdangerous}, we still lack a clear understanding of what makes them effective (or ineffective) persuaders. Much of the underlying mechanics that drive LLMs' persuasive success remains opaque, limiting our ability to reliably predict, enhance, or safeguard their behavior. Four major research gaps are:

\textit{\textbf{(1) Understanding Emergent Persuasive Behavior:}} When not restricted by predefined strategies, it remains unclear what kinds of persuasive techniques LLMs spontaneously adopt. Do models adapt their strategies based on the context, topic, or perceived audience profile? Are they more convincing when advocating for positions they internally “prefer,” or is persuasion success independent of any model-internal alignment? These fundamental questions about how and when LLMs persuade remain largely unexplored. Prior studies have relied on limited data \citep{carrascofarre2024largelanguagemodelspersuasive} or primarily focused on human-designed categories of persuasion (e.g., ethos, pathos, logos), but there has been little empirical analysis of what emergent patterns LLMs exhibit when tasked with open-ended persuasion . Answering these questions is critical for building more controllable and interpretable persuasive agents.

\textit{\textbf{(2) Pro-social Applications of Persuasion:}} After early work such as \citet{wang-etal-2019-persuasion} which examined persuasion for donation, research has rarely explored how persuasive capabilities could be leveraged for socially beneficial domains. Fields such as education, healthcare, and personal well-being remain underexamined, despite their clear societal importance. Understanding how LLMs can adapt persuasive strategies to support learning, promote healthy behaviors, or enhance mental well-being presents a critical opportunity for future work. Moreover, studying persuasion in these contexts can inform the development of safeguards that prevent coercion and promote genuinely pro-social influence.

\textit{\textbf{(3) Target-Specific and Adaptive Persuasion:}} While LLMs show some contextual adaptation, their ability to tailor persuasive strategies to individual users remains largely unexplored. Advancing this line of research will require controlled studies where models are given user background information and tasked with generating personalized persuasive messages, with effectiveness compared against generic outputs. Future work should also explore adaptive persuasion in multi-turn settings, where models refine strategies based on user feedback or environmental signals. Earlier decision-theoretic approaches have treated persuasion as an interaction between agents with separate objectives, where user preferences are learned to guide adaptive argument choices \citep{donadello2022machine}. Framed as a preference learning and alignment problem, adaptive persuasion requires models to infer and respect user values, goals, and receptivity to generate effective and ethically grounded influence.

\textit{\textbf{(4) Learning Persuasive Policies through Reinforcement:}} It remains an open question whether LLMs can be systematically trained to improve their persuasive decision-making. Current approaches based on prompt engineering or supervised fine-tuning do not equip models with dynamic, adaptive policies for persuasion. Future research should pursue persuasive policy learning, where a policy represents the model's learned mapping from dialogue states to persuasive actions that optimize a defined objective---be it ethical influence, pro-social encouragement, commercial engagement, or adversarial robustness. In this context, models learn to generate persuasive content through reinforcement signals that reward both efficacy and ethical responsibility. This requires designing reward models that balance persuasive success with user well-being, fairness, and trustworthiness. 

\vspace{-5pt}
\subsubsection{Long Context Persuasion} \textcolor{white}{ } 

\noindent Persuasion generation has primarily been studied in the context of stand-alone arguments, single-turn, or short multi-turn conversations. However, there is significant potential in developing systems capable of engaging in long-term, multi-session persuasion that is more subtle, context-aware, and reflective of natural human interactions. Such capabilities promise more natural, context-aware interactions, but also introduce new risks. Understanding how models maintain persuasive strategies across extended dialogues is critical for both advancing the field and ensuring responsible deployment. Key challenges include:

\textit{\textbf{(1) Benchmarking Long-Horizon Persuasion:}} Existing evaluation methods focus on short interactions, not assessing persuasive success over extended conversations. Future benchmarks should track how persuasive influence evolves across multiple turns and sessions, considering factors like memory, trust dynamics, and user resistance over time. \looseness=-1

\textit{\textbf{(2) Persuasive Planning and State Tracking:}} Long-term persuasion requires more than turn-by-turn skill. Models must plan across dialogue horizons, adapt strategies dynamically, and maintain memory of user preferences and prior interactions. Future work should explore methods like hierarchical planning, user modeling, and memory-augmented architectures to support sustained and ethical persuasion.

\vspace{-7pt}
\subsection{AI as Persuadee} 
While recent work has shown that LLMs can be susceptible to persuasive prompts and adversarial strategies \cite{zeng-etal-2024-johnny, xu-etal-2024-earth, bozdag2025persuadecanframeworkevaluating}, the boundaries and underlying mechanisms of these vulnerabilities are not well understood. Insights from steering vectors may further clarify such mechanisms \citep{dunefsky2025oneshot}, and several important directions merit further investigation.

\vspace{-5pt}
\subsubsection{Model Susceptibility to Persuasion.} \textcolor{white}{ } 

\noindent It is not yet clear whether LLMs respond to persuasive strategies in ways analogous to humans. Existing studies have primarily focused on short, single-turn interactions, leaving many open questions about how models internalize influence, especially over longer contexts. Understanding model susceptibility is crucial for both advancing robustness and anticipating new forms of adversarial manipulation. The primary concerns are:

\textit{\textbf{(1) Susceptibility in Long Interactions:}} Most existing work examines persuasion in isolated or a limited number of exchanges, but real-world attacks often unfold across extended dialogues. Research should investigate how LLMs respond when persuasive strategies are diffused over many turns, especially when combined with techniques such as many-shot jailbreaking (MSJ) \citep{anil2024manyshot}. Studying subtle, multi-turn persuasion could reveal deeper model vulnerabilities that are not apparent in short-form settings.

\textit{\textbf{(2) Effective Modalities of Agent Persuasion:}} While natural language is the dominant input modality, it is not yet known whether alternatives such as structured metadata, tool outputs, or interleaved code could more effectively persuade models. Exploring multimodal input may uncover novel vectors of influence and corresponding risks.

\textit{\textbf{(3) Origins of Persuasive Vulnerability:}} It is not yet known which stages of model development (e.g., instruction tuning, reinforcement learning, alignment techniques) contribute most to susceptibility. Future work should aim to isolate these factors to better understand and ultimately protect model responses against manipulative inputs. This behavior might also be examined through models' self-explanations of why they were persuaded, offering insight into which aspects of a persuasive argument influenced the response.

\vspace{-5pt}
\subsubsection{Selective Acceptance and Resistance to Persuasion.} \textcolor{white}{ } 

\noindent Understanding model susceptibility is only the first step; equally important is developing models that can selectively accept or resist persuasion without degrading their general capabilities or instruction-following performance \citep{stengel-eskin-etal-2025-teaching}. Building such systems will require new training objectives, fine-tuning strategies, and evaluation protocols explicitly targeting persuasive resilience. Remaining challenges are:

\textit{\textbf{(1) Defining Selectiveness in Persuasion:}} Persuasion resistance should not emerge from general stubbornness or degraded instruction-following, but from correctly identifying problematic persuasive attempts. We must establish clear definitions and metrics for selective robustness that distinguish between appropriate and inappropriate influence.

\textit{\textbf{(2) Training Selective Persuadees:}} Achieving selective behavior will require novel training techniques, such as fine-tuning on adversarially generated persuasive attacks, training with annotated persuasive feedback, or using preference learning to steer model responses toward robustness while preserving flexibility.

\vspace{-5pt}
\subsubsection{Agent-to-Agent Systems: AI as Persuadee \& Persuader} \textcolor{white}{ }

\noindent Research in computational persuasion has primarily examined human-AI interactions, but as multi-agent systems become increasingly common, novel safety concerns, including agent-to-agent persuasion, are likely to emerge \cite{hammond2025multiagentrisksadvancedai}. Studying persuasion in multi-agent contexts is therefore critical. The main challenges are: 

\textit{\textbf{(1) Persuasion Effects in Multi-Agent Systems:}} Future work must examine how persuasion among agents can influence outcomes in debates, negotiations, and collaborative tasks. It is important to identify when and how more persuasive or more susceptible agents shift system behavior toward undesirable states. Moreover, it remains unclear whether agent-to-agent persuasion will resemble human-like rhetorical strategies through natural language or evolve through novel modalities and mechanisms.

\textit{\textbf{(2) Inter-Agent Dynamics \& Learning Profiles:}} Open questions include whether stronger models (e.g., larger or better-aligned ones) more easily persuade weaker ones, whether weaker agents can effectively persuade stronger ones, and how agents might adapt their trust in others through repeated interaction. Designing safeguards will require careful modeling of agent profiles, trust calibration, and learning dynamics in multi-agent ecosystems.

\vspace{-5pt}
\subsection{AI as Persuasion Judge} Aside from being a Persuader or Persuadee agent, language models also hold great potential to serve as Persuasion Judges where they detect, mitigate, and inform about persuasive attempts, recognize and differentiate harmful persuasion from beneficial encouragements.

\vspace{-5pt}
\subsubsection{Detecting \& Evaluating Persuasion} \textcolor{white}{ }

\noindent Automatically identifying patterns of persuasion—such as rhetorical strategies, emotional appeals, or attempts at undue influence—is a critical use case for both classifiers and generative language models. As reviewed in \S~\ref{evaluating-persuasion}, prior work has explored modeling persuasion and distinguishing successful from unsuccessful persuasive attempts. However, these systems often lack alignment with human preferences or generalize poorly beyond the specific domains they were trained on. Improving detection and evaluation models remains a challenge, with several open research directions:

\textit{\textbf{(1) Creating Reliable Persuasion Datasets: }} Persuasiveness is highly subjective, and even human-annotated datasets often exhibit noise, low annotator agreement, and limited generalizability. Progress in persuasion detection and evaluation will require the collection of more natural, diverse, and high-agreement datasets that better reflect real-world perceptions of persuasive effectiveness and ethical acceptability.

\textit{\textbf{(2) Robust and Aligned Persuasion Detection:}} Current approaches, including prompting and fine-tuning LLMs, still fail to reliably assess the quality or ethical appropriateness of persuasive content. Beyond improving dataset quality, future work must develop more robust detectors that distinguish beneficial from harmful persuasion without introducing systematic biases, ideally grounded in clearer normative frameworks.

\textit{\textbf{(3) Predicting User Responses to Persuasion:}} 
Models should aim to predict how users are likely to respond to persuasive messages. Anticipating user reactions would enable more adaptive and context-sensitive evaluations, supporting downstream applications such as safer generation, persuasion mitigation, and user-personalized feedback.

\vspace{-5pt}
\subsubsection{Identifying Long-Term Persuasion.} \textcolor{white}{ }

\noindent Detecting persuasion across long-term interactions presents distinct challenges beyond single-turn analysis. Rather than simply recognizing persuasive intent in isolated exchanges, judge models must discern how influence evolves gradually through accumulated trust, personalization, and subtle framing. As conversational agents retain user-specific memories and adapt, identifying when a conversation shifts from rapport-building to strategic persuasion becomes increasingly complex. Main challenges in this direction are:

\textit{\textbf{(1) Detecting Agendas and Gradual Steering:}} Exploring whether models can steer users over time while masking persuasive intent through a long history of benign interaction is crucial. New methods are needed to identify early warning signals, such as subtle shifts in tone, strategy, or goal alignment, that could indicate emerging manipulation in extended conversations.

\textit{\textbf{(2) Long-Term Monitoring of Persuasion:}} LLMs themselves could be leveraged to monitor for signs of manipulation in long-context conversations. Models should be trained to self-detect persuasive shifts, issue warnings, or audit past interactions to surface latent persuasive strategies, supporting the development of more trustworthy AI systems.

\vspace{-5pt}
\subsubsection{Generative Adversarial Persuasion: AI as Persuader, Persuadee, and Judge.} \textcolor{white}{ }

\noindent In the preceding sections, we highlighted current limitations in the roles of AI as Persuader, Persuadee, and Persuasion Judge. We believe that progress in these areas can be achieved collectively through a unified framework called: 

\textit{\textbf{(1) Generative Adversarial Persuasion:}} In this setup, a persuasive agent attempts to influence a persuadee model, while a judge model oversees the interaction and evaluates the effectiveness, appropriateness, and potential risks of the persuasive attempt. Drawing inspiration from the structure of Generative Adversarial Networks (GANs), this framework encourages co-evolution among the agents. The persuadee learns to develop resistance to manipulative or unethical persuasion, the persuader learns to improve persuasiveness in its responses, and the judge becomes more accurate in identifying persuasive strategies and assessing their quality. This adversarial multi-agent paradigm offers a promising direction for building more robust, adaptive, and ethically grounded persuasive systems.
\vspace{-10pt}
\section{CONCLUSION} \label{sec:conclusion}
In this survey of AI-driven persuasion, we have systematically examined existing research through the lens of \textbf{AI as Persuader}, \textbf{AI as Persuadee}, and \textbf{AI as Persuasion Judge}, with a focus on three core areas: \textit{Evaluating Persuasion}, \textit{Generating Persuasion}, and \textit{Safeguarding Persuasion}. As persuasion becomes an increasingly inherent capability of large language models, research must prioritize developing robust methods for evaluating persuasive behavior, responsibly generating persuasive content, and carefully safeguarding against the potential risks posed by advanced AI systems. Our survey highlights that persuasion in AI systems poses challenges not only for human users but also within agentic ecosystems where models interact and influence one another. Looking ahead, we have outlined several promising future directions in this rapidly growing area of research, including improved and unified evaluation mechanisms, adaptive persuasive systems, and techniques for selective acceptance of persuasion. While our survey focuses primarily on English text-based persuasion, we acknowledge the importance of expanding this work to cover multilingual and multimodal contexts, which remain underexplored. Moreover, a richer interdisciplinary integration with fields such as psycholinguistics, behavioral psychology, and communication studies can offer valuable insights into phenomena like argument framing, behavior change, and language processing, thereby informing the design, evaluation, and ethical grounding of persuasive AI systems. We call on the research community to explore these diverse dimensions of persuasion to build safer, more ethical, and more trustworthy AI systems.

\vspace{-10pt}
\bibliographystyle{ACM-Reference-Format}
\bibliography{main}

@inproceedings{wang-etal-2019-persuasion,
    title = "Persuasion for Good: Towards a Personalized Persuasive Dialogue System for Social Good",
    author = "Wang, Xuewei  and
      Shi, Weiyan  and
      Kim, Richard  and
      Oh, Yoojung  and
      Yang, Sijia  and
      Zhang, Jingwen  and
      Yu, Zhou",
    editor = "Korhonen, Anna  and
      Traum, David  and
      M{\`a}rquez, Llu{\'\i}s",
    booktitle = "Proceedings of the 57th Annual Meeting of the Association for Computational Linguistics",
    month = jul,
    year = "2019",
    address = "Florence, Italy",
    publisher = "Association for Computational Linguistics",
    url = "https://aclanthology.org/P19-1566",
    doi = "10.18653/v1/P19-1566",
    pages = "5635--5649",
}

@inproceedings{computationalpropagandasurvey,
author = {Da San Martino, Giovanni and Cresci, Stefano and Barr\'{o}n-Cede\~{n}o, Alberto and Yu, Seunghak and Di Pietro, Roberto and Nakov, Preslav},
title = {A survey on computational propaganda detection},
year = {2021},
isbn = {9780999241165},
booktitle = {Proceedings of the Twenty-Ninth International Joint Conference on Artificial Intelligence},
articleno = {672},
numpages = {7},
location = {Yokohama, Yokohama, Japan},
series = {IJCAI'20}
}

@article{Hidey_McKeown_2018, title={Persuasive Influence Detection: The Role of Argument Sequencing}, volume={32}, url={https://ojs.aaai.org/index.php/AAAI/article/view/12003}, DOI={10.1609/aaai.v32i1.12003}, abstractNote={ &lt;p&gt; &lt;!-- p.p1 {margin: 0.0px 0.0px 0.0px 0.0px; font: 9.0px Helvetica} --&gt; &lt;p class=&quot;p1&quot;&gt;Automatic detection of persuasion in online discussion is key to understanding how social media is used. Predicting persuasiveness is difficult, however, due to the need to model world knowledge, dialogue, and sequential reasoning. We focus on modeling the sequence of arguments in social media posts using neural models with embeddings for words, discourse relations, and semantic frames. We demonstrate significant improvement over prior work in detecting successful arguments. We also present an error analysis assessing novice human performance at predicting persuasiveness.&lt;/p&gt; &lt;/p&gt; }, number={1}, journal={Proceedings of the AAAI Conference on Artificial Intelligence}, author={Hidey, Christopher and McKeown, Kathleen}, year={2018}, month={Apr.} }

@book{gass2022persuasion,
  title={Persuasion: Social Influence and Compliance Gaining},
  author={Gass, R.H. and Seiter, J.S.},
  isbn={9781000556773},
  url={https://books.google.com/books?id=leFeEAAAQBAJ},
  year={2022},
  publisher={Taylor \& Francis}
}

@online{durmus2024persuasion,
author = {Esin Durmus and Liane Lovitt and Alex Tamkin and Stuart Ritchie and Jack Clark and Deep Ganguli},
title = {Measuring the Persuasiveness of Language Models},
date = {2024-04-09},
year = {2024},
url = {https://www.anthropic.com/news/measuring-model-persuasiveness},
}

@article{ai-pro-vaccine-karinshak-2023,
author = {Karinshak, Elise and Liu, Sunny Xun and Park, Joon Sung and Hancock, Jeffrey T.},
title = {Working With AI to Persuade: Examining a Large Language Model's Ability to Generate Pro-Vaccination Messages},
year = {2023},
issue_date = {April 2023},
publisher = {Association for Computing Machinery},
address = {New York, NY, USA},
volume = {7},
number = {CSCW1},
url = {https://doi.org/10.1145/3579592},
doi = {10.1145/3579592},
journal = {Proc. ACM Hum.-Comput. Interact.},
month = apr,
articleno = {116},
numpages = {29}
}

@inproceedings{shmueli2019detecting,
  title={Detecting persuasive arguments based on author-reader personality traits and their interaction},
  author={Shmueli-Scheuer, Michal and Herzig, Jonathan and Konopnicki, David and Sandbank, Tommy},
  booktitle={Proceedings of the 27th ACM conference on user modeling, adaptation and personalization},
  pages={211--215},
  year={2019}
}

@online{chatgpt,
    author={OpenAI} ,
    title={Introducing ChatGPT},
    year={2022},
    url={https://openai.com/index/chatgpt},
    note={Accessed: 21 Oct. 2024}
}

@article{tsinganos2022utilizing,
  title={Utilizing convolutional neural networks and word embeddings for early-stage recognition of persuasion in chat-based social engineering attacks},
  author={Tsinganos, Nikolaos and Mavridis, Ioannis and Gritzalis, Dimitris},
  journal={IEEE Access},
  volume={10},
  pages={108517--108529},
  year={2022},
  publisher={IEEE}
}

@inproceedings{winning-args-tan-etal-2016,
author = {Tan, Chenhao and Niculae, Vlad and Danescu-Niculescu-Mizil, Cristian and Lee, Lillian},
title = {Winning Arguments: Interaction Dynamics and Persuasion Strategies in Good-faith Online Discussions},
year = {2016},
isbn = {9781450341431},
publisher = {International World Wide Web Conferences Steering Committee},
address = {Republic and Canton of Geneva, CHE},
url = {https://doi.org/10.1145/2872427.2883081},
doi = {10.1145/2872427.2883081},
booktitle = {Proceedings of the 25th International Conference on World Wide Web},
pages = {613–624},
numpages = {12},
location = {Montr\'{e}al, Qu\'{e}bec, Canada},
series = {WWW '16}
}

@inproceedings{chawla2021casino,
  title={CaSiNo: A Corpus of Campsite Negotiation Dialogues for Automatic Negotiation Systems},
  author={Chawla, Kushal and Ramirez, Jaysa and Clever, Rene and Lucas, Gale and May, Jonathan and Gratch, Jonathan},
  booktitle={Proceedings of the 2021 Conference of the North American Chapter of the Association for Computational Linguistics: Human Language Technologies},
  pages={3167--3185},
  year={2021}
}

@inproceedings{jin-etal-2024-persuading,
    title = "Persuading across Diverse Domains: a Dataset and Persuasion Large Language Model",
    author = "Jin, Chuhao  and
      Ren, Kening  and
      Kong, Lingzhen  and
      Wang, Xiting  and
      Song, Ruihua  and
      Chen, Huan",
    editor = "Ku, Lun-Wei  and
      Martins, Andre  and
      Srikumar, Vivek",
    booktitle = "Proceedings of the 62nd Annual Meeting of the Association for Computational Linguistics (Volume 1: Long Papers)",
    month = aug,
    year = "2024",
    address = "Bangkok, Thailand",
    publisher = "Association for Computational Linguistics",
    url = "https://aclanthology.org/2024.acl-long.92",
    doi = "10.18653/v1/2024.acl-long.92",
    pages = "1678--1706",
}

@inproceedings{xu-etal-2024-earth,
    title = "The Earth is Flat because...: Investigating {LLM}s{'} Belief towards Misinformation via Persuasive Conversation",
    author = "Xu, Rongwu  and
      Lin, Brian  and
      Yang, Shujian  and
      Zhang, Tianqi  and
      Shi, Weiyan  and
      Zhang, Tianwei  and
      Fang, Zhixuan  and
      Xu, Wei  and
      Qiu, Han",
    editor = "Ku, Lun-Wei  and
      Martins, Andre  and
      Srikumar, Vivek",
    booktitle = "Proceedings of the 62nd Annual Meeting of the Association for Computational Linguistics (Volume 1: Long Papers)",
    month = aug,
    year = "2024",
    address = "Bangkok, Thailand",
    publisher = "Association for Computational Linguistics",
    url = "https://aclanthology.org/2024.acl-long.858",
    doi = "10.18653/v1/2024.acl-long.858",
    pages = "16259--16303",
}

@inproceedings{lewis-etal-2017-deal,
    title = "Deal or No Deal? End-to-End Learning of Negotiation Dialogues",
    author = "Lewis, Mike  and
      Yarats, Denis  and
      Dauphin, Yann  and
      Parikh, Devi  and
      Batra, Dhruv",
    editor = "Palmer, Martha  and
      Hwa, Rebecca  and
      Riedel, Sebastian",
    booktitle = "Proceedings of the 2017 Conference on Empirical Methods in Natural Language Processing",
    month = sep,
    year = "2017",
    address = "Copenhagen, Denmark",
    publisher = "Association for Computational Linguistics",
    url = "https://aclanthology.org/D17-1259",
    doi = "10.18653/v1/D17-1259",
    pages = "2443--2453",
}

@inproceedings{dimitrov-etal-2021-semeval,
    title = "{S}em{E}val-2021 Task 6: Detection of Persuasion Techniques in Texts and Images",
    author = "Dimitrov, Dimitar  and
      Bin Ali, Bishr  and
      Shaar, Shaden  and
      Alam, Firoj  and
      Silvestri, Fabrizio  and
      Firooz, Hamed  and
      Nakov, Preslav  and
      Da San Martino, Giovanni",
    editor = "Palmer, Alexis  and
      Schneider, Nathan  and
      Schluter, Natalie  and
      Emerson, Guy  and
      Herbelot, Aurelie  and
      Zhu, Xiaodan",
    booktitle = "Proceedings of the 15th International Workshop on Semantic Evaluation (SemEval-2021)",
    month = aug,
    year = "2021",
    address = "Online",
    publisher = "Association for Computational Linguistics",
    url = "https://aclanthology.org/2021.semeval-1.7",
    doi = "10.18653/v1/2021.semeval-1.7",
    pages = "70--98",
}

@inproceedings{sakurai-miyao-2024-evaluating,
    title = "Evaluating Intention Detection Capability of Large Language Models in Persuasive Dialogues",
    author = "Sakurai, Hiromasa  and
      Miyao, Yusuke",
    editor = "Ku, Lun-Wei  and
      Martins, Andre  and
      Srikumar, Vivek",
    booktitle = "Proceedings of the 62nd Annual Meeting of the Association for Computational Linguistics (Volume 1: Long Papers)",
    month = aug,
    year = "2024",
    address = "Bangkok, Thailand",
    publisher = "Association for Computational Linguistics",
    url = "https://aclanthology.org/2024.acl-long.90",
    doi = "10.18653/v1/2024.acl-long.90",
    pages = "1635--1657",
}

@inproceedings{dutt-etal-2020-keeping,
    title = "Keeping Up Appearances: Computational Modeling of Face Acts in Persuasion Oriented Discussions",
    author = "Dutt, Ritam  and
      Joshi, Rishabh  and
      Rose, Carolyn",
    editor = "Webber, Bonnie  and
      Cohn, Trevor  and
      He, Yulan  and
      Liu, Yang",
    booktitle = "Proceedings of the 2020 Conference on Empirical Methods in Natural Language Processing (EMNLP)",
    month = nov,
    year = "2020",
    address = "Online",
    publisher = "Association for Computational Linguistics",
    url = "https://aclanthology.org/2020.emnlp-main.605",
    doi = "10.18653/v1/2020.emnlp-main.605",
    pages = "7473--7485",
}

@inproceedings{zeng-etal-2024-johnny,
    title = "How Johnny Can Persuade {LLM}s to Jailbreak Them: Rethinking Persuasion to Challenge {AI} Safety by Humanizing {LLM}s",
    author = "Zeng, Yi  and
      Lin, Hongpeng  and
      Zhang, Jingwen  and
      Yang, Diyi  and
      Jia, Ruoxi  and
      Shi, Weiyan",
    editor = "Ku, Lun-Wei  and
      Martins, Andre  and
      Srikumar, Vivek",
    booktitle = "Proceedings of the 62nd Annual Meeting of the Association for Computational Linguistics (Volume 1: Long Papers)",
    month = aug,
    year = "2024",
    address = "Bangkok, Thailand",
    publisher = "Association for Computational Linguistics",
    url = "https://aclanthology.org/2024.acl-long.773",
    doi = "10.18653/v1/2024.acl-long.773",
    pages = "14322--14350",
}

@misc{pöyhönen2022multilingualpersuasiondetectionvideo,
      title={Multilingual Persuasion Detection: Video Games as an Invaluable Data Source for NLP}, 
      author={Teemu Pöyhönen and Mika Hämäläinen and Khalid Alnajjar},
      year={2022},
      eprint={2207.04453},
      archivePrefix={arXiv},
      primaryClass={cs.CL},
      url={https://arxiv.org/abs/2207.04453}, 
}

@article{Chen_Yang_2021, 
    title={Weakly-Supervised Hierarchical Models for Predicting Persuasive Strategies in Good-faith Textual Requests}, 
    volume={35}, 
    url={https://ojs.aaai.org/index.php/AAAI/article/view/17498}, 
    DOI={10.1609/aaai.v35i14.17498}, 
    abstractNote={Modeling persuasive language has the potential to better facilitate our decision-making processes. Despite its importance, computational modeling of persuasion is still in its infancy, largely due to the lack of benchmark datasets that can provide quantitative labels of persuasive strategies to expedite this line of research. To this end, we introduce a large-scale multi-domain text corpus for modeling persuasive strategies in good-faith text requests. Moreover, we design a hierarchical weakly-supervised latent variable model that can leverage partially labeled data to predict such associated persuasive strategies for each sentence, where the supervision comes from both the overall document-level labels and very limited sentence-level labels. Experimental results showed that our proposed method outperformed existing semi-supervised baselines significantly. We have publicly released our code at https://github.com/GT-SALT/Persuasion_Strategy_WVAE.}, 
    number={14}, journal={Proceedings of the AAAI Conference on Artificial Intelligence}, 
    author={Chen, Jiaao and Yang, Diyi}, 
    year={2021}, 
    month={May}, 
    pages={12648-12656} 
}

@inproceedings{effects_of_persuasive_dialogue_shi20,
author = {Shi, Weiyan and Wang, Xuewei and Oh, Yoo Jung and Zhang, Jingwen and Sahay, Saurav and Yu, Zhou},
title = {Effects of Persuasive Dialogues: Testing Bot Identities and Inquiry Strategies},
year = {2020},
isbn = {9781450367080},
publisher = {Association for Computing Machinery},
address = {New York, NY, USA},
url = {https://doi.org/10.1145/3313831.3376843},
doi = {10.1145/3313831.3376843},
booktitle = {Proceedings of the 2020 CHI Conference on Human Factors in Computing Systems},
pages = {1–13},
numpages = {13},
location = {Honolulu, HI, USA},
series = {CHI '20}
}

@inproceedings{samad-etal-2022-empathetic,
    title = "Empathetic Persuasion: Reinforcing Empathy and Persuasiveness in Dialogue Systems",
    author = "Samad, Azlaan Mustafa  and
      Mishra, Kshitij  and
      Firdaus, Mauajama  and
      Ekbal, Asif",
    editor = "Carpuat, Marine  and
      de Marneffe, Marie-Catherine  and
      Meza Ruiz, Ivan Vladimir",
    booktitle = "Findings of the Association for Computational Linguistics: NAACL 2022",
    month = jul,
    year = "2022",
    address = "Seattle, United States",
    publisher = "Association for Computational Linguistics",
    url = "https://aclanthology.org/2022.findings-naacl.63",
    doi = "10.18653/v1/2022.findings-naacl.63",
    pages = "844--856",
}

@article{bianchi2024llms,
      title={How Well Can LLMs Negotiate? NegotiationArena Platform and Analysis}, 
      author={Federico Bianchi and Patrick John Chia and Mert Yuksekgonul and Jacopo Tagliabue and Dan Jurafsky and James Zou},
      year={2024},
      eprint={2402.05863},
      journal={arXiv},
}

@misc{li2024llmdefensesrobustmultiturn,
      title={LLM Defenses Are Not Robust to Multi-Turn Human Jailbreaks Yet}, 
      author={Nathaniel Li and Ziwen Han and Ian Steneker and Willow Primack and Riley Goodside and Hugh Zhang and Zifan Wang and Cristina Menghini and Summer Yue},
      year={2024},
      eprint={2408.15221},
      archivePrefix={arXiv},
      primaryClass={cs.LG},
      url={https://arxiv.org/abs/2408.15221}, 
}

@misc{michael2023debatehelpssuperviseunreliable,
      title={Debate Helps Supervise Unreliable Experts}, 
      author={Julian Michael and Salsabila Mahdi and David Rein and Jackson Petty and Julien Dirani and Vishakh Padmakumar and Samuel R. Bowman},
      year={2023},
      eprint={2311.08702},
      archivePrefix={arXiv},
      primaryClass={cs.AI},
      url={https://arxiv.org/abs/2311.08702}, 
}

@inproceedings{
du2024improving,
title={Improving Factuality and Reasoning in Language Models through Multiagent Debate},
author={Yilun Du and Shuang Li and Antonio Torralba and Joshua B. Tenenbaum and Igor Mordatch},
booktitle={Forty-first International Conference on Machine Learning},
year={2024},
url={https://openreview.net/forum?id=zj7YuTE4t8}
}

@article{CHEN202147,
title = {Persuasive dialogue understanding: The baselines and negative results},
journal = {Neurocomputing},
volume = {431},
pages = {47-56},
year = {2021},
issn = {0925-2312},
doi = {https://doi.org/10.1016/j.neucom.2020.11.040},
url = {https://www.sciencedirect.com/science/article/pii/S0925231220318336},
author = {Hui Chen and Deepanway Ghosal and Navonil Majumder and Amir Hussain and Soujanya Poria}
}

@inproceedings{da-san-martino-etal-2020-semeval,
    title = "{S}em{E}val-2020 Task 11: Detection of Propaganda Techniques in News Articles",
    author = "Da San Martino, Giovanni  and
      Barr{\'o}n-Cede{\~n}o, Alberto  and
      Wachsmuth, Henning  and
      Petrov, Rostislav  and
      Nakov, Preslav",
    editor = "Herbelot, Aurelie  and
      Zhu, Xiaodan  and
      Palmer, Alexis  and
      Schneider, Nathan  and
      May, Jonathan  and
      Shutova, Ekaterina",
    booktitle = "Proceedings of the Fourteenth Workshop on Semantic Evaluation",
    month = dec,
    year = "2020",
    address = "Barcelona (online)",
    publisher = "International Committee for Computational Linguistics",
    url = "https://aclanthology.org/2020.semeval-1.186",
    doi = "10.18653/v1/2020.semeval-1.186",
    pages = "1377--1414",
}

@inproceedings{dimitrov-etal-2024-semeval,
    title = "{S}em{E}val-2024 Task 4: Multilingual Detection of Persuasion Techniques in Memes",
    author = "Dimitrov, Dimitar  and
      Alam, Firoj  and
      Hasanain, Maram  and
      Hasnat, Abul  and
      Silvestri, Fabrizio  and
      Nakov, Preslav  and
      Da San Martino, Giovanni",
    editor = {Ojha, Atul Kr.  and
      Do{\u{g}}ru{\"o}z, A. Seza  and
      Tayyar Madabushi, Harish  and
      Da San Martino, Giovanni  and
      Rosenthal, Sara  and
      Ros{\'a}, Aiala},
    booktitle = "Proceedings of the 18th International Workshop on Semantic Evaluation (SemEval-2024)",
    month = jun,
    year = "2024",
    address = "Mexico City, Mexico",
    publisher = "Association for Computational Linguistics",
    url = "https://aclanthology.org/2024.semeval-1.275",
    doi = "10.18653/v1/2024.semeval-1.275",
    pages = "2009--2026",
}

@inproceedings{guerini-etal-2015-echoes,
    title = "Echoes of Persuasion: The Effect of Euphony in Persuasive Communication",
    author = {Guerini, Marco  and
      {\"O}zbal, G{\"o}zde  and
      Strapparava, Carlo},
    editor = "Mihalcea, Rada  and
      Chai, Joyce  and
      Sarkar, Anoop",
    booktitle = "Proceedings of the 2015 Conference of the North {A}merican Chapter of the Association for Computational Linguistics: Human Language Technologies",
    month = may # "{--}" # jun,
    year = "2015",
    address = "Denver, Colorado",
    publisher = "Association for Computational Linguistics",
    url = "https://aclanthology.org/N15-1172",
    doi = "10.3115/v1/N15-1172",
    pages = "1483--1493",
}

@misc{o1systemcard2024,
  title        = {OpenAI o1 System Card},
  author       = {OpenAI},
  year         = {2024},
  month        = {December},
  url          = {https://cdn.openai.com/o1-system-card-20241205.pdf},
  note         = {Accessed: 2024-12-10}
}

@misc{gpt45systemcard2025,
  title        = {OpenAI GPT-4.5 System Card},
  author       = {OpenAI},
  year         = {2025},
  month        = {February},
  url          = {https://cdn.openai.com/gpt-4-5-system-card-2272025.pdf},
  note         = {Accessed: 2025-03-06}
}

@misc{bozdag2025persuadecanframeworkevaluating,
      title={Persuade Me if You Can: A Framework for Evaluating Persuasion Effectiveness and Susceptibility Among Large Language Models}, 
      author={Nimet Beyza Bozdag and Shuhaib Mehri and Gokhan Tur and Dilek Hakkani-Tür},
      year={2025},
      eprint={2503.01829},
      archivePrefix={arXiv},
      primaryClass={cs.CL},
      url={https://arxiv.org/abs/2503.01829}, 
}

@misc{singh2024measuringimprovingpersuasivenesslarge,
      title={Measuring and Improving Persuasiveness of Large Language Models}, 
      author={Somesh Singh and Yaman K Singla and Harini SI and Balaji Krishnamurthy},
      year={2024},
      eprint={2410.02653},
      archivePrefix={arXiv},
      primaryClass={cs.CL},
      url={https://arxiv.org/abs/2410.02653}, 
}

@misc{phuong2024evaluatingfrontiermodelsdangerous,
      title={Evaluating Frontier Models for Dangerous Capabilities}, 
      author={Mary Phuong and Matthew Aitchison and Elliot Catt and Sarah Cogan and Alexandre Kaskasoli and Victoria Krakovna and David Lindner and Matthew Rahtz and Yannis Assael and Sarah Hodkinson and Heidi Howard and Tom Lieberum and Ramana Kumar and Maria Abi Raad and Albert Webson and Lewis Ho and Sharon Lin and Sebastian Farquhar and Marcus Hutter and Gregoire Deletang and Anian Ruoss and Seliem El-Sayed and Sasha Brown and Anca Dragan and Rohin Shah and Allan Dafoe and Toby Shevlane},
      year={2024},
      eprint={2403.13793},
      archivePrefix={arXiv},
      primaryClass={cs.LG},
      url={https://arxiv.org/abs/2403.13793}, 
}

@inproceedings{singh2023exploiting,
  title={Exploiting large language models (llms) through deception techniques and persuasion principles},
  author={Singh, Sonali and Abri, Faranak and Namin, Akbar Siami},
  booktitle={2023 IEEE International Conference on Big Data (BigData)},
  pages={2508--2517},
  year={2023},
  organization={IEEE}
}

@article{hasanain2024can,
  title={Can gpt-4 identify propaganda? annotation and detection of propaganda spans in news articles},
  author={Hasanain, Maram and Ahmed, Fatema and Alam, Firoj},
  journal={arXiv preprint arXiv:2402.17478},
  year={2024}
}

@article{dutt2021resper,
  title={Resper: Computationally modelling resisting strategies in persuasive conversations},
  author={Dutt, Ritam and Sinha, Sayan and Joshi, Rishabh and Chakraborty, Surya Shekhar and Riggs, Meredith and Yan, Xinru and Bao, Haogang and Ros{\'e}, Carolyn Penstein},
  journal={arXiv preprint arXiv:2101.10545},
  year={2021}
}

@article{burtell2023artificial,
  title={Artificial influence: An analysis of AI-driven persuasion},
  author={Burtell, Matthew and Woodside, Thomas},
  journal={arXiv preprint arXiv:2303.08721},
  year={2023}
}

@misc{elsayed2024mechanismbasedapproachmitigatingharms,
      title={A Mechanism-Based Approach to Mitigating Harms from Persuasive Generative AI}, 
      author={Seliem El-Sayed and Canfer Akbulut and Amanda McCroskery and Geoff Keeling and Zachary Kenton and Zaria Jalan and Nahema Marchal and Arianna Manzini and Toby Shevlane and Shannon Vallor and Daniel Susser and Matija Franklin and Sophie Bridgers and Harry Law and Matthew Rahtz and Murray Shanahan and Michael Henry Tessler and Arthur Douillard and Tom Everitt and Sasha Brown},
      year={2024},
      eprint={2404.15058},
      archivePrefix={arXiv},
      primaryClass={cs.CY},
      url={https://arxiv.org/abs/2404.15058}, 
}

@article{saenger2024autopersuade,
  title={AutoPersuade: A Framework for Evaluating and Explaining Persuasive Arguments},
  author={Saenger, Till Raphael and Hinck, Musashi and Grimmer, Justin and Stewart, Brandon M},
  journal={arXiv preprint arXiv:2410.08917},
  year={2024}
}

@inproceedings{lukin-etal-2017-argument,
    title = "Argument Strength is in the Eye of the Beholder: Audience Effects in Persuasion",
    author = "Lukin, Stephanie  and
      Anand, Pranav  and
      Walker, Marilyn  and
      Whittaker, Steve",
    editor = "Lapata, Mirella  and
      Blunsom, Phil  and
      Koller, Alexander",
    booktitle = "Proceedings of the 15th Conference of the {E}uropean Chapter of the Association for Computational Linguistics: Volume 1, Long Papers",
    month = apr,
    year = "2017",
    address = "Valencia, Spain",
    publisher = "Association for Computational Linguistics",
    url = "https://aclanthology.org/E17-1070/",
    pages = "742--753"
}

@article{matz2024potential,
  title={The potential of generative AI for personalized persuasion at scale},
  author={Matz, Sandra C and Teeny, Jacob D and Vaid, Sumer S and Peters, Heinrich and Harari, Gabriella M and Cerf, Moran},
  journal={Scientific Reports},
  volume={14},
  number={1},
  pages={4692},
  year={2024},
  publisher={Nature Publishing Group UK London}
}

@article{simchon2024persuasive,
  title={The persuasive effects of political microtargeting in the age of generative artificial intelligence},
  author={Simchon, Almog and Edwards, Matthew and Lewandowsky, Stephan},
  journal={PNAS nexus},
  volume={3},
  number={2},
  pages={pgae035},
  year={2024},
  publisher={Oxford University Press US}
}

@article{kaptein2015personalizing,
  title={Personalizing persuasive technologies: Explicit and implicit personalization using persuasion profiles},
  author={Kaptein, Maurits and Markopoulos, Panos and De Ruyter, Boris and Aarts, Emile},
  journal={International Journal of Human-Computer Studies},
  volume={77},
  pages={38--51},
  year={2015},
  publisher={Elsevier}
}

@article{convperscontrolledtrial,
  publtype={informal},
  author={Francesco Salvi and Manoel Horta Ribeiro and Riccardo Gallotti and Robert West},
  title={On the Conversational Persuasiveness of Large Language Models: A Randomized Controlled Trial},
  year={2024},
  cdate={1704067200000},
  journal={CoRR},
  volume={abs/2403.14380},
  url={https://doi.org/10.48550/arXiv.2403.14380}
}

@article{TIWARI2022116303,
title = {A persona aware persuasive dialogue policy for dynamic and co-operative goal setting},
journal = {Expert Systems with Applications},
volume = {195},
pages = {116303},
year = {2022},
issn = {0957-4174},
doi = {https://doi.org/10.1016/j.eswa.2021.116303},
url = {https://www.sciencedirect.com/science/article/pii/S0957417421016067},
author = {Abhisek Tiwari and Tulika Saha and Sriparna Saha and Shubhashis Sengupta and Anutosh Maitra and Roshni Ramnani and Pushpak Bhattacharyya}
}

@article{ruiz2024persuasion,
  title={Persuasion-enhanced computational argumentative reasoning through argumentation-based persuasive frameworks},
  author={Ruiz-Dolz, Ramon and Taverner, Joaquin and Heras Barber{\'a}, Stella M and Garc{\'\i}a-Fornes, Ana},
  journal={User Modeling and User-Adapted Interaction},
  volume={34},
  number={1},
  pages={229--258},
  year={2024},
  publisher={Springer}
}

@inproceedings{meguellati2024good,
  title={How good are LLMS in generating personalized advertisements?},
  author={Meguellati, Elyas and Han, Lei and Bernstein, Abraham and Sadiq, Shazia and Demartini, Gianluca},
  booktitle={Companion Proceedings of the ACM Web Conference 2024},
  pages={826--829},
  year={2024}
}

@inproceedings{tiwari-etal-2022-persona,
    title = "Persona or Context? Towards Building Context adaptive Personalized Persuasive Virtual Sales Assistant",
    author = "Tiwari, Abhisek  and
      Saha, Sriparna  and
      Sengupta, Shubhashis  and
      Maitra, Anutosh  and
      Ramnani, Roshni  and
      Bhattacharyya, Pushpak",
    editor = "He, Yulan  and
      Ji, Heng  and
      Li, Sujian  and
      Liu, Yang  and
      Chang, Chua-Hui",
    booktitle = "Proceedings of the 2nd Conference of the Asia-Pacific Chapter of the Association for Computational Linguistics and the 12th International Joint Conference on Natural Language Processing (Volume 1: Long Papers)",
    month = nov,
    year = "2022",
    address = "Online only",
    publisher = "Association for Computational Linguistics",
    url = "https://aclanthology.org/2022.aacl-main.76/",
    doi = "10.18653/v1/2022.aacl-main.76",
    pages = "1035--1047"
}

@article{tiwari2023towards,
  title={Towards personalized persuasive dialogue generation for adversarial task oriented dialogue setting},
  author={Tiwari, Abhisek and Khandwe, Abhijeet and Saha, Sriparna and Ramnani, Roshni and Maitra, Anutosh and Sengupta, Shubhashis},
  journal={Expert Systems with Applications},
  volume={213},
  pages={118775},
  year={2023},
  publisher={Elsevier}
}

@inproceedings{habernal-gurevych-2016-argument,
    title = "Which argument is more convincing? Analyzing and predicting convincingness of Web arguments using bidirectional {LSTM}",
    author = "Habernal, Ivan  and
      Gurevych, Iryna",
    editor = "Erk, Katrin  and
      Smith, Noah A.",
    booktitle = "Proceedings of the 54th Annual Meeting of the Association for Computational Linguistics (Volume 1: Long Papers)",
    month = aug,
    year = "2016",
    address = "Berlin, Germany",
    publisher = "Association for Computational Linguistics",
    url = "https://aclanthology.org/P16-1150/",
    doi = "10.18653/v1/P16-1150",
    pages = "1589--1599"
}

@article{simpson-gurevych-2018-finding,
    title = "Finding Convincing Arguments Using Scalable {B}ayesian Preference Learning",
    author = "Simpson, Edwin  and
      Gurevych, Iryna",
    editor = "Lee, Lillian  and
      Johnson, Mark  and
      Toutanova, Kristina  and
      Roark, Brian",
    journal = "Transactions of the Association for Computational Linguistics",
    volume = "6",
    year = "2018",
    address = "Cambridge, MA",
    publisher = "MIT Press",
    url = "https://aclanthology.org/Q18-1026/",
    doi = "10.1162/tacl_a_00026",
    pages = "357--371"
}

@inproceedings{toledo-etal-2019-automatic,
    title = "Automatic Argument Quality Assessment - New Datasets and Methods",
    author = "Toledo, Assaf  and
      Gretz, Shai  and
      Cohen-Karlik, Edo  and
      Friedman, Roni  and
      Venezian, Elad  and
      Lahav, Dan  and
      Jacovi, Michal  and
      Aharonov, Ranit  and
      Slonim, Noam",
    editor = "Inui, Kentaro  and
      Jiang, Jing  and
      Ng, Vincent  and
      Wan, Xiaojun",
    booktitle = "Proceedings of the 2019 Conference on Empirical Methods in Natural Language Processing and the 9th International Joint Conference on Natural Language Processing (EMNLP-IJCNLP)",
    month = nov,
    year = "2019",
    address = "Hong Kong, China",
    publisher = "Association for Computational Linguistics",
    url = "https://aclanthology.org/D19-1564/",
    doi = "10.18653/v1/D19-1564",
    pages = "5625--5635"
}

@misc{pauli2025measuringbenchmarkinglargelanguage,
      title={Measuring and Benchmarking Large Language Models' Capabilities to Generate Persuasive Language}, 
      author={Amalie Brogaard Pauli and Isabelle Augenstein and Ira Assent},
      year={2025},
      eprint={2406.17753},
      archivePrefix={arXiv},
      primaryClass={cs.CL},
      url={https://arxiv.org/abs/2406.17753}, 
}

@inproceedings{rescala2024languagemodelsrecognizeconvincing,
    title = "Can Language Models Recognize Convincing Arguments?",
    author = "Rescala, Paula  and
      Ribeiro, Manoel Horta  and
      Hu, Tiancheng  and
      West, Robert",
    editor = "Al-Onaizan, Yaser  and
      Bansal, Mohit  and
      Chen, Yun-Nung",
    booktitle = "Findings of the Association for Computational Linguistics: EMNLP 2024",
    month = nov,
    year = "2024",
    address = "Miami, Florida, USA",
    publisher = "Association for Computational Linguistics",
    url = "https://aclanthology.org/2024.findings-emnlp.515/",
    doi = "10.18653/v1/2024.findings-emnlp.515",
    pages = "8826--8837"
}

@article{reducingconspiracy,
    author = {Thomas H. Costello  and Gordon Pennycook  and David G. Rand },
    title = {Durably reducing conspiracy beliefs through dialogues with AI},
    journal = {Science},
    volume = {385},
    number = {6714},
    pages = {eadq1814},
    year = {2024},
    doi = {10.1126/science.adq1814},
    URL = {https://www.science.org/doi/abs/10.1126/science.adq1814},
    eprint = {https://www.science.org/doi/pdf/10.1126/science.adq1814}
}

@inproceedings{furumai-etal-2024-zero,
    title = "Zero-shot Persuasive Chatbots with {LLM}-Generated Strategies and Information Retrieval",
    author = "Furumai, Kazuaki  and
      Legaspi, Roberto  and
      Romero, Julio Cesar Vizcarra  and
      Yamazaki, Yudai  and
      Nishimura, Yasutaka  and
      Semnani, Sina  and
      Ikeda, Kazushi  and
      Shi, Weiyan  and
      Lam, Monica",
    editor = "Al-Onaizan, Yaser  and
      Bansal, Mohit  and
      Chen, Yun-Nung",
    booktitle = "Findings of the Association for Computational Linguistics: EMNLP 2024",
    month = nov,
    year = "2024",
    address = "Miami, Florida, USA",
    publisher = "Association for Computational Linguistics",
    url = "https://aclanthology.org/2024.findings-emnlp.656/",
    doi = "10.18653/v1/2024.findings-emnlp.656",
    pages = "11224--11249"
}

@inproceedings{chen-etal-2022-seamlessly,
    title = "Seamlessly Integrating Factual Information and Social Content with Persuasive Dialogue",
    author = "Chen, Maximillian  and
      Shi, Weiyan  and
      Yan, Feifan  and
      Hou, Ryan  and
      Zhang, Jingwen  and
      Sahay, Saurav  and
      Yu, Zhou",
    editor = "He, Yulan  and
      Ji, Heng  and
      Li, Sujian  and
      Liu, Yang  and
      Chang, Chua-Hui",
    booktitle = "Proceedings of the 2nd Conference of the Asia-Pacific Chapter of the Association for Computational Linguistics and the 12th International Joint Conference on Natural Language Processing (Volume 1: Long Papers)",
    month = nov,
    year = "2022",
    address = "Online only",
    publisher = "Association for Computational Linguistics",
    url = "https://aclanthology.org/2022.aacl-main.31/",
    doi = "10.18653/v1/2022.aacl-main.31",
    pages = "399--413"
}

@inproceedings{liu-etal-2021-towards,
    title = "Towards Emotional Support Dialog Systems",
    author = "Liu, Siyang  and
      Zheng, Chujie  and
      Demasi, Orianna  and
      Sabour, Sahand  and
      Li, Yu  and
      Yu, Zhou  and
      Jiang, Yong  and
      Huang, Minlie",
    editor = "Zong, Chengqing  and
      Xia, Fei  and
      Li, Wenjie  and
      Navigli, Roberto",
    booktitle = "Proceedings of the 59th Annual Meeting of the Association for Computational Linguistics and the 11th International Joint Conference on Natural Language Processing (Volume 1: Long Papers)",
    month = aug,
    year = "2021",
    address = "Online",
    publisher = "Association for Computational Linguistics",
    url = "https://aclanthology.org/2021.acl-long.269/",
    doi = "10.18653/v1/2021.acl-long.269",
    pages = "3469--3483"
}

@misc{zhu2025multiagentbenchevaluatingcollaborationcompetition,
      title={MultiAgentBench: Evaluating the Collaboration and Competition of LLM agents}, 
      author={Kunlun Zhu and Hongyi Du and Zhaochen Hong and Xiaocheng Yang and Shuyi Guo and Zhe Wang and Zhenhailong Wang and Cheng Qian and Xiangru Tang and Heng Ji and Jiaxuan You},
      year={2025},
      eprint={2503.01935},
      archivePrefix={arXiv},
      primaryClass={cs.MA},
      url={https://arxiv.org/abs/2503.01935}, 
}

@misc{breum2023persuasivepowerlargelanguage,
      title={The Persuasive Power of Large Language Models}, 
      author={Simon Martin Breum and Daniel Vædele Egdal and Victor Gram Mortensen and Anders Giovanni Møller and Luca Maria Aiello},
      year={2023},
      eprint={2312.15523},
      archivePrefix={arXiv},
      primaryClass={cs.CY},
      url={https://arxiv.org/abs/2312.15523}, 
}

@article{10.1093/joc/jqad024,
    author = {Huang, Guanxiong and Wang, Sai},
    title = {Is artificial intelligence more persuasive than humans? A meta-analysis},
    journal = {Journal of Communication},
    volume = {73},
    number = {6},
    pages = {552-562},
    year = {2023},
    month = {08},
    issn = {0021-9916},
    doi = {10.1093/joc/jqad024},
    url = {https://doi.org/10.1093/joc/jqad024},
    eprint = {https://academic.oup.com/joc/article-pdf/73/6/552/54463455/jqad024.pdf},
}

@article{10.1093/pnasnexus/pgae034,
    author = {Goldstein, Josh A and Chao, Jason and Grossman, Shelby and Stamos, Alex and Tomz, Michael},
    title = {How persuasive is AI-generated propaganda?},
    journal = {PNAS Nexus},
    volume = {3},
    number = {2},
    pages = {pgae034},
    year = {2024},
    month = {02},
    issn = {2752-6542},
    doi = {10.1093/pnasnexus/pgae034},
    url = {https://doi.org/10.1093/pnasnexus/pgae034},
    eprint = {https://academic.oup.com/pnasnexus/article-pdf/3/2/pgae034/56712546/pgae034.pdf},
}

@incollection{mcguire1969nature,
  author       = {McGuire, William J.},
  title        = {The nature of attitudes and attitude change},
  booktitle    = {The Handbook of Social Psychology},
  editor       = {Aronson, Elliot and Lindzey, Gardner},
  edition      = {2nd},
  volume       = {3},
  pages        = {136--314},
  year         = {1969},
  publisher    = {Addison-Wesley},
  address      = {Massachusetts}
}

@incollection{petty1986elaboration,
	title = {The {Elaboration} {Likelihood} {Model} of {Persuasion}},
	volume = {19},
	url = {https://www.sciencedirect.com/science/article/pii/S0065260108602142},
	urldate = {2025-03-13},
	booktitle = {Advances in {Experimental} {Social} {Psychology}},
	publisher = {Academic Press},
	author = {Petty, Richard E. and Cacioppo, John T.},
	editor = {Berkowitz, Leonard},
	month = jan,
	year = {1986},
	doi = {10.1016/S0065-2601(08)60214-2},
	pages = {123--205},
}

@article{chaiken1980heuristic,
	title = {Heuristic versus systematic information processing and the use of source versus message cues in persuasion},
	volume = {39},
	issn = {1939-1315},
	doi = {10.1037/0022-3514.39.5.752},
	number = {5},
	journal = {Journal of Personality and Social Psychology},
	author = {Chaiken, Shelly},
	year = {1980},
	note = {Place: US
Publisher: American Psychological Association},
	pages = {752--766},
}

@article{druckman2022framework,
	title = {A {Framework} for the {Study} of {Persuasion}},
	volume = {25},
	issn = {1094-2939, 1545-1577},
	url = {https://www.annualreviews.org/content/journals/10.1146/annurev-polisci-051120-110428},
	doi = {10.1146/annurev-polisci-051120-110428},
	language = {en},
	number = {Volume 25, 2022},
	urldate = {2025-03-06},
	journal = {Annual Review of Political Science},
	author = {Druckman, James N.},
	month = may,
	year = {2022},
	note = {Publisher: Annual Reviews},
	pages = {65--88},
}

@article{cialdini2001science,
	title = {The {Science} of {Persuasion}},
	volume = {284},
	issn = {0036-8733},
	url = {https://www.jstor.org/stable/26059056},
	number = {2},
	urldate = {2025-03-06},
	journal = {Scientific American},
	author = {Cialdini, Robert B.},
	year = {2001},
	note = {Publisher: Scientific American, a division of Nature America, Inc.},
	pages = {76--81},
}

@book{festinger1957theory,
	series = {A theory of cognitive dissonance},
	title = {A theory of cognitive dissonance},
	isbn = {978-0-8047-0131-0 978-0-8047-0911-8},
	publisher = {Stanford University Press},
	author = {Festinger, Leon},
	year = {1957},
	note = {Pages: xi, 291},
}

@incollection{asch1951effects,
	address = {Oxford, England},
	title = {Effects of group pressure upon the modification and distortion of judgments},
	booktitle = {Groups, leadership and men; research in human relations},
	publisher = {Carnegie Press},
	author = {Asch, S. E.},
	year = {1951},
	pages = {177--190},
}

@article{milgram1963behavioral,
	title = {Behavioral {Study} of obedience},
	volume = {67},
	issn = {0096-851X},
	doi = {10.1037/h0040525},
	number = {4},
	journal = {The Journal of Abnormal and Social Psychology},
	author = {Milgram, Stanley},
	year = {1963},
	note = {Place: US
Publisher: American Psychological Association},
	pages = {371--378},
}

@book{brehm1966theory,
	address = {Oxford, England},
	series = {A theory of psychological reactance},
	title = {A theory of psychological reactance},
	publisher = {Academic Press},
	author = {Brehm, Jack W.},
	year = {1966},
	note = {Pages: x, 135},
}

@inproceedings{fogg1997captology,
	address = {New York, NY, USA},
	series = {{CHI} {EA} '97},
	title = {Captology: the study of computers as persuasive technologies},
	isbn = {978-0-89791-926-5},
	shorttitle = {Captology},
	url = {https://dl.acm.org/doi/10.1145/1120212.1120301},
	doi = {10.1145/1120212.1120301},
	urldate = {2025-01-05},
	booktitle = {{CHI} '97 {Extended} {Abstracts} on {Human} {Factors} in {Computing} {Systems}},
	publisher = {Association for Computing Machinery},
	author = {Fogg, BJ},
	month = mar,
	year = {1997},
	pages = {129},
}

@inproceedings{fogg1998persuasive,
	address = {USA},
	series = {{CHI} '98},
	title = {Persuasive computers: perspectives and research directions},
	isbn = {978-0-201-30987-4},
	shorttitle = {Persuasive computers},
	url = {https://dl.acm.org/doi/10.1145/274644.274677},
	doi = {10.1145/274644.274677},
	urldate = {2025-01-05},
	booktitle = {Proceedings of the {SIGCHI} {Conference} on {Human} {Factors} in {Computing} {Systems}},
	publisher = {ACM Press/Addison-Wesley Publishing Co.},
	author = {Fogg, BJ},
	month = jan,
	year = {1998},
	pages = {225--232},
}

@inproceedings{fogg2009behavior,
	address = {New York, NY, USA},
	series = {Persuasive '09},
	title = {A behavior model for persuasive design},
	isbn = {978-1-60558-376-1},
	url = {https://dl.acm.org/doi/10.1145/1541948.1541999},
	doi = {10.1145/1541948.1541999},
	urldate = {2025-03-10},
	booktitle = {Proceedings of the 4th {International} {Conference} on {Persuasive} {Technology}},
	publisher = {Association for Computing Machinery},
	author = {Fogg, BJ},
	month = apr,
	year = {2009},
	pages = {1--7},
}

@inproceedings{fogg2009creating,
	address = {New York, NY, USA},
	series = {Persuasive '09},
	title = {Creating persuasive technologies: an eight-step design process},
	isbn = {978-1-60558-376-1},
	shorttitle = {Creating persuasive technologies},
	url = {https://dl.acm.org/doi/10.1145/1541948.1542005},
	doi = {10.1145/1541948.1542005},
	urldate = {2025-03-10},
	booktitle = {Proceedings of the 4th {International} {Conference} on {Persuasive} {Technology}},
	publisher = {Association for Computing Machinery},
	author = {Fogg, BJ},
	month = apr,
	year = {2009},
	pages = {1--6},
}

@inproceedings{consolvo2009theory-driven,
	address = {Boston MA USA},
	title = {Theory-driven design strategies for technologies that support behavior change in everyday life},
	isbn = {978-1-60558-246-7},
	url = {https://dl.acm.org/doi/10.1145/1518701.1518766},
	doi = {10.1145/1518701.1518766},
	language = {en},
	urldate = {2025-03-10},
	booktitle = {Proceedings of the {SIGCHI} {Conference} on {Human} {Factors} in {Computing} {Systems}},
	publisher = {ACM},
	author = {Consolvo, Sunny and McDonald, David W. and Landay, James A.},
	month = apr,
	year = {2009},
	pages = {405--414},
}

@inproceedings{jafarinaimi2005breakaway,
	address = {New York, NY, USA},
	series = {{CHI} {EA} '05},
	title = {Breakaway: an ambient display designed to change human behavior},
	isbn = {978-1-59593-002-6},
	shorttitle = {Breakaway},
	url = {https://dl.acm.org/doi/10.1145/1056808.1057063},
	doi = {10.1145/1056808.1057063},
	urldate = {2025-03-13},
	booktitle = {{CHI} '05 {Extended} {Abstracts} on {Human} {Factors} in {Computing} {Systems}},
	publisher = {Association for Computing Machinery},
	author = {Jafarinaimi, Nassim and Forlizzi, Jodi and Hurst, Amy and Zimmerman, John},
	month = apr,
	year = {2005},
	pages = {1945--1948},
}

@inproceedings{consolvo2008activity,
	address = {Florence Italy},
	title = {Activity sensing in the wild: a field trial of ubifit garden},
	isbn = {978-1-60558-011-1},
	shorttitle = {Activity sensing in the wild},
	url = {https://dl.acm.org/doi/10.1145/1357054.1357335},
	doi = {10.1145/1357054.1357335},
	language = {en},
	urldate = {2025-03-10},
	booktitle = {Proceedings of the {SIGCHI} {Conference} on {Human} {Factors} in {Computing} {Systems}},
	publisher = {ACM},
	author = {Consolvo, Sunny and McDonald, David W. and Toscos, Tammy and Chen, Mike Y. and Froehlich, Jon and Harrison, Beverly and Klasnja, Predrag and LaMarca, Anthony and LeGrand, Louis and Libby, Ryan and Smith, Ian and Landay, James A.},
	month = apr,
	year = {2008},
	pages = {1797--1806},
}

@inproceedings{dey2017art,
	address = {Portland Oregon USA},
	title = {The {Art} and {Science} of {Persuasion}: {Not} {All} {Crowdfunding} {Campaign} {Videos} are {The} {Same}},
	isbn = {978-1-4503-4335-0},
	shorttitle = {The {Art} and {Science} of {Persuasion}},
	url = {https://dl.acm.org/doi/10.1145/2998181.2998229},
	doi = {10.1145/2998181.2998229},
	language = {en},
	urldate = {2025-03-06},
	booktitle = {Proceedings of the 2017 {ACM} {Conference} on {Computer} {Supported} {Cooperative} {Work} and {Social} {Computing}},
	publisher = {ACM},
	author = {Dey, Sanorita and Duff, Brittany and Karahalios, Karrie and Fu, Wai-Tat},
	month = feb,
	year = {2017},
	pages = {755--769},
}

@article{xiao2019should,
	title = {Should {We} {Use} an {Abstract} {Comic} {Form} to {Persuade}?: {Experiments} with {Online} {Charitable} {Donation}},
	volume = {3},
	issn = {2573-0142},
	shorttitle = {Should {We} {Use} an {Abstract} {Comic} {Form} to {Persuade}?},
	url = {https://dl.acm.org/doi/10.1145/3359177},
	doi = {10.1145/3359177},
	language = {en},
	number = {CSCW},
	urldate = {2025-03-06},
	journal = {Proceedings of the ACM on Human-Computer Interaction},
	author = {Xiao, Ziang and Ho, Po-Shiun and Wang, Xinran and Karahalios, Karrie and Sundaram, Hari},
	month = nov,
	year = {2019},
	pages = {1--28},
}

@article{farrelly2009influence,
	title = {The {Influence} of the {National} truth® {Campaign} on {Smoking} {Initiation}},
	volume = {36},
	issn = {0749-3797, 1873-2607},
	url = {https://www.ajpmonline.org/article/S0749-3797(09)00074-9/fulltext},
	doi = {10.1016/j.amepre.2009.01.019},
	language = {English},
	number = {5},
	urldate = {2025-03-14},
	journal = {American Journal of Preventive Medicine},
	author = {Farrelly, Matthew C. and Nonnemaker, James and Davis, Kevin C. and Hussin, Altijani},
	month = may,
	year = {2009},
	note = {Publisher: Elsevier},
	pages = {379--384},
}

@article{danciu2014manipulative,
	title = {Manipulative marketing: persuasion and manipulation of the consumer through advertising},
	volume = {XXI},
	shorttitle = {Manipulative marketing},
	url = {https://ideas.repec.org//a/agr/journl/vxxiy2014i2(591)p19-34.html},
	language = {en},
	number = {2(591)},
	urldate = {2025-03-14},
	journal = {Theoretical and Applied Economics},
	author = {Danciu, Victor},
	year = {2014},
	note = {Publisher: Asociatia Generala a Economistilor din Romania / Editura Economica},
	pages = {19--34},
}

@article{markova2008persuasion,
	title = {Persuasion and {Propaganda}},
	volume = {55},
	issn = {0392-1921, 1467-7695},
	url = {https://www.cambridge.org/core/journals/diogenes/article/persuasion-and-propaganda/5B877DC7CC09B164EC95FAB20F209EDF},
	doi = {10.1177/0392192107087916},
	language = {en},
	number = {1},
	urldate = {2025-03-14},
	journal = {Diogenes},
	author = {Marková, Ivana},
	month = feb,
	year = {2008},
	pages = {37--51},
}

@misc{idziejczak2025themgamebasedframeworkassessing,
      title={Among Them: A game-based framework for assessing persuasion capabilities of LLMs}, 
      author={Mateusz Idziejczak and Vasyl Korzavatykh and Mateusz Stawicki and Andrii Chmutov and Marcin Korcz and Iwo Błądek and Dariusz Brzezinski},
      year={2025},
      eprint={2502.20426},
      archivePrefix={arXiv},
      primaryClass={cs.CL},
      url={https://arxiv.org/abs/2502.20426}, 
}

@misc{bai_voelkel_muldowney_eichstaedt_willer_2025,
 title={AI-Generated Messages Can Be Used to Persuade Humans on Policy Issues},
 url={osf.io/stakv_v5},
 DOI={10.31219/osf.io/stakv_v5},
 publisher={OSF Preprints},
 author={Bai, Hui and Voelkel, Jan G and Muldowney, Shane and Eichstaedt, johannes C and Willer, Robb},
 year={2025},
 month={Mar}
}

@article{gdppersuasion,
 ISSN = {00028282},
 URL = {http://www.jstor.org/stable/2117917},
 author = {Donald McCloskey and Arjo Klamer},
 journal = {The American Economic Review},
 number = {2},
 pages = {191--195},
 publisher = {American Economic Association},
 title = {One Quarter of GDP is Persuasion},
 urldate = {2025-03-18},
 volume = {85},
 year = {1995}
}

@article{gdppersuasion2, 
    author = {Antioch, Gerry},
    year = {2013},
    issue_date = {2013},
    publisher = {Department of the Treasury},
    number = {1},
    url = {https://search.informit.org/doi/10.3316/informit.558637667306970},
    journal = {Economic Round-Up},
    title = {Persuasion is now 30 per cent of US GDP},
    pages = {1–10},
    numpages = {10}
}

@inproceedings{shi-etal-2021-refine-imitate,
    title = "Refine and Imitate: Reducing Repetition and Inconsistency in Persuasion Dialogues via Reinforcement Learning and Human Demonstration",
    author = "Shi, Weiyan  and
      Li, Yu  and
      Sahay, Saurav  and
      Yu, Zhou",
    editor = "Moens, Marie-Francine  and
      Huang, Xuanjing  and
      Specia, Lucia  and
      Yih, Scott Wen-tau",
    booktitle = "Findings of the Association for Computational Linguistics: EMNLP 2021",
    month = nov,
    year = "2021",
    address = "Punta Cana, Dominican Republic",
    publisher = "Association for Computational Linguistics",
    url = "https://aclanthology.org/2021.findings-emnlp.295/",
    doi = "10.18653/v1/2021.findings-emnlp.295",
    pages = "3478--3492"
}

@article{cima2024contextualized,
  title={Contextualized Counterspeech: Strategies for Adaptation, Personalization, and Evaluation},
  author={Cima, Lorenzo and Miaschi, Alessio and Trujillo, Amaury and Avvenuti, Marco and Dell'Orletta, Felice and Cresci, Stefano},
  journal={arXiv preprint arXiv:2412.07338},
  year={2024}
}

@inproceedings{hunter2019towards,
  title={Towards computational persuasion via natural language argumentation dialogues},
  author={Hunter, Anthony and Chalaguine, Lisa and Czernuszenko, Tomasz and Hadoux, Emmanuel and Polberg, Sylwia},
  booktitle={KI 2019: Advances in Artificial Intelligence: 42nd German Conference on AI, Kassel, Germany, September 23--26, 2019, Proceedings 42},
  pages={18--33},
  year={2019},
  organization={Springer}
}

@misc{zhang2025persuasiondoubleblindmultidomaindialogue,
      title={Persuasion Should be Double-Blind: A Multi-Domain Dialogue Dataset With Faithfulness Based on Causal Theory of Mind}, 
      author={Dingyi Zhang and Deyu Zhou},
      year={2025},
      eprint={2502.21297},
      archivePrefix={arXiv},
      primaryClass={cs.CL},
      url={https://arxiv.org/abs/2502.21297}, 
}

@inproceedings{wei-etal-2016-post,
    title = "Is This Post Persuasive? Ranking Argumentative Comments in Online Forum",
    author = "Wei, Zhongyu  and
      Liu, Yang  and
      Li, Yi",
    editor = "Erk, Katrin  and
      Smith, Noah A.",
    booktitle = "Proceedings of the 54th Annual Meeting of the Association for Computational Linguistics (Volume 2: Short Papers)",
    month = aug,
    year = "2016",
    address = "Berlin, Germany",
    publisher = "Association for Computational Linguistics",
    url = "https://aclanthology.org/P16-2032/",
    doi = "10.18653/v1/P16-2032",
    pages = "195--200"
}

@misc{dughmi2016algorithmicbayesianpersuasion,
      title={Algorithmic Bayesian Persuasion}, 
      author={Shaddin Dughmi and Haifeng Xu},
      year={2016},
      eprint={1503.05988},
      archivePrefix={arXiv},
      primaryClass={cs.GT},
      url={https://arxiv.org/abs/1503.05988}, 
}

@article{bayesianpersuasion,
    Author = {Kamenica, Emir and Gentzkow, Matthew},
    Title = {Bayesian Persuasion},
    Journal = {American Economic Review},
    Volume = {101},
    Number = {6},
    Year = {2011},
    Month = {October},
    Pages = {2590–2615},
    DOI = {10.1257/aer.101.6.2590},
    URL = {https://www.aeaweb.org/articles?id=10.1257/aer.101.6.2590}
}

@inproceedings{hidey-etal-2017-analyzing,
    title = "Analyzing the Semantic Types of Claims and Premises in an Online Persuasive Forum",
    author = "Hidey, Christopher  and
      Musi, Elena  and
      Hwang, Alyssa  and
      Muresan, Smaranda  and
      McKeown, Kathy",
    editor = "Habernal, Ivan  and
      Gurevych, Iryna  and
      Ashley, Kevin  and
      Cardie, Claire  and
      Green, Nancy  and
      Litman, Diane  and
      Petasis, Georgios  and
      Reed, Chris  and
      Slonim, Noam  and
      Walker, Vern",
    booktitle = "Proceedings of the 4th Workshop on Argument Mining",
    month = sep,
    year = "2017",
    address = "Copenhagen, Denmark",
    publisher = "Association for Computational Linguistics",
    url = "https://aclanthology.org/W17-5102/",
    doi = "10.18653/v1/W17-5102",
    pages = "11--21"
}

@inproceedings{Khazaei2017,
  author    = {Taraneh Khazaei and Lu Xiao and Robert Mercer},
  title     = {Writing to persuade: Analysis and detection of persuasive discourse},
  booktitle = {iConference 2017 Proceedings},
  year      = {2017},
  publisher = {iSchools},
  url       = {http://hdl.handle.net/2142/96673}
}

@article{Addawood_Badawy_Lerman_Ferrara_2019, 
    title={Linguistic Cues to Deception: Identifying Political Trolls on Social Media}, 
    volume={13}, 
    url={https://ojs.aaai.org/index.php/ICWSM/article/view/3205},
    DOI={10.1609/icwsm.v13i01.3205}, 
    abstractNote={&lt;p&gt;The ease with which information can be shared on social media has opened it up to abuse and manipulation. One example of a manipulation campaign that has garnered much attention recently was the alleged Russian interference in the 2016 U.S. elections, with Russia accused of, among other things, using trolls and malicious accounts to spread misinformation and politically biased information. To take an in-depth look at this manipulation campaign, we collected a dataset of 13 million election-related posts shared on Twitter in 2016 by over a million distinct users. This dataset includes accounts associated with the identified Russian trolls as well as users sharing posts in the same time period on a variety of topics around the 2016 elections. To study how these trolls attempted to manipulate public opinion, we identified 49 theoretically grounded linguistic markers of deception and measured their use by troll and non-troll accounts. We show that deceptive language cues can help to accurately identify trolls, with average F1 score of 82% and recall 88%.&lt;/p&gt;}, 
    number={01},
    journal={Proceedings of the International AAAI Conference on Web and Social Media}, 
    author={Addawood, Aseel and Badawy, Adam and Lerman, Kristina and Ferrara, Emilio}, 
    year={2019}, 
    month={Jul.}, 
    pages={15-25} 
}

@inproceedings{yang-etal-2019-lets,
    title = "Let`s Make Your Request More Persuasive: Modeling Persuasive Strategies via Semi-Supervised Neural Nets on Crowdfunding Platforms",
    author = "Yang, Diyi  and
      Chen, Jiaao  and
      Yang, Zichao  and
      Jurafsky, Dan  and
      Hovy, Eduard",
    editor = "Burstein, Jill  and
      Doran, Christy  and
      Solorio, Thamar",
    booktitle = "Proceedings of the 2019 Conference of the North {A}merican Chapter of the Association for Computational Linguistics: Human Language Technologies, Volume 1 (Long and Short Papers)",
    month = jun,
    year = "2019",
    address = "Minneapolis, Minnesota",
    publisher = "Association for Computational Linguistics",
    url = "https://aclanthology.org/N19-1364/",
    doi = "10.18653/v1/N19-1364",
    pages = "3620--3630"
}

@article{dutta-changing-views-2019,
    title = {Changing views: Persuasion modeling and argument extraction from online discussions},
    journal = {Information Processing \& Management},
    volume = {57},
    number = {2},
    pages = {102085},
    year = {2020},
    issn = {0306-4573},
    doi = {https://doi.org/10.1016/j.ipm.2019.102085},
    url = {https://www.sciencedirect.com/science/article/pii/S0306457319301165},
    author = {Subhabrata Dutta and Dipankar Das and Tanmoy Chakraborty}
}

@inproceedings{chakrabarty-etal-2019-ampersand,
    title = "{AMPERSAND}: Argument Mining for {PERS}u{A}sive o{N}line Discussions",
    author = "Chakrabarty, Tuhin  and
      Hidey, Christopher  and
      Muresan, Smaranda  and
      McKeown, Kathy  and
      Hwang, Alyssa",
    editor = "Inui, Kentaro  and
      Jiang, Jing  and
      Ng, Vincent  and
      Wan, Xiaojun",
    booktitle = "Proceedings of the 2019 Conference on Empirical Methods in Natural Language Processing and the 9th International Joint Conference on Natural Language Processing (EMNLP-IJCNLP)",
    month = nov,
    year = "2019",
    address = "Hong Kong, China",
    publisher = "Association for Computational Linguistics",
    url = "https://aclanthology.org/D19-1291/",
    doi = "10.18653/v1/D19-1291",
    pages = "2933--2943"
}

@inproceedings{shaikh-etal-2020-examining,
    title = "Examining the Ordering of Rhetorical Strategies in Persuasive Requests",
    author = "Shaikh, Omar  and
      Chen, Jiaao  and
      Saad-Falcon, Jon  and
      Chau, Polo  and
      Yang, Diyi",
    editor = "Cohn, Trevor  and
      He, Yulan  and
      Liu, Yang",
    booktitle = "Findings of the Association for Computational Linguistics: EMNLP 2020",
    month = nov,
    year = "2020",
    address = "Online",
    publisher = "Association for Computational Linguistics",
    url = "https://aclanthology.org/2020.findings-emnlp.116/",
    doi = "10.18653/v1/2020.findings-emnlp.116",
    pages = "1299--1306"
}

@inproceedings{pauli-etal-2022-modelling,
    title = "Modelling Persuasion through Misuse of Rhetorical Appeals",
    author = "Pauli, Amalie  and
      Derczynski, Leon  and
      Assent, Ira",
    editor = "Biester, Laura  and
      Demszky, Dorottya  and
      Jin, Zhijing  and
      Sachan, Mrinmaya  and
      Tetreault, Joel  and
      Wilson, Steven  and
      Xiao, Lu  and
      Zhao, Jieyu",
    booktitle = "Proceedings of the Second Workshop on NLP for Positive Impact (NLP4PI)",
    month = dec,
    year = "2022",
    address = "Abu Dhabi, United Arab Emirates (Hybrid)",
    publisher = "Association for Computational Linguistics",
    url = "https://aclanthology.org/2022.nlp4pi-1.11/",
    doi = "10.18653/v1/2022.nlp4pi-1.11",
    pages = "89--100"
}

@misc{wojtowicz2024persuasionhardcomputationalcomplexity,
      title={When and Why is Persuasion Hard? A Computational Complexity Result}, 
      author={Zachary Wojtowicz},
      year={2024},
      eprint={2408.07923},
      archivePrefix={arXiv},
      primaryClass={cs.CY},
      url={https://arxiv.org/abs/2408.07923}, 
}

@misc{li2025verbalizedbayesianpersuasion,
      title={Verbalized Bayesian Persuasion}, 
      author={Wenhao Li and Yue Lin and Xiangfeng Wang and Bo Jin and Hongyuan Zha and Baoxiang Wang},
      year={2025},
      eprint={2502.01587},
      archivePrefix={arXiv},
      primaryClass={cs.GT},
      url={https://arxiv.org/abs/2502.01587}, 
}

@misc{musi2018changemyviewconcessionsconcessionsincrease,
      title={ChangeMyView Through Concessions: Do Concessions Increase Persuasion?}, 
      author={Elena Musi and Debanjan Ghosh and Smaranda Muresan},
      year={2018},
      eprint={1806.03223},
      archivePrefix={arXiv},
      primaryClass={cs.CL},
      url={https://arxiv.org/abs/1806.03223}, 
}

@inproceedings{durmus-cardie-2018-exploring,
    title = "Exploring the Role of Prior Beliefs for Argument Persuasion",
    author = "Durmus, Esin  and
      Cardie, Claire",
    editor = "Walker, Marilyn  and
      Ji, Heng  and
      Stent, Amanda",
    booktitle = "Proceedings of the 2018 Conference of the North {A}merican Chapter of the Association for Computational Linguistics: Human Language Technologies, Volume 1 (Long Papers)",
    month = jun,
    year = "2018",
    address = "New Orleans, Louisiana",
    publisher = "Association for Computational Linguistics",
    url = "https://aclanthology.org/N18-1094/",
    doi = "10.18653/v1/N18-1094",
    pages = "1035--1045"
}

@article{Palmer02092023,
    author = {Alexis Palmer and Arthur Spirling and},
    title = {Large Language Models Can Argue in Convincing Ways About Politics, But Humans Dislike AI Authors: implications for Governance},
    journal = {Political Science},
    volume = {75},
    number = {3},
    pages = {281--291},
    year = {2023},
    publisher = {Routledge},
    doi = {10.1080/00323187.2024.2335471},
    URL = { https://doi.org/10.1080/00323187.2024.2335471},
    eprint = {https://doi.org/10.1080/00323187.2024.2335471}
}

@inproceedings{piskorski-etal-2023-semeval,
    title = "{S}em{E}val-2023 Task 3: Detecting the Category, the Framing, and the Persuasion Techniques in Online News in a Multi-lingual Setup",
    author = "Piskorski, Jakub  and
      Stefanovitch, Nicolas  and
      Da San Martino, Giovanni  and
      Nakov, Preslav",
    editor = {Ojha, Atul Kr.  and
      Do{\u{g}}ru{\"o}z, A. Seza  and
      Da San Martino, Giovanni  and
      Tayyar Madabushi, Harish  and
      Kumar, Ritesh  and
      Sartori, Elisa},
    booktitle = "Proceedings of the 17th International Workshop on Semantic Evaluation (SemEval-2023)",
    month = jul,
    year = "2023",
    address = "Toronto, Canada",
    publisher = "Association for Computational Linguistics",
    url = "https://aclanthology.org/2023.semeval-1.317/",
    doi = "10.18653/v1/2023.semeval-1.317",
    pages = "2343--2361"
}

@inproceedings{anil2024manyshot,
    title={Many-shot Jailbreaking},
    author={Cem Anil and Esin DURMUS and Nina Rimsky and Mrinank Sharma and Joe Benton and Sandipan Kundu and Joshua Batson and Meg Tong and Jesse Mu and Daniel J Ford and Francesco Mosconi and Rajashree Agrawal and Rylan Schaeffer and Naomi Bashkansky and Samuel Svenningsen and Mike Lambert and Ansh Radhakrishnan and Carson Denison and Evan J Hubinger and Yuntao Bai and Trenton Bricken and Timothy Maxwell and Nicholas Schiefer and James Sully and Alex Tamkin and Tamera Lanham and Karina Nguyen and Tomasz Korbak and Jared Kaplan and Deep Ganguli and Samuel R. Bowman and Ethan Perez and Roger Baker Grosse and David Duvenaud},
    booktitle={The Thirty-eighth Annual Conference on Neural Information Processing Systems},
    year={2024},
    url={https://openreview.net/forum?id=cw5mgd71jW}
}

@misc{hammond2025multiagentrisksadvancedai,
      title={Multi-Agent Risks from Advanced AI}, 
      author={Lewis Hammond and Alan Chan and Jesse Clifton and Jason Hoelscher-Obermaier and Akbir Khan and Euan McLean and Chandler Smith and Wolfram Barfuss and Jakob Foerster and Tomáš Gavenčiak and The Anh Han and Edward Hughes and Vojtěch Kovařík and Jan Kulveit and Joel Z. Leibo and Caspar Oesterheld and Christian Schroeder de Witt and Nisarg Shah and Michael Wellman and Paolo Bova and Theodor Cimpeanu and Carson Ezell and Quentin Feuillade-Montixi and Matija Franklin and Esben Kran and Igor Krawczuk and Max Lamparth and Niklas Lauffer and Alexander Meinke and Sumeet Motwani and Anka Reuel and Vincent Conitzer and Michael Dennis and Iason Gabriel and Adam Gleave and Gillian Hadfield and Nika Haghtalab and Atoosa Kasirzadeh and Sébastien Krier and Kate Larson and Joel Lehman and David C. Parkes and Georgios Piliouras and Iyad Rahwan},
      year={2025},
      eprint={2502.14143},
      archivePrefix={arXiv},
      primaryClass={cs.MA},
      url={https://arxiv.org/abs/2502.14143}, 
}

@inproceedings{keizer-etal-2017-evaluating,
    title = "Evaluating Persuasion Strategies and Deep Reinforcement Learning methods for Negotiation Dialogue agents",
    author = "Keizer, Simon  and
      Guhe, Markus  and
      Cuay{\'a}huitl, Heriberto  and
      Efstathiou, Ioannis  and
      Engelbrecht, Klaus-Peter  and
      Dobre, Mihai  and
      Lascarides, Alex  and
      Lemon, Oliver",
    editor = "Lapata, Mirella  and
      Blunsom, Phil  and
      Koller, Alexander",
    booktitle = "Proceedings of the 15th Conference of the {E}uropean Chapter of the Association for Computational Linguistics: Volume 2, Short Papers",
    month = apr,
    year = "2017",
    address = "Valencia, Spain",
    publisher = "Association for Computational Linguistics",
    url = "https://aclanthology.org/E17-2077/",
    pages = "480--484"
}

@inproceedings{he-etal-2018-decoupling,
    title = "Decoupling Strategy and Generation in Negotiation Dialogues",
    author = "He, He  and
      Chen, Derek  and
      Balakrishnan, Anusha  and
      Liang, Percy",
    editor = "Riloff, Ellen  and
      Chiang, David  and
      Hockenmaier, Julia  and
      Tsujii, Jun{'}ichi",
    booktitle = "Proceedings of the 2018 Conference on Empirical Methods in Natural Language Processing",
    month = oct # "-" # nov,
    year = "2018",
    address = "Brussels, Belgium",
    publisher = "Association for Computational Linguistics",
    url = "https://aclanthology.org/D18-1256/",
    doi = "10.18653/v1/D18-1256",
    pages = "2333--2343"
}

@inproceedings{
joshi2021dialograph,
title={DialoGraph: Incorporating Interpretable Strategy-Graph Networks into Negotiation Dialogues},
author={Rishabh Joshi and Vidhisha Balachandran and Shikhar Vashishth and Alan Black and Yulia Tsvetkov},
booktitle={International Conference on Learning Representations},
year={2021},
url={https://openreview.net/forum?id=kDnal_bbb-E}
}

@inproceedings{chawla-etal-2021-casino,
    title = "{C}a{S}i{N}o: A Corpus of Campsite Negotiation Dialogues for Automatic Negotiation Systems",
    author = "Chawla, Kushal  and
      Ramirez, Jaysa  and
      Clever, Rene  and
      Lucas, Gale  and
      May, Jonathan  and
      Gratch, Jonathan",
    editor = "Toutanova, Kristina  and
      Rumshisky, Anna  and
      Zettlemoyer, Luke  and
      Hakkani-Tur, Dilek  and
      Beltagy, Iz  and
      Bethard, Steven  and
      Cotterell, Ryan  and
      Chakraborty, Tanmoy  and
      Zhou, Yichao",
    booktitle = "Proceedings of the 2021 Conference of the North American Chapter of the Association for Computational Linguistics: Human Language Technologies",
    month = jun,
    year = "2021",
    address = "Online",
    publisher = "Association for Computational Linguistics",
    url = "https://aclanthology.org/2021.naacl-main.254/",
    doi = "10.18653/v1/2021.naacl-main.254",
    pages = "3167--3185"
}

@misc{rogiers2024persuasionlargelanguagemodels,
      title={Persuasion with Large Language Models: a Survey}, 
      author={Alexander Rogiers and Sander Noels and Maarten Buyl and Tijl De Bie},
      year={2024},
      eprint={2411.06837},
      archivePrefix={arXiv},
      primaryClass={cs.CL},
      url={https://arxiv.org/abs/2411.06837}, 
}

@article{Bakir2018,
  author    = {Vian Bakir and Eric Herring and David Miller and Piers Robinson},
  title     = {Organized Persuasive Communication: A new conceptual framework for research on public relations, propaganda and promotional culture},
  journal   = {Critical Sociology},
  volume    = {45},
  number    = {3},
  pages     = {311--328},
  year      = {2018},
  doi       = {10.1177/0896920518764586}
}

@article{Siddiqi2022,
  author    = {Murtaza Ahmed Siddiqi and Wooguil Pak and Moquddam A. Siddiqi},
  title     = {A Study on the Psychology of Social Engineering-Based Cyberattacks and Existing Countermeasures},
  journal   = {Applied Sciences},
  volume    = {12},
  number    = {12},
  pages     = {6042},
  year      = {2022},
  doi       = {10.3390/app12126042}
}

@article{Lock2020,
  author    = {Irina Lock and Ramona Ludolph},
  title     = {Organizational propaganda on the Internet: A systematic review},
  journal   = {Public Relations Inquiry},
  volume    = {9},
  number    = {1},
  pages     = {103--127},
  year      = {2020},
  doi       = {10.1177/2046147X19870844}
}

@inproceedings{Ferreyra_2020,
   title={Persuasion Meets AI: Ethical Considerations for the Design of Social Engineering Countermeasures},
   url={http://dx.doi.org/10.5220/0010142402040211},
   DOI={10.5220/0010142402040211},
   booktitle={Proceedings of the 12th International Joint Conference on Knowledge Discovery, Knowledge Engineering and Knowledge Management},
   publisher={SCITEPRESS - Science and Technology Publications},
   author={Ferreyra, Nicolás and Aïmeur, Esma and Hage, Hicham and Heisel, Maritta and van Hoogstraten, Catherine},
   year={2020},
   pages={204–211} }

@article{kamenica2011bayesian,
	title = {Bayesian {Persuasion}},
	volume = {101},
	issn = {0002-8282},
	url = {https://www.aeaweb.org/articles?id=10.1257/aer.101.6.2590},
	doi = {10.1257/aer.101.6.2590},
	language = {en},
	number = {6},
	urldate = {2025-03-25},
	journal = {American Economic Review},
	author = {Kamenica, Emir and Gentzkow, Matthew},
	month = oct,
	year = {2011},
	pages = {2590--2615},
}

@article{milgrom1981good,
	title = {Good {News} and {Bad} {News}: {Representation} {Theorems} and {Applications}},
	volume = {12},
	issn = {0361915X},
	shorttitle = {Good {News} and {Bad} {News}},
	url = {https://www.jstor.org/stable/3003562?origin=crossref},
	doi = {10.2307/3003562},
	language = {en},
	number = {2},
	urldate = {2025-04-20},
	journal = {The Bell Journal of Economics},
	author = {Milgrom, Paul R.},
	year = {1981},
	pages = {380},
}

@article{grossman1981informational,
	title = {The {Informational} {Role} of {Warranties} and {Private} {Disclosure} about {Product} {Quality}},
	volume = {24},
	issn = {0022-2186, 1537-5285},
	url = {https://www.journals.uchicago.edu/doi/10.1086/466995},
	doi = {10.1086/466995},
	language = {en},
	number = {3},
	urldate = {2025-04-20},
	journal = {The Journal of Law and Economics},
	author = {Grossman, Sanford J.},
	month = dec,
	year = {1981},
	pages = {461--483},
}

@article{crawford1982strategic,
	title = {Strategic {Information} {Transmission}},
	volume = {50},
	issn = {00129682},
	url = {https://www.jstor.org/stable/1913390?origin=crossref},
	doi = {10.2307/1913390},
	language = {en},
	number = {6},
	urldate = {2025-04-20},
	journal = {Econometrica},
	author = {Crawford, Vincent P. and Sobel, Joel},
	month = nov,
	year = {1982},
	pages = {1431},
}

@article{spence1973job,
	title = {Job {Market} {Signaling}},
	volume = {87},
	issn = {00335533},
	url = {https://academic.oup.com/qje/article-lookup/doi/10.2307/1882010},
	doi = {10.2307/1882010},
	language = {en},
	number = {3},
	urldate = {2025-04-20},
	journal = {The Quarterly Journal of Economics},
	author = {Spence, Michael},
	month = aug,
	year = {1973},
	pages = {355},
}

@inproceedings{factorsofsuccessonlinedebate,
    author = {Durmus, Esin and Cardie, Claire},
    title = {Modeling the Factors of User Success in Online Debate},
    year = {2019},
    isbn = {9781450366748},
    publisher = {Association for Computing Machinery},
    address = {New York, NY, USA},
    url = {https://doi.org/10.1145/3308558.3313676},
    doi = {10.1145/3308558.3313676},
    booktitle = {The World Wide Web Conference},
    pages = {2701–2707},
    numpages = {7},
    location = {San Francisco, CA, USA},
    series = {WWW '19}
    }

@misc{devlin2019bertpretrainingdeepbidirectional,
      title={BERT: Pre-training of Deep Bidirectional Transformers for Language Understanding}, 
      author={Jacob Devlin and Ming-Wei Chang and Kenton Lee and Kristina Toutanova},
      year={2019},
      eprint={1810.04805},
      archivePrefix={arXiv},
      primaryClass={cs.CL},
      url={https://arxiv.org/abs/1810.04805}, 
}

@misc{brown2020languagemodelsfewshotlearners,
      title={Language Models are Few-Shot Learners}, 
      author={Tom B. Brown and Benjamin Mann and Nick Ryder and Melanie Subbiah and Jared Kaplan and Prafulla Dhariwal and Arvind Neelakantan and Pranav Shyam and Girish Sastry and Amanda Askell and Sandhini Agarwal and Ariel Herbert-Voss and Gretchen Krueger and Tom Henighan and Rewon Child and Aditya Ramesh and Daniel M. Ziegler and Jeffrey Wu and Clemens Winter and Christopher Hesse and Mark Chen and Eric Sigler and Mateusz Litwin and Scott Gray and Benjamin Chess and Jack Clark and Christopher Berner and Sam McCandlish and Alec Radford and Ilya Sutskever and Dario Amodei},
      year={2020},
      eprint={2005.14165},
      archivePrefix={arXiv},
      primaryClass={cs.CL},
      url={https://arxiv.org/abs/2005.14165}, 
}

@misc{openai2024gpt4technicalreport,
      title={GPT-4 Technical Report},
      author={OpenAI},
      year={2024},
      eprint={2303.08774},
      archivePrefix={arXiv},
      primaryClass={cs.CL},
      url={https://arxiv.org/abs/2303.08774}, 
}

@inproceedings{durmus-etal-2019-role,
    title = "The Role of Pragmatic and Discourse Context in Determining Argument Impact",
    author = "Durmus, Esin  and
      Ladhak, Faisal  and
      Cardie, Claire",
    editor = "Inui, Kentaro  and
      Jiang, Jing  and
      Ng, Vincent  and
      Wan, Xiaojun",
    booktitle = "Proceedings of the 2019 Conference on Empirical Methods in Natural Language Processing and the 9th International Joint Conference on Natural Language Processing (EMNLP-IJCNLP)",
    month = nov,
    year = "2019",
    address = "Hong Kong, China",
    publisher = "Association for Computational Linguistics",
    url = "https://aclanthology.org/D19-1568/",
    doi = "10.18653/v1/D19-1568",
    pages = "5668--5678"
}

@inproceedings{karande-etal-2024-persuasion,
    title = "Persuasion Games with Large Language Models",
    author = "Karande, Shirish  and
      V, Santhosh  and
      Bhatia, Yash",
    editor = "Lalitha Devi, Sobha  and
      Arora, Karunesh",
    booktitle = "Proceedings of the 21st International Conference on Natural Language Processing (ICON)",
    month = dec,
    year = "2024",
    address = "AU-KBC Research Centre, Chennai, India",
    publisher = "NLP Association of India (NLPAI)",
    url = "https://aclanthology.org/2024.icon-1.67/",
    pages = "576--582"
}

@inproceedings{roller-etal-2021-recipes,
    title = "Recipes for Building an Open-Domain Chatbot",
    author = "Roller, Stephen  and
      Dinan, Emily  and
      Goyal, Naman  and
      Ju, Da  and
      Williamson, Mary  and
      Liu, Yinhan  and
      Xu, Jing  and
      Ott, Myle  and
      Smith, Eric Michael  and
      Boureau, Y-Lan  and
      Weston, Jason",
    editor = "Merlo, Paola  and
      Tiedemann, Jorg  and
      Tsarfaty, Reut",
    booktitle = "Proceedings of the 16th Conference of the European Chapter of the Association for Computational Linguistics: Main Volume",
    month = apr,
    year = "2021",
    address = "Online",
    publisher = "Association for Computational Linguistics",
    url = "https://aclanthology.org/2021.eacl-main.24/",
    doi = "10.18653/v1/2021.eacl-main.24",
    pages = "300--325"
}

@inproceedings{zhang-etal-2020-dialogpt,
    title = "{DIALOGPT} : Large-Scale Generative Pre-training for Conversational Response Generation",
    author = "Zhang, Yizhe  and
      Sun, Siqi  and
      Galley, Michel  and
      Chen, Yen-Chun  and
      Brockett, Chris  and
      Gao, Xiang  and
      Gao, Jianfeng  and
      Liu, Jingjing  and
      Dolan, Bill",
    editor = "Celikyilmaz, Asli  and
      Wen, Tsung-Hsien",
    booktitle = "Proceedings of the 58th Annual Meeting of the Association for Computational Linguistics: System Demonstrations",
    month = jul,
    year = "2020",
    address = "Online",
    publisher = "Association for Computational Linguistics",
    url = "https://aclanthology.org/2020.acl-demos.30/",
    doi = "10.18653/v1/2020.acl-demos.30",
    pages = "270--278"
}

@inproceedings{lewis-etal-2020-bart,
    title = "{BART}: Denoising Sequence-to-Sequence Pre-training for Natural Language Generation, Translation, and Comprehension",
    author = "Lewis, Mike  and
      Liu, Yinhan  and
      Goyal, Naman  and
      Ghazvininejad, Marjan  and
      Mohamed, Abdelrahman  and
      Levy, Omer  and
      Stoyanov, Veselin  and
      Zettlemoyer, Luke",
    editor = "Jurafsky, Dan  and
      Chai, Joyce  and
      Schluter, Natalie  and
      Tetreault, Joel",
    booktitle = "Proceedings of the 58th Annual Meeting of the Association for Computational Linguistics",
    month = jul,
    year = "2020",
    address = "Online",
    publisher = "Association for Computational Linguistics",
    url = "https://aclanthology.org/2020.acl-main.703/",
    doi = "10.18653/v1/2020.acl-main.703",
    pages = "7871--7880"
}

@misc{schulman2017proximalpolicyoptimizationalgorithms,
      title={Proximal Policy Optimization Algorithms}, 
      author={John Schulman and Filip Wolski and Prafulla Dhariwal and Alec Radford and Oleg Klimov},
      year={2017},
      eprint={1707.06347},
      archivePrefix={arXiv},
      primaryClass={cs.LG},
      url={https://arxiv.org/abs/1707.06347}, 
}

@inproceedings{mishra-etal-2022-pepds,
    title = "{PEPDS}: A Polite and Empathetic Persuasive Dialogue System for Charity Donation",
    author = "Mishra, Kshitij  and
      Samad, Azlaan Mustafa  and
      Totala, Palak  and
      Ekbal, Asif",
    editor = "Calzolari, Nicoletta  and
      Huang, Chu-Ren  and
      Kim, Hansaem  and
      Pustejovsky, James  and
      Wanner, Leo  and
      Choi, Key-Sun  and
      Ryu, Pum-Mo  and
      Chen, Hsin-Hsi  and
      Donatelli, Lucia  and
      Ji, Heng  and
      Kurohashi, Sadao  and
      Paggio, Patrizia  and
      Xue, Nianwen  and
      Kim, Seokhwan  and
      Hahm, Younggyun  and
      He, Zhong  and
      Lee, Tony Kyungil  and
      Santus, Enrico  and
      Bond, Francis  and
      Na, Seung-Hoon",
    booktitle = "Proceedings of the 29th International Conference on Computational Linguistics",
    month = oct,
    year = "2022",
    address = "Gyeongju, Republic of Korea",
    publisher = "International Committee on Computational Linguistics",
    url = "https://aclanthology.org/2022.coling-1.34/",
    pages = "424--440"
}

@misc{
hong2025interactive,
title={Interactive Dialogue Agents via Reinforcement Learning with Hindsight Regenerations},
author={Joey Hong and Jessica Lin and Sergey Levine and Anca Dragan},
year={2025},
url={https://openreview.net/forum?id=hrGOMrfc2z}
}

@inproceedings{dpo,
 author = {Rafailov, Rafael and Sharma, Archit and Mitchell, Eric and Manning, Christopher D and Ermon, Stefano and Finn, Chelsea},
 booktitle = {Advances in Neural Information Processing Systems},
 editor = {A. Oh and T. Naumann and A. Globerson and K. Saenko and M. Hardt and S. Levine},
 pages = {53728--53741},
 publisher = {Curran Associates, Inc.},
 title = {Direct Preference Optimization: Your Language Model is Secretly a Reward Model},
 url = {https://proceedings.neurips.cc/paper_files/paper/2023/file/a85b405ed65c6477a4fe8302b5e06ce7-Paper-Conference.pdf},
 volume = {36},
 year = {2023}
}

@inproceedings{sharma2024towards,
    title={Towards Understanding Sycophancy in Language Models},
    author={Mrinank Sharma and Meg Tong and Tomasz Korbak and David Duvenaud and Amanda Askell and Samuel R. Bowman and Esin DURMUS and Zac Hatfield-Dodds and Scott R Johnston and Shauna M Kravec and Timothy Maxwell and Sam McCandlish and Kamal Ndousse and Oliver Rausch and Nicholas Schiefer and Da Yan and Miranda Zhang and Ethan Perez},
    booktitle={The Twelfth International Conference on Learning Representations},
    year={2024},
    url={https://openreview.net/forum?id=tvhaxkMKAn}
}

@inproceedings{perez-etal-2022-red,
    title = "Red Teaming Language Models with Language Models",
    author = "Perez, Ethan  and
      Huang, Saffron  and
      Song, Francis  and
      Cai, Trevor  and
      Ring, Roman  and
      Aslanides, John  and
      Glaese, Amelia  and
      McAleese, Nat  and
      Irving, Geoffrey",
    editor = "Goldberg, Yoav  and
      Kozareva, Zornitsa  and
      Zhang, Yue",
    booktitle = "Proceedings of the 2022 Conference on Empirical Methods in Natural Language Processing",
    month = dec,
    year = "2022",
    address = "Abu Dhabi, United Arab Emirates",
    publisher = "Association for Computational Linguistics",
    url = "https://aclanthology.org/2022.emnlp-main.225/",
    doi = "10.18653/v1/2022.emnlp-main.225",
    pages = "3419--3448"
}

@inproceedings{multimodal-2014,
    author = {Park, Sunghyun and Shim, Han Suk and Chatterjee, Moitreya and Sagae, Kenji and Morency, Louis-Philippe},
    title = {Computational Analysis of Persuasiveness in Social Multimedia: A Novel Dataset and Multimodal Prediction Approach},
    year = {2014},
    isbn = {9781450328852},
    publisher = {Association for Computing Machinery},
    address = {New York, NY, USA},
    url = {https://doi.org/10.1145/2663204.2663260},
    doi = {10.1145/2663204.2663260},
    booktitle = {Proceedings of the 16th International Conference on Multimodal Interaction},
    pages = {50–57},
    numpages = {8},
    location = {Istanbul, Turkey},
    series = {ICMI '14}
}

@article{m2p2,
  author={Bai, Chongyang and Chen, Haipeng and Kumar, Srijan and Leskovec, Jure and Subrahmanian, V. S.},
  journal={IEEE Transactions on Multimedia}, 
  title={M2P2: Multimodal Persuasion Prediction Using Adaptive Fusion}, 
  year={2023},
  volume={25},
  number={},
  pages={942-952},
  doi={10.1109/TMM.2021.3134168}
}

@inproceedings{liu-etal-2022-imagearg,
    title = "{I}mage{A}rg: A Multi-modal Tweet Dataset for Image Persuasiveness Mining",
    author = "Liu, Zhexiong  and
      Guo, Meiqi  and
      Dai, Yue  and
      Litman, Diane",
    editor = "Lapesa, Gabriella  and
      Schneider, Jodi  and
      Jo, Yohan  and
      Saha, Sougata",
    booktitle = "Proceedings of the 9th Workshop on Argument Mining",
    month = oct,
    year = "2022",
    address = "Online and in Gyeongju, Republic of Korea",
    publisher = "International Conference on Computational Linguistics",
    url = "https://aclanthology.org/2022.argmining-1.1/",
    pages = "1--18"
}

@inproceedings{lai-etal-2023-werewolf,
    title = "Werewolf Among Us: Multimodal Resources for Modeling Persuasion Behaviors in Social Deduction Games",
    author = "Lai, Bolin  and
      Zhang, Hongxin  and
      Liu, Miao  and
      Pariani, Aryan  and
      Ryan, Fiona  and
      Jia, Wenqi  and
      Hayati, Shirley Anugrah  and
      Rehg, James  and
      Yang, Diyi",
    editor = "Rogers, Anna  and
      Boyd-Graber, Jordan  and
      Okazaki, Naoaki",
    booktitle = "Findings of the Association for Computational Linguistics: ACL 2023",
    month = jul,
    year = "2023",
    address = "Toronto, Canada",
    publisher = "Association for Computational Linguistics",
    url = "https://aclanthology.org/2023.findings-acl.411/",
    doi = "10.18653/v1/2023.findings-acl.411",
    pages = "6570--6588"
}

@article{strat-in-advertisement, 
    title={Persuasion Strategies in Advertisements}, 
    volume={37}, 
    url={https://ojs.aaai.org/index.php/AAAI/article/view/25076}, 
    DOI={10.1609/aaai.v37i1.25076}, 
    abstractNote={Modeling what makes an advertisement persuasive, i.e., eliciting the desired response from consumer, is critical to the study of propaganda, social psychology, and marketing. Despite its importance, computational modeling of persuasion in computer vision is still in its infancy, primarily due to the lack of benchmark datasets that can provide persuasion-strategy labels associated with ads. Motivated by persuasion literature in social psychology and marketing, we introduce an extensive vocabulary of persuasion strategies and build the first ad image corpus annotated with persuasion strategies. We then formulate the task of persuasion strategy prediction with multi-modal learning, where we design a multi-task attention fusion model that can leverage other ad-understanding tasks to predict persuasion strategies. The dataset also provides image segmentation masks, which labels persuasion strategies in the corresponding ad images on the test split. We publicly release our code and dataset at https://midas-research.github.io/persuasion-advertisements/.}, 
    number={1}, 
    journal={Proceedings of the AAAI Conference on Artificial Intelligence}, 
    author={Kumar, Yaman and Jha, Rajat and Gupta, Arunim and Aggarwal, Milan and Garg, Aditya and Malyan, Tushar and Bhardwaj, Ayush and Ratn Shah, Rajiv and Krishnamurthy, Balaji and Chen, Changyou}, 
    year={2023}, 
    month={Jun.}, 
    pages={57-66} 
}

@book{aristotle1984rhetoric,
	title = {Rhetoric},
	isbn = {978-0-394-33924-5},
	publisher = {Modern Library},
	author = {{Aristotle} and Roberts, W.R. and Bywater, I.},
	year = {1984},
	lccn = {84003466},
}

@misc{wu2025groundedpersuasivelanguagegeneration,
      title={Grounded Persuasive Language Generation for Automated Marketing}, 
      author={Jibang Wu and Chenghao Yang and Simon Mahns and Chaoqi Wang and Hao Zhu and Fei Fang and Haifeng Xu},
      year={2025},
      eprint={2502.16810},
      archivePrefix={arXiv},
      primaryClass={cs.AI},
      url={https://arxiv.org/abs/2502.16810}, 
}

@inproceedings{stengel-eskin-etal-2025-teaching,
    title = "Teaching Models to Balance Resisting and Accepting Persuasion",
    author = "Stengel-Eskin, Elias  and
      Hase, Peter  and
      Bansal, Mohit",
    editor = "Chiruzzo, Luis  and
      Ritter, Alan  and
      Wang, Lu",
    booktitle = "Proceedings of the 2025 Conference of the Nations of the Americas Chapter of the Association for Computational Linguistics: Human Language Technologies (Volume 1: Long Papers)",
    month = apr,
    year = "2025",
    address = "Albuquerque, New Mexico",
    publisher = "Association for Computational Linguistics",
    url = "https://aclanthology.org/2025.naacl-long.412/",
    pages = "8108--8122",
    ISBN = "979-8-89176-189-6"
}

@inproceedings{llms-in-hci,
    author = {H\"{a}m\"{a}l\"{a}inen, Perttu and Tavast, Mikke and Kunnari, Anton},
    title = {Evaluating Large Language Models in Generating Synthetic HCI Research Data: a Case Study},
    year = {2023},
    isbn = {9781450394215},
    publisher = {Association for Computing Machinery},
    address = {New York, NY, USA},
    url = {https://doi.org/10.1145/3544548.3580688},
    doi = {10.1145/3544548.3580688},
    booktitle = {Proceedings of the 2023 CHI Conference on Human Factors in Computing Systems},
    articleno = {433},
    numpages = {19},
    location = {Hamburg, Germany},
    series = {CHI '23}
}

@misc{carrascofarre2024largelanguagemodelspersuasive,
      title={Large Language Models are as persuasive as humans, but how? About the cognitive effort and moral-emotional language of LLM arguments}, 
      author={Carlos Carrasco-Farre},
      year={2024},
      eprint={2404.09329},
      archivePrefix={arXiv},
      primaryClass={cs.CL},
      url={https://arxiv.org/abs/2404.09329}, 
}

@inproceedings{xu2024shadowcast,
    title={Shadowcast: Stealthy Data Poisoning Attacks Against Vision-Language Models},
    author={Yuancheng Xu and Jiarui Yao and Manli Shu and Yanchao Sun and Zichu Wu and Ning Yu and Tom Goldstein and Furong Huang},
    booktitle={The Thirty-eighth Annual Conference on Neural Information Processing Systems},
    year={2024},
    url={https://openreview.net/forum?id=JhqyeppMiD}
}

@misc{havin2025aichangemind,
      title={Can (A)I Change Your Mind?}, 
      author={Miriam Havin and Timna Wharton Kleinman and Moran Koren and Yaniv Dover and Ariel Goldstein},
      year={2025},
      eprint={2503.01844},
      archivePrefix={arXiv},
      primaryClass={cs.CL},
      url={https://arxiv.org/abs/2503.01844}, 
}

@inproceedings{liu-etal-2025-propainsight,
    title = "{P}ropa{I}nsight: Toward Deeper Understanding of Propaganda in Terms of Techniques, Appeals, and Intent",
    author = "Liu, Jiateng  and
      Ai, Lin  and
      Liu, Zizhou  and
      Karisani, Payam  and
      Hui, Zheng  and
      Fung, Yi  and
      Nakov, Preslav  and
      Hirschberg, Julia  and
      Ji, Heng",
    editor = "Rambow, Owen  and
      Wanner, Leo  and
      Apidianaki, Marianna  and
      Al-Khalifa, Hend  and
      Eugenio, Barbara Di  and
      Schockaert, Steven",
    booktitle = "Proceedings of the 31st International Conference on Computational Linguistics",
    month = jan,
    year = "2025",
    address = "Abu Dhabi, UAE",
    publisher = "Association for Computational Linguistics",
    url = "https://aclanthology.org/2025.coling-main.376/",
    pages = "5607--5628"
}

@misc{borah2025persuasionplayunderstandingmisinformation,
      title={Persuasion at Play: Understanding Misinformation Dynamics in Demographic-Aware Human-LLM Interactions}, 
      author={Angana Borah and Rada Mihalcea and Verónica Pérez-Rosas},
      year={2025},
      eprint={2503.02038},
      archivePrefix={arXiv},
      primaryClass={cs.CL},
      url={https://arxiv.org/abs/2503.02038}, 
}

@inproceedings{kamali-etal-2024-using,
    title = "Using Persuasive Writing Strategies to Explain and Detect Health Misinformation",
    author = "Kamali, Danial  and
      Romain, Joseph D.  and
      Liu, Huiyi  and
      Peng, Wei  and
      Meng, Jingbo  and
      Kordjamshidi, Parisa",
    editor = "Calzolari, Nicoletta  and
      Kan, Min-Yen  and
      Hoste, Veronique  and
      Lenci, Alessandro  and
      Sakti, Sakriani  and
      Xue, Nianwen",
    booktitle = "Proceedings of the 2024 Joint International Conference on Computational Linguistics, Language Resources and Evaluation (LREC-COLING 2024)",
    month = may,
    year = "2024",
    address = "Torino, Italia",
    publisher = "ELRA and ICCL",
    url = "https://aclanthology.org/2024.lrec-main.1501/",
    pages = "17285--17309"
}

@inproceedings{ma-etal-2025-communication,
    title = "Communication Makes Perfect: Persuasion Dataset Construction via Multi-{LLM} Communication",
    author = "Ma, Weicheng  and
      Zhang, Hefan  and
      Yang, Ivory  and
      Ji, Shiyu  and
      Chen, Joice  and
      Hashemi, Farnoosh  and
      Mohole, Shubham  and
      Gearey, Ethan  and
      Macy, Michael  and
      Hassanpour, Saeed  and
      Vosoughi, Soroush",
    editor = "Chiruzzo, Luis  and
      Ritter, Alan  and
      Wang, Lu",
    booktitle = "Proceedings of the 2025 Conference of the Nations of the Americas Chapter of the Association for Computational Linguistics: Human Language Technologies (Volume 1: Long Papers)",
    month = apr,
    year = "2025",
    address = "Albuquerque, New Mexico",
    publisher = "Association for Computational Linguistics",
    url = "https://aclanthology.org/2025.naacl-long.203/",
    pages = "4017--4045",
    ISBN = "979-8-89176-189-6"
}

@inproceedings{han-etal-2024-lm,
    title = "{LM}-Infinite: Zero-Shot Extreme Length Generalization for Large Language Models",
    author = "Han, Chi  and
      Wang, Qifan  and
      Peng, Hao  and
      Xiong, Wenhan  and
      Chen, Yu  and
      Ji, Heng  and
      Wang, Sinong",
    editor = "Duh, Kevin  and
      Gomez, Helena  and
      Bethard, Steven",
    booktitle = "Proceedings of the 2024 Conference of the North American Chapter of the Association for Computational Linguistics: Human Language Technologies (Volume 1: Long Papers)",
    month = jun,
    year = "2024",
    address = "Mexico City, Mexico",
    publisher = "Association for Computational Linguistics",
    url = "https://aclanthology.org/2024.naacl-long.222/",
    doi = "10.18653/v1/2024.naacl-long.222",
    pages = "3991--4008"
}

@misc{sabour2025humandecisionmakingsusceptibleaidriven,
      title={Human Decision-making is Susceptible to AI-driven Manipulation}, 
      author={Sahand Sabour and June M. Liu and Siyang Liu and Chris Z. Yao and Shiyao Cui and Xuanming Zhang and Wen Zhang and Yaru Cao and Advait Bhat and Jian Guan and Wei Wu and Rada Mihalcea and Hongning Wang and Tim Althoff and Tatia M. C. Lee and Minlie Huang},
      year={2025},
      eprint={2502.07663},
      archivePrefix={arXiv},
      primaryClass={cs.AI},
      url={https://arxiv.org/abs/2502.07663}, 
}

@article{decodingpersuasion2024,
author={Bassi, Davide  and Fomsgaard, Søren  and Pereira-Fariña, Martín },
title={Decoding persuasion: a survey on ML and NLP methods for the study of online persuasion},  
journal={Frontiers in Communication},     
volume={Volume 9 - 2024},
year={2024},
url={https://www.frontiersin.org/journals/communication/articles/10.3389/fcomm.2024.1457433},
doi={10.3389/fcomm.2024.1457433},
issn={2297-900X}}

@article{RadanlievAIEthics,
    author = {Petar Radanliev},
    title = {AI Ethics: Integrating Transparency, Fairness, and Privacy in AI Development},
    journal = {Applied Artificial Intelligence},
    volume = {39},
    number = {1},
    pages = {2463722},
    year = {2025},
    publisher = {Taylor \& Francis},
    doi = {10.1080/08839514.2025.2463722},
    URL = {https://doi.org/10.1080/08839514.2025.2463722},
    eprint = {https://doi.org/10.1080/08839514.2025.2463722}
}

@ARTICLE{RadanlievEthicsResponsibleAI,
    author={Radanliev, Petar  and Santos, Omar  and Brandon-Jones, Alistair  and Joinson, Adam },
    title={Ethics and responsible AI deployment},  
    journal={Frontiers in Artificial Intelligence},    
    volume={Volume 7 - 2024},
    year={2024},
    url={https://www.frontiersin.org/journals/artificial-intelligence/articles/10.3389/frai.2024.1377011},
    doi={10.3389/frai.2024.1377011},
    issn={2624-8212}
}

@article{Godber_Origgi_2023, 
    title={Telling Propaganda from Legitimate Political Persuasion}, 
    volume={20}, 
    DOI={10.1017/epi.2023.10}, 
    number={3},
    journal={Episteme}, 
    author={Godber, Amelia and Origgi, Gloria}, 
    year={2023}, 
    pages={778–797}
}

@book{okeefe2015persuasion,
  title     = {Persuasion: Theory and Research},
  author    = {O'Keefe, Daniel J.},
  year      = {2015},
  edition   = {3rd},
  publisher = {SAGE Publications},
  address   = {Thousand Oaks, CA}
}

@article{lawrence-reed-2019-argument,
    title = "Argument Mining: A Survey",
    author = "Lawrence, John  and
      Reed, Chris",
    journal = "Computational Linguistics",
    volume = "45",
    number = "4",
    month = dec,
    year = "2019",
    address = "Cambridge, MA",
    publisher = "MIT Press",
    url = "https://aclanthology.org/J19-4006/",
    doi = "10.1162/coli_a_00364",
    pages = "765--818"
}

@article{Nguyen_Litman_2018, title={Argument Mining for Improving the Automated Scoring of Persuasive Essays}, volume={32}, url={https://ojs.aaai.org/index.php/AAAI/article/view/12046}, DOI={10.1609/aaai.v32i1.12046}, abstractNote={ &lt;p&gt; End-to-end argument mining has enabled the development of new automated essay scoring (AES) systems that use argumentative features (e.g., number of claims, number of support relations) in addition to traditional legacy features (e.g., grammar, discourse structure) when scoring persuasive essays. While prior research has proposed different argumentative features as well as empirically demonstrated their utility for AES, these studies have all had important limitations. In this paper we identify a set of desiderata for evaluating the use of argument mining for AES, introduce an end-to-end argument mining system and associated argumentative feature sets, and present the results of several studies that both satisfy the desiderata and demonstrate the value-added of argument mining for scoring persuasive essays. &lt;/p&gt; }, number={1}, journal={Proceedings of the AAAI Conference on Artificial Intelligence}, author={Nguyen, Huy and Litman, Diane}, year={2018}, month={Apr.} }

@inproceedings{dusmanu-etal-2017-argument,
    title = "Argument Mining on {T}witter: Arguments, Facts and Sources",
    author = "Dusmanu, Mihai  and
      Cabrio, Elena  and
      Villata, Serena",
    editor = "Palmer, Martha  and
      Hwa, Rebecca  and
      Riedel, Sebastian",
    booktitle = "Proceedings of the 2017 Conference on Empirical Methods in Natural Language Processing",
    month = sep,
    year = "2017",
    address = "Copenhagen, Denmark",
    publisher = "Association for Computational Linguistics",
    url = "https://aclanthology.org/D17-1245/",
    doi = "10.18653/v1/D17-1245",
    pages = "2317--2322"
}

@article{stab-gurevych-2017-parsing,
    title = "Parsing Argumentation Structures in Persuasive Essays",
    author = "Stab, Christian  and
      Gurevych, Iryna",
    journal = "Computational Linguistics",
    volume = "43",
    number = "3",
    month = sep,
    year = "2017",
    address = "Cambridge, MA",
    publisher = "MIT Press",
    url = "https://aclanthology.org/J17-3005/",
    doi = "10.1162/COLI_a_00295",
    pages = "619--659"
}

@article{slonim_autonomous_debating,
  author    = {Noam Slonim and Yonatan Bilu and Carlos Alzate and Roy Bar-Haim and Ben Bogin and Francesca Bonin and Leshem Choshen and Edo Cohen-Karlik and Lena Dankin and Lilach Edelstein and Liat Ein-Dor and Roni Friedman-Melamed and Assaf Gavron and Ariel Gera and Martin Gleize and Shai Gretz and Dan Gutfreund and Alon Halfon and Daniel Hershcovich and Ron Hoory and Yufang Hou and Shay Hummel and Michal Jacovi and Charles Jochim and Yoav Kantor and Yoav Katz and David Konopnicki and Zvi Kons and Lili Kotlerman and Dalia Krieger and Dan Lahav and Tamar Lavee and Ran Levy and Naftali Liberman and Yosi Mass and Amir Menczel and Shachar Mirkin and Guy Moshkowich and Shila Ofek-Koifman and Matan Orbach and Ella Rabinovich and Ruty Rinott and Slava Shechtman and Dafna Sheinwald and Eyal Shnarch and Ilya Shnayderman and Aya Soffer and Artem Spector and Benjamin Sznajder and Assaf Toledo and Orith Toledo-Ronen and Elad Venezian and Ranit Aharonov},
  title     = {An autonomous debating system},
  journal   = {Nature},
  year      = {2021},
  volume    = {591},
  number    = {7850},
  pages     = {379--384},
  doi       = {10.1038/s41586-021-03215-w},
  url       = {https://doi.org/10.1038/s41586-021-03215-w},
  issn      = {1476-4687}
}

@article{atkinson_artificial_argumentation, title={Towards Artificial Argumentation}, volume={38}, url={https://ojs.aaai.org/aimagazine/index.php/aimagazine/article/view/2704}, DOI={10.1609/aimag.v38i3.2704}, abstractNote={The field of computational models of argument is emerging as an important aspect of artificial intelligence research. The reason for this is based on the recognition that if we are to develop robust intelligent systems, then it is imperative that they can handle incomplete and inconsistent information in a way that somehow emulates the way humans tackle such a complex task. And one of the key ways that humans do this is to use argumentation either internally, by evaluating arguments and counterarguments‚ or externally, by for instance entering into a discussion or debate where arguments are exchanged. As we report in this review, recent developments in the field are leading to technology for artificial argumentation, in the legal, medical, and e-government domains, and interesting tools for argument mining, for debating technologies, and for argumentation solvers are emerging.}, number={3}, journal={AI Magazine}, author={Atkinson, Katie and Baroni, Pietro and Giacomin, Massimiliano and Hunter, Anthony and Prakken, Henry and Reed, Chris and Simari, Guillermo and Thimm, Matthias and Villata, Serena}, year={2017}, month={Oct.}, pages={25-36} }

@article{Castagna_2024,
   title={Computational Argumentation-based Chatbots: A Survey},
   volume={80},
   ISSN={1076-9757},
   url={http://dx.doi.org/10.1613/jair.1.15407},
   DOI={10.1613/jair.1.15407},
   journal={Journal of Artificial Intelligence Research},
   publisher={AI Access Foundation},
   author={Castagna, Federico and Kökciyan, Nadin and Sassoon, Isabel and Parsons, Simon and Sklar, Elizabeth},
   year={2024},
   month=aug, pages={1271–1310} }

@inproceedings{donadello2022machine,
  title={Machine Learning for Utility Prediction in Argument-Based Computational Persuasion},
  author={Donadello, Ivan and Hunter, Anthony and Teso, Stefano and Dragoni, Mauro},
  booktitle={Proceedings of the AAAI Conference on Artificial Intelligence},
  volume={36},
  number={5},
  pages={5144--5152},
  year={2022},
  doi={10.1609/aaai.v36i5.20499},
  publisher={AAAI Press}
}

@article{ibm_rank_30,
author = {Gretz, Shai and Friedman, Roni and Cohen, Edo and Toledo, Assaf and Lahav, Dan and Aharonov, Ranit and Slonim, Noam},
year = {2020},
month = {04},
pages = {7805-7813},
title = {A Large-Scale Dataset for Argument Quality Ranking: Construction and Analysis},
volume = {34},
journal = {Proceedings of the AAAI Conference on Artificial Intelligence},
doi = {10.1609/aaai.v34i05.6285}
}

@inproceedings{
dunefsky2025oneshot,
title={One-shot Optimized Steering Vectors Mediate Safety-relevant Behaviors in {LLM}s},
author={Jacob Dunefsky and Arman Cohan},
booktitle={Second Conference on Language Modeling},
year={2025},
url={https://openreview.net/forum?id=teW4nIZ1gy}
}
\vspace{-5pt}

\end{document}